\documentclass{article}
\usepackage[T1]{fontenc}
\usepackage[preprint]{iclr2027_conference}
\usepackage{times}
\usepackage{amsmath,amssymb,graphicx,booktabs,array,tabularx}
\usepackage{hyperref}
\usepackage{url}
\newcommand{\Candidates}{17{,}225}
\newcommand{\CellLines}{17}
\newcommand{\Controls}{21{,}687}
\newcommand{\EligiblePairs}{269{,}769}
\newcommand{\EligibleTargets}{5{,}108}
\newcommand{\EpochBudget}{30}
\newcommand{\Features}{10{,}073}

\newcommand{\FourReplicateMRR}{0.1352}

\newcommand{\FourUnseenMRR}{0.1456}
\newcommand{\HardNegatives}{20}
\newcommand{\HiddenWidth}{128}
\newcommand{\Inferred}{9{,}095}
\newcommand{\Landmarks}{978}
\newcommand{\LearningRate}{\ensuremath{1.00\times10^{-3}}}

\newcommand{\Nodes}{17{,}308}

\newcommand{\OneReplicateMRR}{0.1213}

\newcommand{\OneUnseenMRR}{0.1463}
\newcommand{\Patience}{5}
\newcommand{\Plates}{915}

\newcommand{\RankingMargin}{1}
\newcommand{\RankingWeight}{0.1}
\newcommand{\RetainedPairs}{270{,}331}

\newcommand{\TargetVocabulary}{5{,}119}
\newcommand{\Tissues}{11}
\newcommand{\TrainBatch}{16}
\newcommand{\TrainingSeed}{20260903}
\newcommand{\WeightDecay}{\ensuremath{1.00\times10^{-4}}}
\newcommand{\RequestedUp}{2}
\newcommand{\RequestedDown}{155}
\newcommand{\RequestAmplitude}{2}
\newcommand{\RequestedTotal}{157}
\newcommand{\GeometryRows}{86{,}540}
\newcommand{\TopFiftyCount}{50}
\newcommand{\TopFiftyEdited}{20}
\newcommand{\PyglRank}{6}

\newcommand{\PyglDifference}{+8.8719}
\newcommand{\PyglNodesNinetyAfterThree}{8}
\newcommand{\PyglFirstMemberRank}{5}
\newcommand{\PyglSecondMemberRank}{10}

\newcommand{\PyglSuppressionGapTwo}{+0.54}

\newcommand{\PyglSuppressionGapFour}{+0.09}
\newcommand{\PyglSuppressionCount}{89}
\newcommand{\PyglFinalMessages}{+4.58}
\newcommand{\PyglFinalResidual}{+4.30}
\newcommand{\CcdcRank}{3}

\newcommand{\CcdcNodesNinetyAfterThree}{1}
\newcommand{\CcdcFirstMemberRank}{4}
\newcommand{\CcdcSecondMemberRank}{2}

\newcommand{\CcdcSuppressionGapOne}{+0.92}

\newcommand{\CcdcSuppressionGapThree}{+0.63}

\newcommand{\CcdcSuppressionCount}{96}
\newcommand{\CcdcFinalMessages}{$-0.31$}
\newcommand{\CcdcFinalResidual}{+12.02}

\graphicspath{{figures/}}
\hypersetup{hidelinks,pdftitle={Hypothesis Generation and Semantic Attribution in Inverse Perturbation Modeling},pdfauthor={Pascal Passigan, Manolis Kellis}}
\newcommand{\R}{\mathbb{R}}
\newcommand{\one}{\mathbf{1}}
\newcommand{\silu}{\operatorname{SiLU}}
\newcommand{\LN}{\operatorname{LN}}
\title{Hypothesis Generation\\and Semantic Attribution\\in Inverse Perturbation Modeling}
\author{Pascal Passigan \& Manolis Kellis\\
Massachusetts Institute of Technology\\
\texttt{\{ppxscal,manoli\}@mit.edu}}

\begin{document}
\maketitle

\begin{abstract}
Understanding gene regulatory relationships at scale can help identify intervention points and anticipate unintended effects. However, connecting a black-box model's learned computation to biological knowledge remains difficult, particularly for unfamiliar cellular states. We investigate whether a sparse, typed gene network can generate inspectable hypotheses for steering malignant transcriptional programs toward expression patterns associated with favorable prognosis. Using LINCS L1000 CRISPR/Cas9 profiles paired with matched plate controls, we train a graph neural network to recover the perturbed gene from baseline and endpoint expression. Its nodes combine ProTrek-derived protein representations with expression features; its edges encode signed regulatory relationships and other gene interactions. We use this inverse formulation to nominate interventions for a desired multigene expression change and examine the biological relationships supporting each nomination. We query the model in a specified cancer-cell context with endpoints that upregulate favorable prognostic markers and downregulate unfavorable ones.
To examine the sensitivity of nominations with respect to structures such as gene states, typed messages, and residual connections, we use integrated gradients and conductance along a path from the cell baseline towards the requested expression endpoint. In A549, the model nominates TTK although its expression is unchanged in the request. The account assigns supporting credit through the checkpoint-associated MAD1L1 relationship and opposing credit through the E2F4 transcriptional-regulatory channel, distinguishing their contributions within the same nomination. Across 333 screened nominations in nine held-out cell lines, request-path attribution shows positive line-averaged correspondence with finite message-deletion effects at every transition. Existing TTK knockout measurements show requested-direction agreement across other markers at the 98.2nd percentile (10 wells); ten of seventeen measured cases in the illustrative screen meet the predefined support criterion. Together, these analyses make competing biological accounts examinable through their distinct contributions to a nomination, with evidence from perturbation measurements and experimental literature.
\end{abstract}

\section{Introduction}
Recent interest in the virtual cell has focused attention on perturbation experiments and modeling \citep{bunne2024}. A central goal is to predict the change in transcriptional state following a genetic or small-molecule intervention and generalize to unseen cell contexts. Perturb-seq made such measurements scalable by combining pooled genetic screens with single-cell expression profiling \citep{adamson2016,dixit2016}, later extending to genome-scale surveys \citep{replogle2022}.

Forward models predict expression responses to specified interventions \citep{klein2025cellflow,roohani2024gears,lotfollahi2023cpa,adduri2025state}; inverse methods nominate interventions for a desired state \citep{gonzalez2026}. Both can learn from the same observations. We choose the inverse formulation for two main reasons. First, its input lets us specify a desired multigene expression change. Here we request changes in prognostic biomarkers. Other possible requests could represent movement toward a treatment-sensitive cancer-cell state \citep{bell2019resistance}, lineage maturation in immature cancer cells \citep{sykes2016differentiation}, or a functional beta-cell expression program \citep{pagliuca2014beta}. Second, we want a window into the biology supporting a nomination for that change. For treatment sensitivity, we could examine support associated with treatment-response features and broader proliferation changes; for differentiation, lineage-maturation and cell-cycle features; and for beta-cell function, cell-identity, secretion-related, and stress-response features. Gradient-based attribution separates supporting and opposing contributions from requested changes and named relationships, making the model's support for different biological explanations examinable. These broader applications would require appropriate perturbation data and functional testing in the relevant cellular context. Forward prediction followed by response matching is also a valid approach; our choice concerns the nomination we want to interrogate, without assuming a ranking or computational advantage.

A recent perturbation-prediction benchmark found that the evaluated deep learning methods did not outperform simple linear baselines \citep{ahlmanneltze2025}, while subsequent work showed that conclusions depend on metric calibration \citep{calibrated2026}. We evaluate predictive performance alongside a separate question: how does a requested expression program contribute to an intervention nomination through named biological relationships?

Logistic regression offers a familiar starting point: its coefficients connect named inputs to a logit. In an interacting model, however, contributions depend on other inputs and on the comparison being made. We therefore ask whether the representation can preserve the scientific identities needed to inspect those dependencies. We retain the graph's named relationships while allowing their contributions to depend on the surrounding expression program through nonlinear updates. This provides a model in which to examine how combinations of requested changes support or oppose an intervention nomination.

\begin{figure}[!t]
\centering
\includegraphics[width=\linewidth]{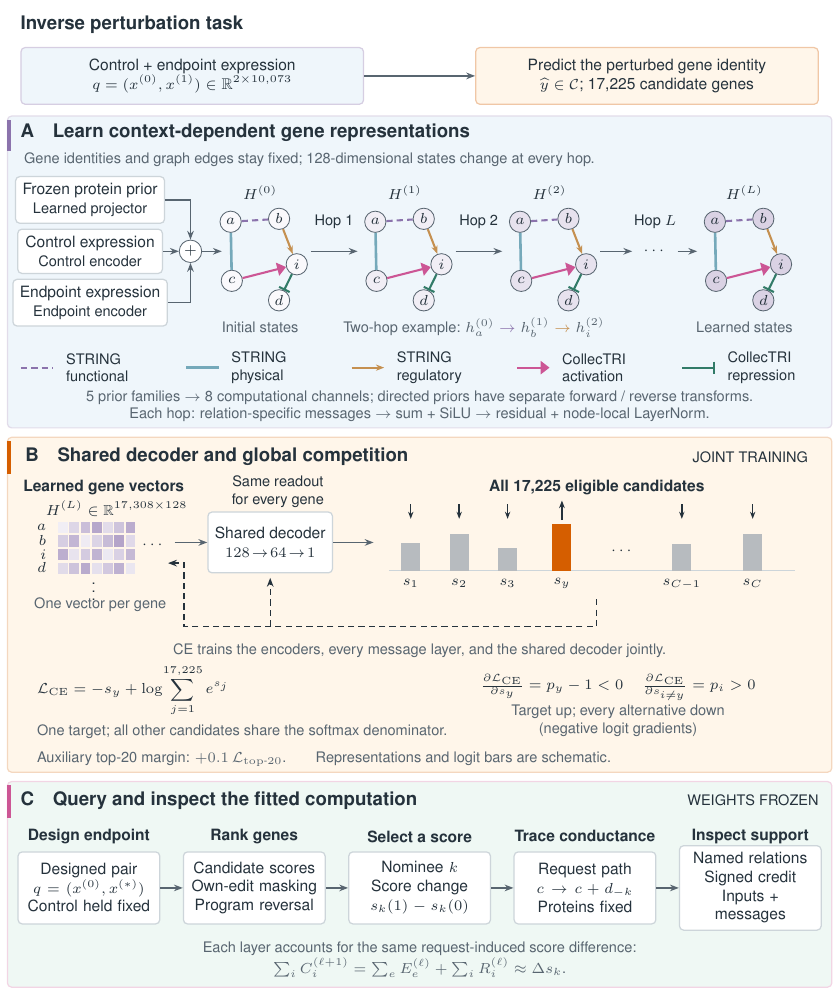}
\caption[Inverse perturbation prediction and computational inspection]{\input{architecture_caption}}
\label{fig:architecture}
\end{figure}

To make these dependencies inspectable, we combine protein features, paired expression, and typed graph messages at gene/protein nodes (Figure~\ref{fig:architecture}). We use semantic attribution to quantify how requested changes contribute to a nomination through those identities and relationships. The graph supplies available relationships; the fitted account quantifies their contributions for a particular request, including opposing contributions from different relationship types connecting the same genes. With 33 distinct neighbors per gene on average, corresponding to approximately 52 incoming typed messages per node excluding self-loops, the graph offers many possible relationships to examine. Their relative contributions can help prioritize hypotheses for subsequent wet-lab investigation. We illustrate this account with TTK in A549, a nomination whose score rises under the requested program and falls when that program is reversed (Section~\ref{sec:cases}), test request-path attributions against finite message deletion across every screened nomination in 9 held-out lines (Section~\ref{sec:suppression}), and compare the requested directions with measured knockout responses (Section~\ref{sec:program}). Predictive evaluation provides context for the learned representations we inspect. The original PYGL and CCDC86 examples are retained as contrasts between own-edit-dependent nomination and measured other-marker support (Appendix~\ref{app:legacy}).

\section{Related work}
\textbf{Perturbation prediction and inverse design.} PDGrapher couples perturbagen discovery with an intervention-conditioned response model \citep{gonzalez2026}. Causal differential networks recover intervention targets from observational and perturbed distributions \citep{wu2025}. Our direct inverse classifier learns on a fixed communication graph.

\textbf{Biologically structured representations.} DCell and DrugCell show how biological entities can organize neural computation \citep{ma2018,kuenzi2020}. We build on this principle using gene-indexed states and relation-specific message passing \citep{bronstein2021,schlichtkrull2018} to inspect individual inverse nominations.

\textbf{Attribution and graph explanations.} Integrated gradients (IG) allocate a scalar difference relative to a reference \citep{sundararajan2017}, while conductance follows it through hidden representations \citep{dhamdhere2019}. For graph explanations, GNNExplainer optimizes subgraph and feature masks \citep{ying2019}, whereas GNN-LRP attributes predictions to higher-order relevant walks \citep{schnake2020}. Our signed allocation through gene states, typed messages, and residuals precedes display selection. Routes then organize marginal transitions without higher-order route scores. Our evaluation concerns signed accounting and finite message effects; visual plausibility alone is not validation \citep{adebayo2018}.

\section{Learning context-dependent gene representations}\label{sec:learning}
We learn these representations by identifying knockouts from paired expression, allowing graph neighbors to shape each gene's state.

\subsection{Observed pairs and gene-indexed states}
For training, each LINCS L1000 observation pairs a CRISPR/Cas9 knockout endpoint with matched plate controls and labels the targeted gene \citep{subramanian2017,clue_perturbagens}. The Level-3 cohort contains \EligiblePairs{} eligible pairs across \CellLines{} lines and \Tissues{} tissues. LINCS BING supplies \Features{} measured and inferred expression coordinates (Appendix~\ref{app:data}). Training-only means and standard deviations normalize baseline and endpoint expression. We test new cell contexts by holding out whole lines in multi-line tissues; singleton tissues permit only physical-replicate holdouts. At each graph depth, we fit three models, H1/H2/H3, on different training/held-out partitions with a shared seed. Their training data overlap. Whole-plate validation selects checkpoints (Appendix~\ref{app:data}).

The \Nodes{} nodes comprise expression genes, retained perturbation-target labels, and their one-hop neighbors in STRING and CollecTRI \citep{szklarczyk2025,mullerdott2023}. Directed families have separate forward/reverse transport channels; computational reversal does not reverse biological causality. At each node,
\begin{equation}
h_i^{(0)}=m_i^pP_\theta(z_i)+m_i^x a_i
\big[B_\theta(x_i^{(0)})+A_\theta(x_i^{(1)})\big].
\label{eq:encode}
\end{equation}
The shared encoders combine frozen ProTrek-derived vectors \citep{su2026} and paired expression. Availability masks suppress missing inputs; $a_i$ masks both expression channels together during training. The protein projector is learned.

Typed messages contextualize these states through repeated updates:
\begin{align}
M_i^{(\ell)}&=\sum_{r,j}\widetilde A^r_{ij}W_{\ell,r}h_j^{(\ell)},\label{eq:message}\\
h_i^{(\ell+1)}&=\LN_\ell\!\left(h_i^{(\ell)}+\silu(M_i^{(\ell)})\right),
\quad\ell=0,\ldots,L-1.\label{eq:update}
\end{align}
We apply SiLU to the aggregated messages before the residual update, providing a smooth nonlinear transformation of the combined neighborhood signal. Adjacency fixes where messages travel; transforms and node-local LayerNorm are learned separately at each layer. The residual carries the current state, including earlier messages. \mbox{$H^{(L)}\in\R^{N\times\HiddenWidth}$} retains one identifiable row per gene.

\subsection{Shared decoding and predictive context}
The shared decoder ranks \Candidates{} nodes with mapped ProTrek vectors; 83 others remain in the computation but are ineligible candidates. Genes without positive training examples remain eligible; unseen-target accuracy is untested. Training uses global cross-entropy (CE) plus a margin against the hardest \HardNegatives{} incorrect candidates (Appendix~\ref{app:training}). Genetic labels keep the intended target explicit; multitarget chemical action is examined later through annotations \mbox{\citep{keiser2009}}.

We compare the GNN with a protein-informed node-local predictor and an expression-only global multilayer perceptron (MLP). Predictive performance depends on the setting: GNN versus node-local mean reciprocal rank (MRR) is 0.1344 versus 0.1150 for held replicates and 0.1433 versus 0.1499 for unseen cells (three-seed tissue-balanced means; Table~\ref{tab:baselines}). These measure observed-knockout recovery, not intervention recommendation quality. The node-local candidate's raw score cannot depend on other genes' edits; the global MLP can use those edits and support input attribution. We use the GNN to examine named, typed relationships that participate in the cross-gene computation producing the score. The fixed-budget comparison does not establish best-achievable graph-free performance. Depth comparisons likewise do not establish a uniformly preferred depth (Appendix~\ref{app:training}).

\section{From a desired program to a nomination}\label{sec:request}
After training, we freeze the weights and replace the observed endpoint with a requested profile to nominate interventions. Human Protein Atlas (HPA) survival associations supply expression directions, not causal effects \citep{uhlen2017,hpa_validation,hpa_methods}. For A549, validation-tier markers in \emph{Lung Adenocarcinoma (validation)} specify \RequestedUp{} increases and \RequestedDown{} decreases, including decreases in the unfavorable markers MAD2L1, BUB1B and PLK1. Each edit has amplitude \RequestAmplitude{} H1-reference SDs; other coordinates remain unchanged. HPA supplies the directions; PRECOG provides a separate concordance check, not a requirement for selecting markers (Appendix~\ref{app:query-audit}).

One native baseline/endpoint pair is transformed through each model's normalizer. Four-hop partition models H1 and H3 both held A549 out of training and validation; their mean raw logit defines the ranking. Unfavorable markers requested to decrease, such as the originally illustrated PYGL and CCDC86, remain eligible knockout candidates. Directly reducing an unfavorable marker agrees with its requested direction; a single-gene perturbation may also affect other coordinates. Candidates are ranked against the complete multigene request. Of the top \TopFiftyCount{} nominees, \TopFiftyEdited{} have edited own inputs. We therefore test whether nominations persist when their own requested edit is removed (Section~\ref{sec:cases}).

\section{Following a nomination through nodes and messages}\label{sec:explanation}
A nomination becomes more informative when we can identify the requested edits that support it and the relationships through which the model combines that information. We trace the score change from no change to the requested endpoint.

\subsection{Fix the question and the path}\label{sec:request-path}
For nominee $k$, conductance explains a score difference between a reference input and a target input along a straight path, with weights and graph fixed. Write $p_i=P_\theta(z_i)$ and $r_i(x^{(0)},x^{(1)})=m_i^x[B_\theta(x_i^{(0)})+A_\theta(x_i^{(1)})]$, the two terms of Equation~\ref{eq:encode} with inference mask $a_i=1$. In general, either the expression endpoint or the proteins move:
\begin{align}
h_i^{(0)}(\alpha)&=r_i\big(x^{(0)},x^{(1)}(\alpha)\big)+m_i^p\,p_i(\alpha),\label{eq:path}\\
\Delta s_k&=s_k(1)-s_k(0),\qquad 0\le\alpha\le1.\label{eq:difference}
\end{align}
\textbf{Request path (Section~\ref{sec:cases}).} Actual proteins stay fixed, $p_i(\alpha)=p_i$, and the endpoint moves from no change to the request. For baseline $c$ and request $d$, let $d_{-k}$ set candidate $k$'s own edit to zero; then $x^{(0)}=c$ and $x^{(1)}(\alpha)=c+\alpha d_{-k}$. For member $f$, $\Delta s_k$ is the raw response $a_{kf}=s_{kf}(c,c+d_{-k})-s_{kf}(c,c)$; the reversed response is $b_{kf}=s_{kf}(c,c-d_{-k})-s_{kf}(c,c)$. Expression IG allocates $a_{kf}$ to individual requested edits (Appendix Equation~\ref{eq:expression-ig}), and the conductances below carry the same quantity through nodes, messages, and residuals. This path asks which requested changes move a nomination.

\textbf{Protein path.} With the request fixed, projected proteins move jointly from their common mean to their actual values (Appendix~\ref{app:numerics}). This examines protein identity in the requested context. Its totals and signs are not interchangeable with the request path.

\textbf{Measure what each state and connection carries.} Conductance combines the score's sensitivity to a representation with that representation's movement along the path \citep{dhamdhere2019}. The node value is
\begin{equation}
C_i^{(\ell)}=\int_0^1
\left\langle\frac{\partial s_k}{\partial h_i^{(\ell)}},
\frac{d h_i^{(\ell)}}{d\alpha}\right\rangle d\alpha.
\label{eq:node-conductance}
\end{equation}
At the initial state, $C_i^{(0)}$ is input IG summed within gene $i$: over requested expression edits on the request path and over protein features on the protein path. After message passing, $C_i^{(\ell)}$ follows the allocation carried by that gene's current representation, which can include earlier information from other genes. Every value concerns the same final nominated logit.

For an edge $e=(j\to i,r)$, let $m_e^{(\ell)}=\widetilde A^r_{ij}W_{\ell,r}h_j^{(\ell)}$. The receiver sums these messages into $M_i^{(\ell)}$, while $u_i^{(\ell)}=h_i^{(\ell)}+\silu(M_i^{(\ell)})$ is the pre-normalization state. Their two branches have separate conductances:
\begin{align}
E_e^{(\ell)}&=\int_0^1\left\langle
\frac{\partial s_k}{\partial M_i^{(\ell)}},
\frac{d m_e^{(\ell)}}{d\alpha}\right\rangle d\alpha,\label{eq:message-conductance}\\
R_i^{(\ell)}&=\int_0^1\left\langle
\frac{\partial s_k}{\partial u_i^{(\ell)}},
\frac{d h_i^{(\ell)}}{d\alpha}\right\rangle d\alpha.\label{eq:residual-conductance}
\end{align}
The residual carries the current representation, including earlier messages. The chain rule reconciles these contributions with the receiving node and the same total difference:
\begin{equation}
C_i^{(\ell+1)}=\sum_{e:e\to i}E_e^{(\ell)}+R_i^{(\ell)},\qquad
\sum_i C_i^{(\ell+1)}=\sum_eE_e^{(\ell)}+\sum_iR_i^{(\ell)}=\Delta s_k.
\label{eq:accounting}
\end{equation}
At each layer, node states provide one account; incoming messages and residuals provide the corresponding branch account. Each complete layer recovers the same difference, so layer totals are not added; on the request path, summed input IG equals initial node conductance and the largest layer discrepancy is $1.58\times10^{-8}$ (Appendix~\ref{app:query-audit}). Positive and negative values support or oppose that logit difference; relationship signs retain their separate biological-prior meaning.

\textbf{Select a readable view.} H1 and H3 receive the same native request. We average corresponding signed scalar contributions, not hidden feature coordinates. A positive mean can hide disagreement, so TTK routes require positive credit at every transition in both fits; the original protein-path displays use a positive mean (Appendix~\ref{app:display}).

\begin{figure}[!t]
\centering
\includegraphics[width=\linewidth]{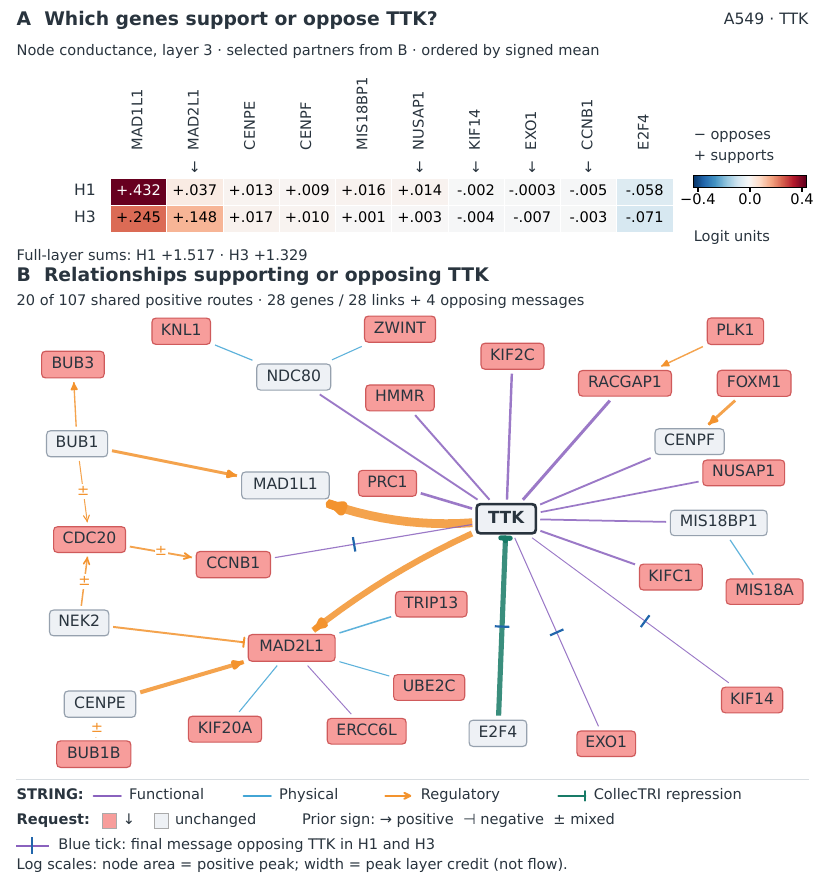}
\caption{\textbf{MAD1L1 carries strong support for TTK; E2F4 and CCNB1 contribute opposition.} A: one signed strip of node conductance at $h^{(3)}$, before the final message step. Among partners shown in B, excluding TTK, we select the six strongest shared-positive and four strongest shared-negative states, then order them by signed H1/H3 mean. Arrows mark requested expression changes; a blank means no edit. Red supports and blue opposes the request-induced score change. Full-layer sums include all nodes, not just this selection. B: twenty shared positive routes (28 genes, 28 relationships) with four shared opposing final messages marked by blue ticks. E2F4 dominates this shared opposition; CCNB1, EXO1 and KIF14 are much smaller. Node fill denotes the request; colors and arrowheads retain biological-prior meanings, not attribution sign or transport direction. Width uses a common logarithmic magnitude scale. This is a route-based view, not a global ranking of edges. Selection and coverage: Appendix~\ref{app:ttk-figure}; explicit layers and transport: Figure~\ref{fig:ttk-signed-routes}.}

\label{fig:ttk-case}
\end{figure}

Edited markers anchor widest positive routes through messages and residuals. We select complete member route witnesses, not paths assembled from unrelated fragments. Figure~\ref{fig:ttk-case} overlays final-message opposition on the shared positive core; Appendix Figure~\ref{fig:ttk-signed-routes} separately shows routes of both signs. Signed profiles and per-layer coverage report what the partial display omits (Appendix~\ref{app:display}).

\section{A program-driven nomination}\label{sec:cases}
We first ask whether a nomination is supported by the other requested changes, beyond the requested decrease in the nominated gene itself. To test this, we restore only that gene's endpoint expression to its baseline value and keep the rest of the request unchanged. PYGL and CCDC86 lose their high rankings under this control, showing that their own requested decreases strongly influence their nominations. Their measured knockouts provide separate evidence about effects on the other requested markers (Table~\ref{tab:request-controls-main}). We therefore screen for candidates whose scores rise under the remaining request and fall under its reversal in every eligible model. Requiring candidate-specific masked rank $\le200$ and mean $(a_{kf}-b_{kf})/2\ge0.5$ yields 19 cases in six of fourteen lines. We use TTK/A549 as a worked example because its own expression was never edited. This choice is retrospective; the population test includes every nomination passing without the illustrative 0.5 floor.

\begin{table}[t]
\centering\small
\caption{\textbf{Dependence on the nominee's own expression change and measured effects on other markers.} A549 ranks are shown for the separately fitted H1/H3 models, both of which exclude A549 from training and validation. H2 trained on A549 and is not included here. For each nominee, we restore its requested endpoint expression to baseline, leave all other requested changes unchanged, and score all candidates again. TTK's input is already unchanged, so its ranks remain 18/17. Measured percentiles compare knockout agreement with the requested directions across the other markers, excluding the nominee's own marker (Section~\ref{sec:program}). Lower ranks and higher percentiles indicate stronger nomination scores and measured agreement, respectively.}
\label{tab:request-controls-main}
\begin{tabular}{lrrr}
\toprule
Nominee & \shortstack{Rank with\\full request} & \shortstack{Rank with nominee's\\expression unchanged} & \shortstack{Measured agreement\\percentile (wells)} \\
\midrule
TTK & 18/17 & 18/17 (no own edit) & 98.2 (10) \\
PYGL & 5/10 & 3084/2214 & 70.8 (17) \\
CCDC86 & 4/2 & 2670/2455 & 88.5 (9) \\
\bottomrule
\end{tabular}
\end{table}

\subsection{TTK in A549: from requested edits to an unedited intermediary}
The A549 request does not ask the model to decrease TTK. Nevertheless, the checkpoint kinase TTK/MPS1 \citep{zheng2019ttk} ranks 2/3 under the candidate-specific masked screen in H1/H3. In this screening procedure, each competing candidate is scored with its own requested change removed. TTK's input does not change, but its competitors' scores can change, explaining the difference from its 18/17 ranks in Table~\ref{tab:request-controls-main}. The requested changes raise its score by $+1.517/{+}1.329$; reversing them lowers it by $-1.834/{-}1.216$. Both models held A549 out, so we can ask what in the other genes' requested changes supports this nomination.

At the input, decreasing MAD2L1 supplies the strongest support in both models ($+0.251/{+}0.236$). Decreases in BUB1B and PLK1 also support both fits. Other edits behave differently: BRCA1 down opposes in H1, and TK1 down opposes in H3. The model therefore does not assign the same sign to every decrease in this program (Table~\ref{tab:ttk-biology}).

Figure~\ref{fig:ttk-case}A follows the combined request into the gene states before the final message step. MAD1L1 carries substantial node conductance ($+0.432/{+}0.245$), despite zero input IG: its later representation carries information received from other genes. The strip orders selected partners from support to opposition, including CCNB1 and E2F4; arrows mark only requested expression changes. Panel B then shows the relationships carrying the account. The final MAD1L1-to-TTK reverse-regulatory message supplies the strongest supporting relationship in the mean account ($+0.420/{+}0.150$). E2F4 supplies substantial shared opposition ($-0.058/{-}0.071$). Of all 99 final typed messages into TTK, 43 support it and four oppose it in both fits; the other shared opponents, CCNB1, KIF14 and EXO1, are much smaller (each below 0.007 in magnitude). Thus an unchanged input can become an important intermediary, while another relationship in the same neighborhood opposes the nomination.

These allocations suggest a specific biological reading. MPS1 phosphorylates KNL1, BUB1 and MAD1 (encoded by MAD1L1) to organize checkpoint signaling \citep{ji2017mps1}; MAD2 and BUBR1, encoded by MAD2L1 and BUB1B, participate in the mitotic checkpoint complex \citep{sudakin2001mcc}. PLK1 strengthens MPS1-dependent signaling \citep{vonschubert2015plk1}. Support from the requested decreases in these components, together with the unedited MAD1L1 intermediary, is consistent with nominating loss of their checkpoint coordinator. In contrast, E2F4/DREAM regulates TTK transcription \citep{muller2017timing}, and its channel opposes the nomination. The account distinguishes checkpoint-associated support from transcriptional-regulatory opposition; it does not establish which mechanism operates in A549 (Table~\ref{tab:ttk-biology}).

The dependence extends beyond the immediate neighborhood: 30.2\%/26.3\% of absolute input IG originates outside TTK's 92 immediate prior neighbors. Named inputs and final MAD1L1 support also persist at a smaller request size. Recurrence across H1/H3 tests partition sensitivity: these separately trained models share a seed and 64\% of training wells. The DANG comparison changes the request, cell and model together and shows only partial recurrence (Appendix~\ref{app:ttk-figure}). Section~\ref{sec:program} then asks whether existing knockout measurements support the requested direction.

\textbf{The same partner can contribute through opposing channels.} The distinction extends beyond choosing which neighboring gene matters. For SORL1/A549, APP's physical channel supports the nomination in H1/H3 ($+0.0436/{+}0.0108$), while its functional channel opposes it ($-0.0149/{-}0.0062$). For CAMSAP2/DANG in H2, CKAP2 likewise supports through its physical channel ($+0.0604$) and opposes through its functional channel ($-0.0301$). These are final-transition ($3\to4$) request-path contributions in score units. Collapsing each pair to one unlabeled edge would hide the opposing allocations. Together with TTK, they illustrate how a quantitative query yields relationship-specific hypotheses beyond a target name. Appendix~\ref{app:competing-accounts} retains all 19 cases with coverage labels and expanded accounts for six complete-field cases. SORL1/A549 uses complete fields; CAMSAP2/DANG remains panel-based. Opposing channels remain active in the computation; their signs do not establish mutually exclusive biological mechanisms.

\begin{figure}[!t]
\centering
\includegraphics[width=\linewidth]{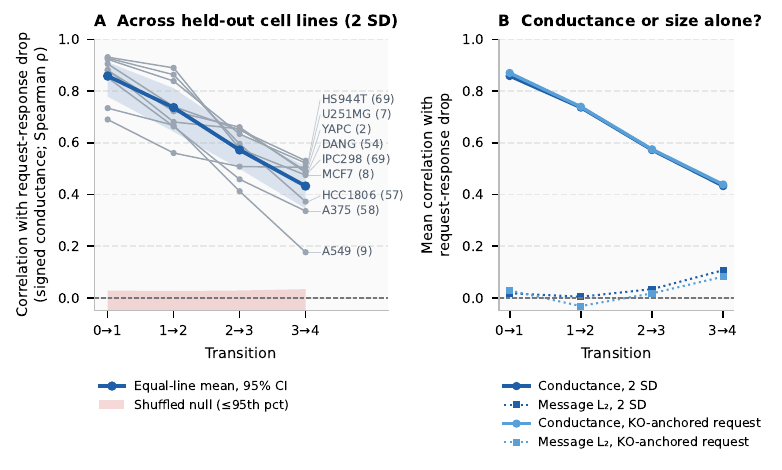}
\caption{\textbf{Mean attribution--deletion correspondence is positive in all nine held-out lines and weakens with depth.} Both panels use request-response drop: how much deleting one message changes the nomination's request-minus-no-change score difference. A: per-line Spearman correlation with signed conductance at the two-SD request; parentheses give nominations per line. Blue shows the equal-line mean and 95\% bootstrap interval over lines and cases; red shows the shuffled null up to its 95th percentile. B: does signed conductance track request-response drop better than raw message size ($L_2$ norm at the requested endpoint)? Solid lines show conductance; dotted lines show message size. Colors distinguish two-SD edits from smaller edits scaled to observed knockout displacements (Section~\ref{sec:program}). Message size is unsigned; this comparison tests signed effects, not their absolute magnitudes. Displayed means weight lines equally; both panels use stratified messages (signed extremes plus controls), not all edges. Pre-declared protocol, Appendix~\ref{app:request-population}.}
\label{fig:request-population}
\end{figure}

\subsection{Checking conductance with finite message suppression}\label{sec:suppression}
We next test the computational account by suppressing messages. We remove one message at a time and ask how much this changes the nominee's response to the request. We measure that response as the difference between its score at the requested endpoint and its score when no expression change is requested. We then compare the reduction in this response with the message's signed conductance. We suppress one typed, directed message at one transition and recompute downstream states (Appendix~\ref{app:request-population}). For member $f$, let $s_{k,f}^{\setminus e}(\alpha)$ denote the score after removing message $e$, where endpoints 0 and 1 are no change and the requested expression. \mbox{The \emph{request-response drop} (Gap drop) is}
\begin{equation}
D^{\rm gap}_{e,f}=[s_{k,f}(1)-s_{k,f}(0)]
-[s_{k,f}^{\setminus e}(1)-s_{k,f}^{\setminus e}(0)].\label{eq:suppression-gap}
\end{equation}
This subtracts the deletion effect at no change from its effect at the requested endpoint, comparing the same nominee in both conditions. It is a finite intervention check, not an attribution identity; no integral is recomputed. Population correlations are computed within each member's panel, then averaged over members, nominees and lines equally. The original protein-path case panels instead compare member-averaged drops with mean signed conductance (Appendix~\ref{app:suppression}).

\textbf{Request-path population study.} We pre-declared this test (protocol, pool, and decision rule frozen before any run; Appendix~\ref{app:request-population}). The pool is every nomination passing the screen without its 0.5 floor; in the 9 held-out lines it contains 333 nominations (359 model runs; five lines contribute one eligible model). Nominees within a line share one request, so we weight lines equally. Each run's panel holds signed extremes plus low-$|E|$ and random controls, fixed before deletion. Mean within-panel Spearman correlation between signed message conductance and request-response drop is positive in all 9 lines at every transition, with equal-line means 0.86/0.74/0.57/0.43 from $0\to1$ to $3\to4$; no permuted mean reaches any observed value ($p=0.001$, 1,000 permutations). At the observed-knockout-anchored request size the means are 0.87/0.74/0.57/0.44, again positive in all 9 lines. We also ask whether a message's size alone tracks its deletion effect: its $L_2$ norm at the requested endpoint gives correlations of 0.00--0.11 with request-response drop at two SD (Figure~\ref{fig:request-population}B). Correspondence declines with depth, and TTK's final-transition case panel is negative in both fits (Table~\ref{tab:expression-suppression}). The allocation is therefore an imperfect predictor of the deletion effect, especially at later layers. Pairwise deletion and normalization controls do not establish a general explanation for this disagreement (Appendix~\ref{app:normalization-repair}). These tests assess the fitted computation; they do not establish that displayed routes are necessary biological mechanisms.

\textbf{Separate protein-path checks.} These show positive mean deletion correspondence but weak early-layer cross-refit agreement; they are not request-path independent-seed replications (Appendices~\ref{app:population}--\ref{app:controls}).

\section{Testing the requested program against measurements}\label{sec:program}
\textbf{Measured responses.} The signed account makes a hypothesis about TTK worth examining, but the computation alone cannot show what a knockout does. We therefore compare existing LINCS knockout wells with the requested marker directions, excluding the target's own marker. Measured TTK knockout agreement reaches the 98.2nd percentile among 5,115 A549 targets (10 wells). The input-attribution profile also corresponds to measured requested-direction marker displacement ($\rho=0.492$; 98.8th percentile against 4,654 other knockouts with at least three wells; Appendix~\ref{app:ttk-knockout}). The first statistic measures program agreement; the second asks whether the more strongly credited markers move further in the requested direction.

The full 19-case computational screen gives a mixed picture: ten cases meet the pre-set measured-support criterion ($\geq$75th percentile), seven are at or below the 50th percentile, and two are unmeasured (Table~\ref{tab:screen-measured}). These are retrospective, case-specific results. TTK's high percentile corresponds to a mean signed displacement of 0.418 reference SD, far below a uniform two-SD shift. For comparison, CKS2/HS944T reaches the 95.0th percentile with a mean displacement of 0.074 SD (six wells). The measurements support directions for particular nominations, not attainment of the whole requested vector.

\textbf{Request size.} Because two-SD edits exceed typical knockout displacements, we also query sizes anchored to the measurements. At 0.527 SD in A549, TTK retains a positive request response and negative reversed response, with masked ranks 14/7; MAD2L1 remains its strongest input. CKS2/HS944T weakens at 0.275 SD: its rank falls from 10 to 54 and response from $+0.768$ to $+0.050$ (Appendix~\ref{app:amplitude}). Even TTK's persistent nomination does not imply an unchanged internal account: H3's $2\to3$ message-field cosine is $-0.172$ across sizes (Table~\ref{tab:field-shape}).

\textbf{What existing signatures already tell us.} We include a retrospective reference that scores each knockout's measured responses in other lines for agreement with the requested marker directions (4,970 common A549 targets; Table~\ref{tab:forward}). It asks how informative those existing responses are for the query. It is not a trained forward model and does not reproduce each inverse fold's information boundary, so it does not establish comparative ranking performance.

\textbf{Compound context.} The explorer links nominees to source-reported drug annotations (Appendix~\ref{app:compounds}). These do not establish efficacy; an inhibitor need not reproduce a knockout.

\section{Limitations and conclusion}\label{sec:limits}
Observed-pair recovery does not establish intervention efficacy. Survival associations specify directions, not causal effects or amplitudes, and bulk patient cohorts only coarsely match cell-line contexts. L1000 averages cell populations, so differences may reflect within-cell changes or shifts in cell-state proportions; these accounts do not isolate phase-specific mechanisms. Training does not filter by technical quality control (QC). The original cases' baseline-sensitivity checks do not extend to TTK or CKS2 (Appendix~\ref{app:qc-sensitivity}). Protein references and interpolation are computational comparisons; biological priors and suppression checks do not establish that a displayed route is a necessary mechanism.

The resulting account connects a quantitative expression request to biological relationships that can be examined. TTK separates recurring checkpoint-associated support from transcriptional-regulatory opposition; SORL1 and CAMSAP2 show why even the same partner cannot be assigned one contribution sign across relationship types. Population suppression assesses computational correspondence, while measured knockout responses provide complementary, case-specific evidence. Together, these results make intervention hypotheses more specific and open to experimental challenge. The biological mechanism responsible for the measured response remains to be tested.

\clearpage
\section*{AI use statement}
Generative AI tools assisted with conceptual framing and methodology design, implementation and debugging, data-cleaning and identifier-mapping code, numerical verification code, interpretation of saved results, literature discovery, figure preparation, and drafting and editing this manuscript. Expression requests follow an explicit rule applied to the stated source data. Reported numerical results are outputs of executed code, not numbers generated by a language model. Automated checks cover source and artifact identities, splits and coordinates, checkpoint restoration, selected derivative comparisons, attribution completeness and refinement, and manuscript exports. AI-assisted visual checks informed layout. These checks are not independent biological validation and do not replace the authors' responsibility for the final content.

\section*{Reproducibility statement}
The notebook presents the observations, model, learning procedure, request, and conductance calculations beside their shared implementation. The recorded load-mode execution has 39 code cells and identifies both original protein-reference cases, their member calculations, signed arrays, profiles, and finite suppression results. Frozen paper exports retain source hashes and numerical locators and build without raw assay matrices or model checkpoints. Code, model weights, the notebook, frozen evidence, and the supplementary outputs cited in the text will be released upon publication. Appendix~\ref{app:reproduction} distinguishes this current evidence from a separate historical full-training execution.

\bibliography{references}

\begin{thebibliography}{57}
\providecommand{\natexlab}[1]{#1}
\providecommand{\url}[1]{\texttt{#1}}
\expandafter\ifx\csname urlstyle\endcsname\relax
  \providecommand{\doi}[1]{doi: #1}\else
  \providecommand{\doi}{doi: \begingroup \urlstyle{rm}\Url}\fi

\bibitem[Adamson et~al.(2016)Adamson, Norman, Jost, Cho, Nu{\~n}ez, Chen,
  Villalta, Gilbert, Horlbeck, Hein, Pak, Gray, Gross, Dixit, Parnas, Regev,
  and Weissman]{adamson2016}
Britt Adamson, Thomas~M. Norman, Marco Jost, Min~Y. Cho, James~K. Nu{\~n}ez,
  Yuwen Chen, Jacqueline~E. Villalta, Luke~A. Gilbert, Max~A. Horlbeck,
  Marco~Y. Hein, Ryan~A. Pak, Andrew~N. Gray, Carol~A. Gross, Atray Dixit, Oren
  Parnas, Aviv Regev, and Jonathan~S. Weissman.
\newblock {A Multiplexed Single-Cell CRISPR Screening Platform Enables
  Systematic Dissection of the Unfolded Protein Response}.
\newblock \emph{Cell}, 167\penalty0 (7):\penalty0 1867--1882.e21, 2016.
\newblock \doi{10.1016/j.cell.2016.11.048}.

\bibitem[Adduri et~al.(2025)Adduri, Gautam, Bevilacqua, Imran, Shah,
  Naghipourfar, Teyssier, Ilango, Nagaraj, Dong, Ricci-Tam, Carpenter,
  Subramanyam, Winters, Tirukkovular, Sullivan, Plosky, Eraslan, Youngblut,
  Leskovec, Gilbert, Konermann, Hsu, Dobin, Burke, Goodarzi, and
  Roohani]{adduri2025state}
Abhinav~K. Adduri, Dhruv Gautam, Beatrice Bevilacqua, Alishba Imran, Rohan
  Shah, Mohsen Naghipourfar, Noam Teyssier, Rajesh Ilango, Sanjay Nagaraj,
  Mingze Dong, Chiara Ricci-Tam, Christopher Carpenter, Vishvak Subramanyam,
  Aidan Winters, Sravya Tirukkovular, Jeremy Sullivan, Brian~S. Plosky, Basak
  Eraslan, Nicholas~D. Youngblut, Jure Leskovec, Luke~A. Gilbert, Silvana
  Konermann, Patrick~D. Hsu, Alexander Dobin, Dave~P. Burke, Hani Goodarzi, and
  Yusuf~H. Roohani.
\newblock {Predicting cellular responses to perturbation across diverse
  contexts with State}.
\newblock \emph{bioRxiv}, 2025.
\newblock \doi{10.1101/2025.06.26.661135}.
\newblock URL
  \url{https://www.biorxiv.org/content/10.1101/2025.06.26.661135v1}.
\newblock Preprint, version 1.

\bibitem[Adebayo et~al.(2018)Adebayo, Gilmer, Muelly, Goodfellow, Hardt, and
  Kim]{adebayo2018}
Julius Adebayo, Justin Gilmer, Michael Muelly, Ian Goodfellow, Moritz Hardt,
  and Been Kim.
\newblock {Sanity Checks for Saliency Maps}.
\newblock In \emph{Advances in Neural Information Processing Systems},
  volume~31, 2018.
\newblock URL
  \url{https://papers.nips.cc/paper/8160-sanity-checks-for-saliency-maps}.

\bibitem[Ahlmann-Eltze et~al.(2025)Ahlmann-Eltze, Huber, and
  Anders]{ahlmanneltze2025}
Constantin Ahlmann-Eltze, Wolfgang Huber, and Simon Anders.
\newblock {Deep-learning-based gene perturbation effect prediction does not yet
  outperform simple linear baselines}.
\newblock \emph{Nature Methods}, 22\penalty0 (8):\penalty0 1657--1661, 2025.
\newblock \doi{10.1038/s41592-025-02772-6}.

\bibitem[Bell et~al.(2019)Bell, Fennell, Chan, et~al.]{bell2019resistance}
Charles~C. Bell, Katie~A. Fennell, Yih-Chih Chan, et~al.
\newblock Targeting enhancer switching overcomes non-genetic drug resistance in
  acute myeloid leukaemia.
\newblock \emph{Nature Communications}, 10:\penalty0 2723, 2019.
\newblock \doi{10.1038/s41467-019-10652-9}.

\bibitem[{Broad Institute}(2017)]{clue_perturbagens}
{Broad Institute}.
\newblock {Perturbagen types and controls}.
\newblock \url{https://clue.io/connectopedia/perturbagen_types_and_controls},
  2017.
\newblock Page last modified December 8, 2017; accessed September 25, 2026.

\bibitem[{Broad Institute}(2018{\natexlab{a}})]{cluecrisprqc}
{Broad Institute}.
\newblock {CRISPR v1.0 Data release README}.
\newblock
  \url{https://s3.amazonaws.com/assets.clue.io/clue/public/resources/Broad_Institute_LINCS_CRISPRv1.0_early_access_data_deposition_README.pdf},
  2018{\natexlab{a}}.
\newblock July 2018 early-access release documentation; accessed September 25,
  2026.

\bibitem[{Broad Institute}(2018{\natexlab{b}})]{clueqc}
{Broad Institute}.
\newblock {Is the L1000 data "fully QC'd"?}
\newblock \url{https://clue.io/connectopedia/l1000_qc}, 2018{\natexlab{b}}.
\newblock Page last modified January 5, 2018; accessed September 25, 2026.

\bibitem[{Broad Institute}(2025)]{broad2025}
{Broad Institute}.
\newblock {Drug Repurposing Hub: drug information annotation snapshot}.
\newblock
  \url{https://repo-hub.broadinstitute.org/public/data/repo-drug-annotation-20250818.txt},
  2025.
\newblock Snapshot dated August 18, 2025; checksum and queried rows retained
  with the code.

\bibitem[Bronstein et~al.(2021)Bronstein, Bruna, Cohen, and
  Veli{\v{c}}kovi{\'c}]{bronstein2021}
Michael~M. Bronstein, Joan Bruna, Taco Cohen, and Petar Veli{\v{c}}kovi{\'c}.
\newblock {Geometric Deep Learning: Grids, Groups, Graphs, Geodesics, and
  Gauges}, 2021.
\newblock URL \url{https://arxiv.org/abs/2104.13478v2}.
\newblock Version 2, revised May 2, 2021.

\bibitem[Bunne et~al.(2024)Bunne, Roohani, Rosen, Gupta, Zhang, Roed,
  Alexandrov, AlQuraishi, Brennan, Burkhardt, Califano, Cool, Dernburg, Ewing,
  Fox, Haury, Herr, Horvitz, Hsu, Jain, Johnson, Kalil, Kelley, Kelley,
  Kreshuk, Mitchison, Otte, Shendure, Sofroniew, Theis, Theodoris, Upadhyayula,
  Valer, Wang, Xing, Yeung-Levy, Zitnik, Karaletsos, Regev, Lundberg, Leskovec,
  and Quake]{bunne2024}
Charlotte Bunne, Yusuf Roohani, Yanay Rosen, Ankit Gupta, Xikun Zhang, Marcel
  Roed, Theo Alexandrov, Mohammed AlQuraishi, Patricia Brennan, Daniel~B.
  Burkhardt, Andrea Califano, Jonah Cool, Abby~F. Dernburg, Kirsty Ewing,
  Emily~B. Fox, Matthias Haury, Amy~E. Herr, Eric Horvitz, Patrick~D. Hsu,
  Viren Jain, Gregory~R. Johnson, Thomas Kalil, David~R. Kelley, Shana~O.
  Kelley, Anna Kreshuk, Tim Mitchison, Stephani Otte, Jay Shendure, Nicholas~J.
  Sofroniew, Fabian Theis, Christina~V. Theodoris, Srigokul Upadhyayula, Marc
  Valer, Bo~Wang, Eric Xing, Serena Yeung-Levy, Marinka Zitnik, Theofanis
  Karaletsos, Aviv Regev, Emma Lundberg, Jure Leskovec, and Stephen~R. Quake.
\newblock {How to build the virtual cell with artificial intelligence:
  Priorities and opportunities}.
\newblock \emph{Cell}, 187\penalty0 (25):\penalty0 7045--7063, 2024.
\newblock \doi{10.1016/j.cell.2024.11.015}.

\bibitem[Chen et~al.(2011)]{nfytop2011}
C.~F. Chen et~al.
\newblock Novel regulation of nuclear factor-{YB} by {miR-485-3p} affects the
  expression of {DNA} topoisomerase {II} alpha and drug responsiveness.
\newblock \emph{Molecular Pharmacology}, 2011.
\newblock \doi{10.1124/mol.110.069633}.

\bibitem[Chini \& Chen(2003)Chini and Chen]{claspin2003}
Claudia~C. Chini and Junjie Chen.
\newblock Human claspin is required for replication checkpoint control.
\newblock \emph{Journal of Biological Chemistry}, 2003.
\newblock \doi{10.1074/jbc.M301136200}.

\bibitem[Corsello et~al.(2017)Corsello, Bittker, Liu, Gould, McCarren,
  Hirschman, Johnston, Vrcic, Wong, Khan, Asiedu, Narayan, Mader, Subramanian,
  and Golub]{corsello2017}
Steven~M Corsello, Joshua~A Bittker, Zihan Liu, Joshua Gould, Patrick McCarren,
  Jodi~E Hirschman, Stephen~E Johnston, Anita Vrcic, Bang Wong, Mariya Khan,
  Jacob Asiedu, Rajiv Narayan, Christopher~C Mader, Aravind Subramanian, and
  Todd~R Golub.
\newblock {The Drug Repurposing Hub: a next-generation drug library and
  information resource}.
\newblock \emph{Nature Medicine}, 23\penalty0 (4):\penalty0 405--408, 2017.
\newblock \doi{10.1038/nm.4306}.

\bibitem[Dhamdhere et~al.(2019)Dhamdhere, Sundararajan, and Yan]{dhamdhere2019}
Kedar Dhamdhere, Mukund Sundararajan, and Qiqi Yan.
\newblock {How Important Is a Neuron?}
\newblock In \emph{International Conference on Learning Representations}, 2019.
\newblock URL \url{https://openreview.net/forum?id=SylKoo0cKm}.

\bibitem[Dixit et~al.(2016)Dixit, Parnas, Li, Chen, Fulco, Jerby-Arnon,
  Marjanovic, Dionne, Burks, Raychowdhury, Adamson, Norman, Lander, Weissman,
  Friedman, and Regev]{dixit2016}
Atray Dixit, Oren Parnas, Biyu Li, Jenny Chen, Charles~P. Fulco, Livnat
  Jerby-Arnon, Nemanja~D. Marjanovic, Danielle Dionne, Tyler Burks, Raktima
  Raychowdhury, Britt Adamson, Thomas~M. Norman, Eric~S. Lander, Jonathan~S.
  Weissman, Nir Friedman, and Aviv Regev.
\newblock {Perturb-Seq: Dissecting Molecular Circuits with Scalable Single-Cell
  RNA Profiling of Pooled Genetic Screens}.
\newblock \emph{Cell}, 167\penalty0 (7):\penalty0 1853--1866.e17, 2016.
\newblock \doi{10.1016/j.cell.2016.11.038}.

\bibitem[Favaro et~al.(2012)Favaro, Bensaad, Chong, Tennant, Ferguson, Snell,
  Steers, Turley, Li, G{\"u}nther, Buffa, McIntyre, and Harris]{favaro2012}
Elena Favaro, Karim Bensaad, Mei~G. Chong, Daniel~A. Tennant, David J.~P.
  Ferguson, Cameron Snell, Graham Steers, Helen Turley, Ji-Liang Li, Ulrich~L.
  G{\"u}nther, Francesca~M. Buffa, Alan McIntyre, and Adrian~L. Harris.
\newblock {Glucose utilization via glycogen phosphorylase sustains
  proliferation and prevents premature senescence in cancer cells}.
\newblock \emph{Cell Metabolism}, 16\penalty0 (6):\penalty0 751--764, 2012.
\newblock \doi{10.1016/j.cmet.2012.10.017}.

\bibitem[Frontini et~al.(2012)]{cks2012}
M~Frontini et~al.
\newblock {The CDK subunit CKS2 counteracts CKS1 to control cyclin A/CDK2
  activity in maintaining replicative fidelity and neurodevelopment}.
\newblock \emph{Developmental cell}, 2012.
\newblock \doi{10.1016/j.devcel.2012.06.018}.

\bibitem[Gentles et~al.(2015)Gentles, Newman, Liu, Bratman, Feng, Kim, Nair,
  Xu, Khuong, Hoang, Diehn, West, Plevritis, and Alizadeh]{gentles2015precog}
Andrew~J Gentles, Aaron~M Newman, Chih~Long Liu, Scott~V Bratman, Weiguo Feng,
  Dongkyoon Kim, Viswam~S Nair, Yue Xu, Amanda Khuong, Chuong~D Hoang,
  Maximilian Diehn, Robert~B West, Sylvia~K Plevritis, and Ash~A Alizadeh.
\newblock {The prognostic landscape of genes and infiltrating immune cells
  across human cancers}.
\newblock \emph{Nature Medicine}, 21\penalty0 (8):\penalty0 938--945, 2015.
\newblock \doi{10.1038/nm.3909}.

\bibitem[Gonzalez et~al.(2026)Gonzalez, Lin, Herath, Veselkov, Bronstein, and
  Zitnik]{gonzalez2026}
Guadalupe Gonzalez, Xiang Lin, Isuru Herath, Kirill Veselkov, Michael
  Bronstein, and Marinka Zitnik.
\newblock {Combinatorial prediction of therapeutic perturbations using causally
  inspired neural networks}.
\newblock \emph{Nature Biomedical Engineering}, 10\penalty0 (5):\penalty0
  1008--1025, 2026.
\newblock \doi{10.1038/s41551-025-01481-x}.
\newblock First published online September 9, 2025.

\bibitem[He et~al.(2023)He, Lyu, Li, Zhao, Qi, and Lu]{he2023}
Xin-Ling He, Wen-Yu Lyu, Xin-Yuan Li, Hong Zhao, Lu~Qi, and Jin-Jian Lu.
\newblock {Identification of glycogen phosphorylase L as a potential target for
  lung cancer}.
\newblock \emph{Medical Oncology}, 40\penalty0 (7):\penalty0 211, 2023.
\newblock \doi{10.1007/s12032-023-02069-8}.
\newblock URL \url{https://doi.org/10.1007/s12032-023-02069-8}.

\bibitem[{Human Protein Atlas}(n.d.)]{hpa_methods}
{Human Protein Atlas}.
\newblock {The Cancer resource - methods summary}.
\newblock \url{https://www.proteinatlas.org/humanproteome/cancer/method}, n.d.
\newblock Accessed September 25, 2026; the study's query uses pinned version
  25.1 annotations.

\bibitem[Huo et~al.(2026)Huo, Hou, Zhang, Duan, Lan, Li, Zhou, and
  Zhang]{huo2026}
Xingfa Huo, Helei Hou, Chuantao Zhang, Xueqin Duan, Hongwei Lan, Yufeng Li,
  Na~Zhou, and Xiaochun Zhang.
\newblock {Chicoric acid targets PYGL to normalize glycogenolysis-driven
  glycolysis to suppress non-small cell lung cancer}.
\newblock \emph{Cellular \& Molecular Biology Letters}, 31:\penalty0 105, 2026.
\newblock \doi{10.1186/s11658-026-00930-y}.

\bibitem[Ji et~al.(2017)Ji, Gao, Jia, Li, and Yu]{ji2017mps1}
Zhejian Ji, Haishan Gao, Luying Jia, Bing Li, and Hongtao Yu.
\newblock A sequential multi-target {Mps1} phosphorylation cascade promotes
  spindle checkpoint signaling.
\newblock \emph{eLife}, 6:\penalty0 e22513, 2017.
\newblock \doi{10.7554/eLife.22513}.

\bibitem[Keiser et~al.(2009)Keiser, Setola, Irwin, Laggner, Abbas, Hufeisen,
  Jensen, Kuijer, Matos, Tran, Whaley, Glennon, Hert, Thomas, Edwards,
  Shoichet, and Roth]{keiser2009}
Michael~J. Keiser, Vincent Setola, John~J. Irwin, Christian Laggner, Atheir~I.
  Abbas, Sandra~J. Hufeisen, Niels~H. Jensen, Michael~B. Kuijer, Roberto~C.
  Matos, Thuy~B. Tran, Ryan Whaley, Richard~A. Glennon, Jérôme Hert, Kelan
  L.~H. Thomas, Douglas~D. Edwards, Brian~K. Shoichet, and Bryan~L. Roth.
\newblock {Predicting new molecular targets for known drugs}.
\newblock \emph{Nature}, 462\penalty0 (7270):\penalty0 175--181, 2009.
\newblock \doi{10.1038/nature08506}.

\bibitem[Klein et~al.(2025)Klein, Fleck, Bobrovskiy, Zimmermann, Becker, Palma,
  Dony, Tejada-Lapuerta, Huguet, Lin, Azbukina, Sanch{\'i}s-Calleja, Uscidda,
  Szalata, Gander, Regev, Treutlein, Camp, and Theis]{klein2025cellflow}
Dominik Klein, Jonas~Simon Fleck, Daniil Bobrovskiy, Lea Zimmermann, S{\"o}ren
  Becker, Alessandro Palma, Leander Dony, Alejandro Tejada-Lapuerta, Guillaume
  Huguet, Hsiu-Chuan Lin, Nadezhda Azbukina, F{\'a}tima Sanch{\'i}s-Calleja,
  Theo Uscidda, Artur Szalata, Manuel Gander, Aviv Regev, Barbara Treutlein,
  J.~Gray Camp, and Fabian~J. Theis.
\newblock {CellFlow enables generative single-cell phenotype modeling with flow
  matching}.
\newblock \emph{bioRxiv}, 2025.
\newblock \doi{10.1101/2025.04.11.648220}.
\newblock URL
  \url{https://www.biorxiv.org/content/10.1101/2025.04.11.648220v1}.
\newblock Preprint, version 1.

\bibitem[Kuenzi et~al.(2020)Kuenzi, Park, Fong, Sanchez, Lee, Kreisberg, Ma,
  and Ideker]{kuenzi2020}
Brent~M Kuenzi, Jisoo Park, Samson~H Fong, Kyle~S Sanchez, John Lee, Jason~F
  Kreisberg, Jianzhu Ma, and Trey Ideker.
\newblock {Predicting Drug Response and Synergy Using a Deep Learning Model of
  Human Cancer Cells}.
\newblock \emph{Cancer Cell}, 38\penalty0 (5):\penalty0 672--684.e6, 2020.
\newblock \doi{10.1016/j.ccell.2020.09.014}.
\newblock URL \url{https://pubmed.ncbi.nlm.nih.gov/33096023/}.

\bibitem[Liu et~al.(2025)Liu, Du, Huang, Tian, Wang, Chen, and Wang]{liu2025}
Jinping Liu, Dingyu Du, Yukai Huang, Jie Tian, Xuhui Wang, Longyi Chen, and
  Feng Wang.
\newblock {CCDC86-BHLHE40-ATF3 axis promotes aerobic glycolysis and tumor
  development in glioma}.
\newblock \emph{Genes \& Diseases}, 12\penalty0 (6):\penalty0 101643, 2025.
\newblock \doi{10.1016/j.gendis.2025.101643}.
\newblock URL \url{https://doi.org/10.1016/j.gendis.2025.101643}.

\bibitem[Lotfollahi et~al.(2023)Lotfollahi, Klimovskaia~Susmelj, De~Donno,
  Hetzel, Ji, Ibarra, Srivatsan, Naghipourfar, Daza, Martin, Shendure,
  McFaline-Figueroa, Boyeau, Wolf, Yakubova, G{\"u}nnemann, Trapnell,
  Lopez-Paz, and Theis]{lotfollahi2023cpa}
Mohammad Lotfollahi, Anna Klimovskaia~Susmelj, Carlo De~Donno, Leon Hetzel,
  Yuge Ji, Ignacio~L Ibarra, Sanjay~R Srivatsan, Mohsen Naghipourfar, Riza~M
  Daza, Beth Martin, Jay Shendure, Jose~L McFaline-Figueroa, Pierre Boyeau,
  F~Alexander Wolf, Nafissa Yakubova, Stephan G{\"u}nnemann, Cole Trapnell,
  David Lopez-Paz, and Fabian~J Theis.
\newblock {Predicting cellular responses to complex perturbations in
  high-throughput screens}.
\newblock \emph{Molecular Systems Biology}, 19\penalty0 (6):\penalty0 e11517,
  2023.
\newblock \doi{10.15252/msb.202211517}.
\newblock URL \url{https://doi.org/10.15252/msb.202211517}.

\bibitem[Ma et~al.(2018)Ma, Yu, Fong, Ono, Sage, Demchak, Sharan, and
  Ideker]{ma2018}
Jianzhu Ma, Michael~Ku Yu, Samson Fong, Keiichiro Ono, Eric Sage, Barry
  Demchak, Roded Sharan, and Trey Ideker.
\newblock {Using deep learning to model the hierarchical structure and function
  of a cell}.
\newblock \emph{Nature Methods}, 15\penalty0 (4):\penalty0 290--298, 2018.
\newblock \doi{10.1038/nmeth.4627}.
\newblock URL \url{https://pubmed.ncbi.nlm.nih.gov/29505029/}.

\bibitem[McInnes et~al.(2018)McInnes, Healy, and Melville]{mcinnes2018}
Leland McInnes, John Healy, and James Melville.
\newblock {UMAP: Uniform Manifold Approximation and Projection for Dimension
  Reduction}.
\newblock \emph{arXiv preprint arXiv:1802.03426}, 2018.
\newblock URL \url{https://arxiv.org/abs/1802.03426v3}.
\newblock Version 3, revised 2020.

\bibitem[Miller et~al.(2026)]{calibrated2026}
H.~E. Miller et~al.
\newblock Deep learning perturbation models can outperform baselines on
  calibrated metrics.
\newblock \emph{Nature Biotechnology}, 2026.
\newblock \doi{10.1038/s41587-026-03307-w}.

\bibitem[M{\"u}ller et~al.(2017)M{\"u}ller, Stangner, Schmitt, Wintsche, and
  Engeland]{muller2017timing}
Gerd~A. M{\"u}ller, Konstanze Stangner, Thomas Schmitt, Axel Wintsche, and Kurt
  Engeland.
\newblock Timing of transcription during the cell cycle: Protein complexes
  binding to {E2F}, {E2F/CLE}, {CDE/CHR}, or {CHR} promoter elements define
  early and late cell cycle gene expression.
\newblock \emph{Oncotarget}, 8\penalty0 (58):\penalty0 97736--97748, 2017.
\newblock \doi{10.18632/oncotarget.10888}.

\bibitem[M{\"u}ller-Dott et~al.(2023)M{\"u}ller-Dott, Tsirvouli, Vazquez,
  Ramirez~Flores, Badia-I-Mompel, Fallegger, T{\"u}rei, L{\ae}greid, and
  Saez-Rodriguez]{mullerdott2023}
Sophia M{\"u}ller-Dott, Eirini Tsirvouli, Miguel Vazquez, Ricardo~O
  Ramirez~Flores, Pau Badia-I-Mompel, Robin Fallegger, Dénes T{\"u}rei, Astrid
  L{\ae}greid, and Julio Saez-Rodriguez.
\newblock {Expanding the coverage of regulons from high-confidence prior
  knowledge for accurate estimation of transcription factor activities}.
\newblock \emph{Nucleic Acids Research}, 51\penalty0 (20):\penalty0
  10934--10949, 2023.
\newblock \doi{10.1093/nar/gkad841}.
\newblock URL \url{https://doi.org/10.1093/nar/gkad841}.

\bibitem[Pagliuca et~al.(2014)Pagliuca, Millman, G{\"u}rtler, Segel,
  Van~Dervort, Ryu, Peterson, Greiner, and Melton]{pagliuca2014beta}
Felicia~W. Pagliuca, Jeffrey~R. Millman, Mads G{\"u}rtler, Michael Segel, Alana
  Van~Dervort, Jennifer~Hyoje Ryu, Quinn~P. Peterson, Dale Greiner, and
  Douglas~A. Melton.
\newblock Generation of functional human pancreatic {\ensuremath{\beta}} cells
  in vitro.
\newblock \emph{Cell}, 159\penalty0 (2):\penalty0 428--439, 2014.
\newblock \doi{10.1016/j.cell.2014.09.040}.

\bibitem[Replogle et~al.(2022)Replogle, Saunders, Pogson, Hussmann, Lenail,
  Guna, Mascibroda, Wagner, Adelman, Lithwick-Yanai, Iremadze, Oberstrass,
  Lipson, Bonnar, Jost, Norman, and Weissman]{replogle2022}
Joseph~M. Replogle, Reuben~A. Saunders, Angela~N. Pogson, Jeffrey~A. Hussmann,
  Alexander Lenail, Alina Guna, Lauren Mascibroda, Eric~J. Wagner, Karen
  Adelman, Gila Lithwick-Yanai, Nika Iremadze, Florian Oberstrass, Doron
  Lipson, Jessica~L. Bonnar, Marco Jost, Thomas~M. Norman, and Jonathan~S.
  Weissman.
\newblock {Mapping information-rich genotype-phenotype landscapes with
  genome-scale Perturb-seq}.
\newblock \emph{Cell}, 185\penalty0 (14):\penalty0 2559--2575.e28, 2022.
\newblock \doi{10.1016/j.cell.2022.05.013}.

\bibitem[Roohani et~al.(2024)Roohani, Huang, and Leskovec]{roohani2024gears}
Yusuf Roohani, Kexin Huang, and Jure Leskovec.
\newblock {Predicting transcriptional outcomes of novel multigene perturbations
  with GEARS}.
\newblock \emph{Nature Biotechnology}, 42\penalty0 (6):\penalty0 927--935,
  2024.
\newblock \doi{10.1038/s41587-023-01905-6}.
\newblock First published online August 17, 2023.

\bibitem[Schlichtkrull et~al.(2018)Schlichtkrull, Kipf, Bloem, van~den Berg,
  Titov, and Welling]{schlichtkrull2018}
Michael Schlichtkrull, Thomas~N. Kipf, Peter Bloem, Rianne van~den Berg, Ivan
  Titov, and Max Welling.
\newblock {Modeling Relational Data with Graph Convolutional Networks}.
\newblock In \emph{The Semantic Web: ESWC 2018}, pp.\  593--607. Springer,
  2018.
\newblock \doi{10.1007/978-3-319-93417-4_38}.
\newblock URL \url{https://doi.org/10.1007/978-3-319-93417-4_38}.

\bibitem[Schmidt et~al.(2007)]{sorla2007}
Vanessa Schmidt et~al.
\newblock {SorLA/LR11} regulates processing of amyloid precursor protein via
  interaction with adaptors {GGA} and {PACS-1}.
\newblock \emph{Journal of Biological Chemistry}, 2007.
\newblock \doi{10.1074/jbc.M705073200}.

\bibitem[Schnake et~al.(2022)Schnake, Eberle, Lederer, Nakajima, Sch{\"u}tt,
  M{\"u}ller, and Montavon]{schnake2020}
Thomas Schnake, Oliver Eberle, Jonas Lederer, Shinichi Nakajima, Kristof~T.
  Sch{\"u}tt, Klaus-Robert M{\"u}ller, and Gr{\'e}goire Montavon.
\newblock {Higher-Order Explanations of Graph Neural Networks via Relevant
  Walks}.
\newblock \emph{IEEE Transactions on Pattern Analysis and Machine
  Intelligence}, 44\penalty0 (11):\penalty0 7581--7596, 2022.
\newblock \doi{10.1109/TPAMI.2021.3115452}.
\newblock URL \url{https://ieeexplore.ieee.org/document/9547794/}.

\bibitem[Singharajkomron et~al.(2026)]{camsap2026}
Natsaranyatron Singharajkomron et~al.
\newblock {miR-145-5p/CAMSAP2} axis modulates cisplatin sensitivity in
  non-small cell lung cancer.
\newblock \emph{Cancer Cell International}, 2026.
\newblock \doi{10.1186/s12935-026-04209-2}.

\bibitem[Stamatiou et~al.(2023)Stamatiou, Chmielewska, Ohta, Earnshaw, and
  Vagnarelli]{stamatiou2023}
Konstantinos Stamatiou, Aldona Chmielewska, Shinya Ohta, William~C. Earnshaw,
  and Paola Vagnarelli.
\newblock {CCDC86 is a novel Ki-67-interacting protein important for cell
  division}.
\newblock \emph{Journal of Cell Science}, 136\penalty0 (2):\penalty0 jcs260391,
  2023.
\newblock \doi{10.1242/jcs.260391}.
\newblock URL \url{https://doi.org/10.1242/jcs.260391}.

\bibitem[Su et~al.(2026)Su, He, You, Jiang, Zhou, Zhang, Wang, Su, Tolstoy,
  Chang, Lu, and Yuan]{su2026}
Jin Su, Yan He, Shiyang You, Shiyu Jiang, Xibin Zhou, Xuting Zhang, Yuxuan
  Wang, Xining Su, Igor Tolstoy, Xing Chang, Hongyuan Lu, and Fajie Yuan.
\newblock {A trimodal protein language model enables advanced protein
  searches}.
\newblock \emph{Nature Biotechnology}, 44\penalty0 (8):\penalty0 1310--1316,
  2026.
\newblock \doi{10.1038/s41587-025-02836-0}.
\newblock First published online October 2, 2025.

\bibitem[Subramanian et~al.(2017)Subramanian, Narayan, Corsello, Peck, Natoli,
  Lu, Gould, Davis, Tubelli, Asiedu, Lahr, Hirschman, Liu, Donahue, Julian,
  Khan, Wadden, Smith, Lam, Liberzon, Toder, Bagul, Orzechowski, Enache,
  Piccioni, Johnson, Lyons, Berger, Shamji, Brooks, Vrcic, Flynn, Rosains,
  Takeda, Hu, Davison, Lamb, Ardlie, Hogstrom, Greenside, Gray, Clemons,
  Silver, Wu, Zhao, Read-Button, Wu, Haggarty, Ronco, Boehm, Schreiber, Doench,
  Bittker, Root, Wong, and Golub]{subramanian2017}
Aravind Subramanian, Rajiv Narayan, Steven~M Corsello, David~D Peck, Ted~E
  Natoli, Xiaodong Lu, Joshua Gould, John~F Davis, Andrew~A Tubelli, Jacob~K
  Asiedu, David~L Lahr, Jodi~E Hirschman, Zihan Liu, Melanie Donahue, Bina
  Julian, Mariya Khan, David Wadden, Ian~C Smith, Daniel Lam, Arthur Liberzon,
  Courtney Toder, Mukta Bagul, Marek Orzechowski, Oana~M Enache, Federica
  Piccioni, Sarah~A Johnson, Nicholas~J Lyons, Alice~H Berger, Alykhan~F
  Shamji, Angela~N Brooks, Anita Vrcic, Corey Flynn, Jacqueline Rosains,
  David~Y Takeda, Roger Hu, Desiree Davison, Justin Lamb, Kristin Ardlie,
  Larson Hogstrom, Peyton Greenside, Nathanael~S Gray, Paul~A Clemons, Serena
  Silver, Xiaoyun Wu, Wen-Ning Zhao, Willis Read-Button, Xiaohua Wu, Stephen~J
  Haggarty, Lucienne~V Ronco, Jesse~S Boehm, Stuart~L Schreiber, John~G Doench,
  Joshua~A Bittker, David~E Root, Bang Wong, and Todd~R Golub.
\newblock {A Next Generation Connectivity Map: L1000 Platform and the First
  1,000,000 Profiles}.
\newblock \emph{Cell}, 171\penalty0 (6):\penalty0 1437--1452.e17, 2017.
\newblock \doi{10.1016/j.cell.2017.10.049}.
\newblock URL \url{https://doi.org/10.1016/j.cell.2017.10.049}.

\bibitem[Sudakin et~al.(2001)Sudakin, Chan, and Yen]{sudakin2001mcc}
Valery Sudakin, Gordon K.~T. Chan, and Tim~J. Yen.
\newblock Checkpoint inhibition of the {APC/C} in {HeLa} cells is mediated by a
  complex of {BUBR1}, {BUB3}, {CDC20}, and {MAD2}.
\newblock \emph{Journal of Cell Biology}, 154\penalty0 (5):\penalty0 925--936,
  2001.
\newblock \doi{10.1083/jcb.200102093}.

\bibitem[Sundararajan et~al.(2017)Sundararajan, Taly, and
  Yan]{sundararajan2017}
Mukund Sundararajan, Ankur Taly, and Qiqi Yan.
\newblock {Axiomatic Attribution for Deep Networks}.
\newblock In \emph{Proceedings of the 34th International Conference on Machine
  Learning}, volume~70, pp.\  3319--3328. PMLR, 2017.
\newblock URL \url{https://proceedings.mlr.press/v70/sundararajan17a.html}.

\bibitem[Sykes et~al.(2016)Sykes, Kfoury, Mercier,
  et~al.]{sykes2016differentiation}
David~B. Sykes, Youmna~S. Kfoury, Fran{\c{c}}ois~E. Mercier, et~al.
\newblock Inhibition of dihydroorotate dehydrogenase overcomes differentiation
  blockade in acute myeloid leukemia.
\newblock \emph{Cell}, 167\penalty0 (1):\penalty0 171--186.e15, 2016.
\newblock \doi{10.1016/j.cell.2016.08.057}.

\bibitem[Szklarczyk et~al.(2025)Szklarczyk, Nastou, Koutrouli, Kirsch,
  Mehryary, Hachilif, Hu, Peluso, Huang, Fang, Doncheva, Pyysalo, Bork, Jensen,
  and von Mering]{szklarczyk2025}
Damian Szklarczyk, Katerina Nastou, Mikaela Koutrouli, Rebecca Kirsch, Farrokh
  Mehryary, Radja Hachilif, Dewei Hu, Matteo~E Peluso, Qingyao Huang, Tao Fang,
  Nadezhda~T Doncheva, Sampo Pyysalo, Peer Bork, Lars~J Jensen, and Christian
  von Mering.
\newblock {The STRING database in 2025: protein networks with directionality of
  regulation}.
\newblock \emph{Nucleic Acids Research}, 53\penalty0 (D1):\penalty0 D730--D737,
  2025.
\newblock \doi{10.1093/nar/gkae1113}.
\newblock URL \url{https://doi.org/10.1093/nar/gkae1113}.

\bibitem[Tsai et~al.(2020)Tsai, Wu, Chang, Hung, Chang, Chang, Jian, Tsai,
  Huang, Chong, and Hsu]{tsai2020ttk}
Ying-Ming Tsai, Kuan-Li Wu, Yung-Yun Chang, Jen-Yu Hung, Wei-An Chang,
  Chao-Yuan Chang, Shu-Fang Jian, Pei-Hsun Tsai, Yung-Chi Huang, Inn-Wen Chong,
  and Ya-Ling Hsu.
\newblock Upregulation of {Thr/Tyr} kinase increases the cancer progression by
  neurotensin and dihydropyrimidinase-like 3 in lung cancer.
\newblock \emph{International Journal of Molecular Sciences}, 21\penalty0
  (5):\penalty0 1640, 2020.
\newblock \doi{10.3390/ijms21051640}.

\bibitem[Uhlen et~al.(2017)Uhlen, Zhang, Lee, Sj{\"o}stedt, Fagerberg,
  Bidkhori, Benfeitas, Arif, Liu, Edfors, Sanli, von Feilitzen, Oksvold,
  Lundberg, Hober, Nilsson, Mattsson, Schwenk, Brunnstr{\"o}m, Glimelius,
  Sj{\"o}blom, Edqvist, Djureinovic, Micke, Lindskog, Mardinoglu, and
  Ponten]{uhlen2017}
Mathias Uhlen, Cheng Zhang, Sunjae Lee, Evelina Sj{\"o}stedt, Linn Fagerberg,
  Gholamreza Bidkhori, Rui Benfeitas, Muhammad Arif, Zhengtao Liu, Fredrik
  Edfors, Kemal Sanli, Kalle von Feilitzen, Per Oksvold, Emma Lundberg, Sophia
  Hober, Peter Nilsson, Johanna Mattsson, Jochen~M Schwenk, Hans
  Brunnstr{\"o}m, Bengt Glimelius, Tobias Sj{\"o}blom, Per-Henrik Edqvist,
  Dijana Djureinovic, Patrick Micke, Cecilia Lindskog, Adil Mardinoglu, and
  Fredrik Ponten.
\newblock {A pathology atlas of the human cancer transcriptome}.
\newblock \emph{Science}, 357\penalty0 (6352):\penalty0 eaan2507, 2017.
\newblock \doi{10.1126/science.aan2507}.
\newblock URL \url{https://doi.org/10.1126/science.aan2507}.

\bibitem[{\"U}nsal-Ka{\c{c}}maz et~al.(2005){\"U}nsal-Ka{\c{c}}maz, Mullen,
  Kaufmann, and Sancar]{timeless2005}
Keziban {\"U}nsal-Ka{\c{c}}maz, Thomas~E. Mullen, William~K. Kaufmann, and Aziz
  Sancar.
\newblock Coupling of human circadian and cell cycles by the timeless protein.
\newblock \emph{Molecular and Cellular Biology}, 25\penalty0 (8):\penalty0
  3109--3116, 2005.
\newblock \doi{10.1128/MCB.25.8.3109-3116.2005}.

\bibitem[Vassilevska et~al.(2009)Vassilevska, Williams, and
  Yuster]{vassilevska2009}
Virginia Vassilevska, R.~Ryan Williams, and Raphael Yuster.
\newblock {All Pairs Bottleneck Paths and Max-Min Matrix Products in Truly
  Subcubic Time}.
\newblock \emph{Theory of Computing}, 5\penalty0 (9):\penalty0 173--189, 2009.
\newblock \doi{10.4086/toc.2009.v005a009}.
\newblock URL \url{https://toc.cs.uchicago.edu/articles/v005a009/}.

\bibitem[von Schubert et~al.(2015)von Schubert, Cubizolles, Bracher,
  Sliedrecht, Kops, and Nigg]{vonschubert2015plk1}
Conrad von Schubert, Fabien Cubizolles, Jasmine~M. Bracher, Tale Sliedrecht,
  Geert J. P.~L. Kops, and Erich~A. Nigg.
\newblock {Plk1} and {Mps1} cooperatively regulate the spindle assembly
  checkpoint in human cells.
\newblock \emph{Cell Reports}, 12\penalty0 (1):\penalty0 66--78, 2015.
\newblock \doi{10.1016/j.celrep.2015.06.007}.

\bibitem[Wu et~al.(2025)Wu, Padia, Murphy, Barzilay, and Jaakkola]{wu2025}
Menghua Wu, Umesh Padia, Sean~H. Murphy, Regina Barzilay, and Tommi Jaakkola.
\newblock Identifying biological perturbation targets through causal
  differential networks.
\newblock In Aarti Singh, Maryam Fazel, Daniel Hsu, Simon Lacoste-Julien, Felix
  Berkenkamp, Tegan Maharaj, Kiri Wagstaff, and Jerry Zhu (eds.),
  \emph{Proceedings of the 42nd International Conference on Machine Learning},
  volume 267 of \emph{Proceedings of Machine Learning Research}, pp.\
  67537--67561. PMLR, 13--19 Jul 2025.
\newblock URL \url{https://proceedings.mlr.press/v267/wu25v.html}.

\bibitem[Ying et~al.(2019)Ying, Bourgeois, You, Zitnik, and Leskovec]{ying2019}
Zhitao Ying, Dylan Bourgeois, Jiaxuan You, Marinka Zitnik, and Jure Leskovec.
\newblock {GNNExplainer: Generating Explanations for Graph Neural Networks}.
\newblock In \emph{Advances in Neural Information Processing Systems},
  volume~32, 2019.
\newblock URL
  \url{https://proceedings.neurips.cc/paper/2019/hash/d80b7040b773199015de6d3b4293c8ff-Abstract.html}.

\bibitem[Yuan et~al.(2025)Yuan, Zhang, Von~Feilitzen, Zwahlen, Shi, Li, Yang,
  Song, Turkez, Uhlén, and Mardinoglu]{hpa_validation}
Meng Yuan, Cheng Zhang, Kalle Von~Feilitzen, Martin Zwahlen, Mengnan Shi,
  Xiangyu Li, Hong Yang, Xiya Song, Hasan Turkez, Mathias Uhlén, and Adil
  Mardinoglu.
\newblock {The Human Pathology Atlas for deciphering the prognostic features of
  human cancers}.
\newblock \emph{eBioMedicine}, 111:\penalty0 105495, 2025.
\newblock \doi{10.1016/j.ebiom.2024.105495}.
\newblock URL \url{https://doi.org/10.1016/j.ebiom.2024.105495}.
\newblock Published online December 10, 2024.

\bibitem[Zheng et~al.(2019)Zheng, Chen, Kawakami, Chen, Roszik, Mustachio,
  Kurie, Villalobos, Lu, Behrens, Mino, Solis, Silvester, Thu, Cescon,
  Rodriguez-Canales, Wistuba, Mak, Liu, and Dmitrovsky]{zheng2019ttk}
Lin Zheng, Zibo Chen, Masanori Kawakami, Yulong Chen, Jason Roszik, Lisa~Maria
  Mustachio, Jonathan~M. Kurie, Pamela Villalobos, Wei Lu, Carmen Behrens,
  Barbara Mino, Luisa~M. Solis, Jennifer Silvester, Kelsie~L. Thu, David~W
  Cescon, Jaime Rodriguez-Canales, Ignacio~I. Wistuba, Tak~W. Mak, Xi~Liu, and
  Ethan Dmitrovsky.
\newblock {Tyrosine Threonine Kinase Inhibition Eliminates Lung Cancers by
  Augmenting Apoptosis and Polyploidy}.
\newblock \emph{Molecular Cancer Therapeutics}, 18\penalty0 (10):\penalty0
  1775--1786, 2019.
\newblock \doi{10.1158/1535-7163.MCT-18-0864}.

\end{thebibliography}
\bibliographystyle{iclr2027_conference}

\clearpage
\appendix
\section{Data, graph, and variable definitions}\label{app:data}
\subsection{Assay and retention}
The pinned inputs are CLUE's LINCS2020 beta Level-3 CRISPR-treated and control GCTX matrices, sample metadata, and gene annotations. L1000 measures expression abundance rather than transcription rates. The selected \texttt{trt\_xpr} modality denotes CRISPR/Cas9 knockout, not CRISPR interference. Level-3 values are processed expression, not raw counts or Level-5 differential signatures. All retained treated wells are measured at 96 hours; dose is missing and is not interpreted as zero.

Retention requires physical replicate X1, X2, or X3, a nonmissing target label, and matched controls on the same plate. The target vocabulary is the intersection of labels present in X1/X2 and X3 before hybrid splitting. This produces \RetainedPairs{} paired wells on \Plates{} plates, with \Controls{} matched controls. Of \TargetVocabulary{} retained target identities, \EligibleTargets{} map to eligible positive labels, yielding \EligiblePairs{} supervised pairs. Protein-covered candidate nodes form a broader set; a never-positive candidate receives competition-based negative supervision but is not an established biological negative or a validated zero-shot prediction.

LINCS supplies expression estimates inferred from the \Landmarks{} measured landmarks; we use these supplied values without fitting a new inference model. The BING (Best Inferred Genes) space combines landmarks with genes whose inferred expression passed the provider's validation against RNA-seq \citep{subramanian2017}. In the pinned LINCS annotation, 10,174 genes belong to BING and 10,086 of these are annotated protein-coding. The historical identifier intersection retains \Features{} coordinates: all \Landmarks{} landmarks and \Inferred{} inferred genes. This ordered feature table is preserved as a contract; no GTEx expression enters this study. Extending expression coverage lets more gene nodes combine cellular context with protein features and typed relationships. Inferred coordinates remain dependent estimates, and we make no claim of exact variance preservation. Native control values are averaged with float64 accumulation and stored in float32 before normalization.

\subsection{Training-only coordinates and split semantics}
For rotation $f$, eligible inner-training treated wells $T_f$ determine the per-gene moments for both input channels:
\begin{equation}
\mu_{fj}=\frac1{|T_f|}\sum_{n\in T_f}v_{nj},\qquad
\sigma_{fj}^2=\frac1{|T_f|}\sum_{n\in T_f}(v_{nj}-\mu_{fj})^2,\qquad
x_j^{(c)}=\frac{v_j^{(c)}-\mu_{fj}}{s_{fj}}.
\end{equation}
Here $s_{fj}=1$ if $\sigma_{fj}<10^{-6}$ and $s_{fj}=\sigma_{fj}$ otherwise. Moments use population variance, float64 accumulation, and float32 storage. Controls, validation, test, and constructed requests never fit these moments. There is no extra log transform, clipping, rank transform, or per-profile centering.

H1/H2/H3 are hybrid split rotations, not X1/X2/X3 physical replicates or independent random seeds. Multi-line tissues hold out an entire cell line, including every plate and replicate. Singleton tissues hold out one complete physical replicate while other replicates of that line remain in development. Within each development line, a seeded 10\% of whole plates is assigned to validation, rounding up while retaining a training plate. Validation selects the minimum unmasked CE and does not include the outer test.

H1 and H3 each contain 30,921 A549 test rows and no A549 training or validation rows. H2 includes A549 in development and is not a primary consensus member. Repeated holdouts of the same line are not independent biological units. Neither evaluation setting holds out target genes.

\subsection{Baseline quality}
The query baseline is the native matched-control mean for the lexicographically first retained A549 sample whose plate's matched controls all carry \texttt{qc\_pass=1} in the CLUE metadata: \texttt{XPR008\_A549.311\_96H\_X3\_B35}, with 20 controls. Failed and unlabeled controls both disqualify a plate, and the rule uses QC metadata alone, so it is fixed before any model is queried. It is a deterministic selection, not an estimate of a typical cell.

Training retention records but does not filter technical QC. Raw metadata retain separate failed, passing, and missing labels. CLUE cautions against assuming that all downloaded L1000 columns pass QC \citep{clueqc}; its CRISPR documentation also distinguishes technical quality from functional perturbation checks \citep{cluecrisprqc}. The baseline's conductance is an explanation of that model input, not evidence that it represents a typical A549 state. Appendix~\ref{app:qc-sensitivity} reports robustness across all QC-passing A549 plates.

\subsection{Graph construction and protein inputs}
The ordered expression features and resolved retained perturbation targets span 11,791 distinct genes. The unresolved historical label RIEG2 is also retained for audit, giving 11,792 seed nodes. Adding 5,516 one-hop neighbors across STRING v12.5 and CollecTRI yields \Nodes{} nodes; we then retain the selected relations within that universe. Resolved nodes use canonical HGNC symbols; all nodes follow lexicographic symbol order. STRING pairs retain their maximum confidence after mapping; a threshold of 700 removes lower-confidence pairs, and STRING self-pairs are excluded. The evidence-filtered CollecTRI subset requires a PMID-decided sign and at least two distinct underlying resources after normalizing resource suffixes and excluding the umbrella label. Source signed self-edges are retained.

Five biological families become eight transport channels: STRING functional and physical expand bidirectionally within their channels; STRING regulatory, CollecTRI activation, and CollecTRI repression each have separate forward and reverse channels. Collapsing relationship types and reverse copies gives 286,303 unique unordered nonself gene pairs across all \Nodes{} nodes, with 33.1 distinct neighbors per node on average (median 15). Counting transport channels separately gives 900,896 nonself directed typed edges, or 52.1 incoming messages per node per message-passing transition on average; this count excludes self-edges and residual connections. Their learned transformations are not constrained to positive or negative latent effects. STRING confidence determines retention, not the runtime message coefficient. For sender $j$ and receiver $i$,
\begin{equation}
\widetilde A^r_{ij}=\frac{\one[(j\to i)\in E_r]}{
\sqrt{\max(1,d^r_{\mathrm{in},i})\max(1,d^r_{\mathrm{out},j})}}.
\end{equation}
All relation channels read the same preceding state and are summed before SiLU. No additional self-loops are introduced; the residual provides a direct carried-state path.

Protein vectors are reconstructed from the pinned published SwissProt ProTrek-650M FAISS index; setup does not run a new sequence encoder. HGNC UniProt mappings, with STRING accession fallback, provide vectors for \Candidates{} nodes. Candidate eligibility follows this availability mask, rather than an HGNC protein-coding filter. The 83 nodes without mapped vectors remain computational context but are excluded from ranking; five are annotated protein-coding in the pinned HGNC table. The shared projector maps 1,024 dimensions through 256 to 128 with SiLU between affine maps. Baseline and endpoint encoders are separate scalar-to-128 affine maps. Missing protein or expression suppresses that component, including expression-encoder biases when the input is unavailable. Of the \Features{} expression-bearing nodes, 10,071 also have protein vectors.

The official September 4, 2026 HGNC archive replaced an unavailable August 28 source during acquisition. The recorded reconstruction reproduced seven ordered graph tables and both protein arrays byte-for-byte; source metadata retain the two source identities and the equivalence record. This is acquisition provenance, not a change of graph or scientific interpretation.

\begin{table}[t]
\centering
\caption{Notation shared by the model and attribution. Node states are indexed from 0 through $L$; transition $\ell$ connects states $\ell$ and $\ell+1$.}
\begin{tabularx}{\linewidth}{lX}
\toprule
Symbol & Meaning \\
\midrule
$N,F,|\mathcal V_p|,d$ & \Nodes{} graph nodes, \Features{} inputs, \Candidates{} protein-covered candidates, hidden width \HiddenWidth. \\
$q=(x^{(0)},x^{(1)})$ & Ordered baseline/endpoint pair, shape $[2,F]$. \\
$m_i^p,m_i^x,a_i$ & Protein/expression availability and paired training keep mask. \\
$p_i,r_i(q),\bar p$ & Projected protein, expression encoding, and projected protein-population mean. \\
$h_i^{(\ell)},H^{(\ell)}$ & A node vector in $\R^d$ and the all-node matrix in $\R^{N\times d}$. \\
$m_e^{(\ell)},M_i^{(\ell)},u_i^{(\ell)}$ & Individual typed message, receiver message sum, and pre-LayerNorm carried-state-plus-message sum. \\
$s_k,\Delta s_k$ & Nominated raw logit and score difference along the chosen attribution path. \\
$C_i^{(\ell)},E_e^{(\ell)},R_i^{(\ell)}$ & Signed node, message, and residual conductance, in nominated-logit units. \\
\bottomrule
\end{tabularx}
\end{table}

\section{Training and predictive characterization}\label{app:training}
The shared decoder is affine $128\to64$, SiLU, affine $64\to1$. Its eligible outputs compete in one global softmax. For correct target $y$ and the highest-scoring $K_{\mathrm{eff}}=\min(\HardNegatives,|\mathcal V_p|-1)$ incorrect candidates $\mathcal H_K$, training uses
\begin{equation}
\mathcal L(q,y)=-s_y+\log\sum_{i\in\mathcal V_p}e^{s_i}
+\frac{\lambda}{K_{\mathrm{eff}}}\sum_{j\in\mathcal H_K}[m+s_j-s_y]_+,
\qquad\lambda=\RankingWeight,\quad m=\RankingMargin.
\end{equation}
Only the auxiliary margin is restricted to hard negatives. Cross-entropy compares all candidates for every pair. The input graph and protein vectors are frozen; the projector, expression encoders, relation transforms, LayerNorm parameters, and decoder are learned. Expression encoders initialize at zero but remain trainable. A Bernoulli keep mask of probability 0.85 hides each available gene's two expression channels together, without inverted-dropout rescaling. Evaluation uses no random mask.

AdamW uses learning rate \LearningRate{}, weight decay \WeightDecay{}, and gradient-norm clipping at 5, without a scheduler or mixed precision in the declared path. The budget is \EpochBudget{} epochs with patience \Patience{} and minimum CE improvement zero; training and validation use seed \TrainingSeed{}. Four CUDA ranks each process a nominal \TrainBatch{} pairs. Balanced, disjoint, unpadded shards include every training row; local loss sums are scaled by world size over global example count before DDP gradient averaging, including uneven final shards. Validation CE selects the checkpoint, after which full outer tests are evaluated.

\begin{table}[t]
\centering
\caption{Predictive context for the interpreted models. Equal-tissue means keep five unseen-line tissues and six held-replicate tissues separate. Differences are descriptive: these overlapping rotations are not independent seeds, and the study does not isolate depth from capacity or nonlinear computation.}
\begin{tabular}{crrrr}
\toprule
Depth & Unseen MRR & CE & Replicate MRR & CE \\
\midrule
1 & 0.1463 & 7.3530 & 0.1213 & 7.4106 \\
2 & 0.1414 & 7.4310 & 0.1296 & 7.1058 \\
3 & 0.1408 & 7.4816 & 0.1325 & 6.9100 \\
4 & 0.1456 & 7.5455 & 0.1352 & 6.7383 \\
\bottomrule
\end{tabular}

\label{tab:depth}
\end{table}

Across twelve fits, one-to-four-hop held-replicate MRR rises from \OneReplicateMRR{} to \FourReplicateMRR{}, while unseen-line MRR is similar (\OneUnseenMRR{} versus \FourUnseenMRR{}) and CE worsens. Depth changes capacity and reach; no uniformly preferred depth is established.

The twelve fits share the declared graph, ordered features, candidate policy, and non-depth training recipe; within each rotation their split and normalizer identities match. This comparison characterizes depth within the GNN; the matched architecture comparison is reported in Appendix~\ref{app:controls}.

For target rank $r_a$ and target softmax probability $p_a(y_a)$, MRR is $n^{-1}\sum_a r_a^{-1}$ and CE is $-n^{-1}\sum_a\log p_a(y_a)$. A uniformly random ordering of the $|\mathcal V_p|=\Candidates$ candidates has expected MRR $|\mathcal V_p|^{-1}\sum_{r=1}^{|\mathcal V_p|}r^{-1}\approx0.00060$. The held-out rank convention gives ties their average occupied rank: $r_a=1+g_a+(e_a-1)/2$, with $g_a$ strictly greater scores and $e_a$ equal scores including the target. Hits uses the recorded \texttt{torch.topk} convention. The nomination table separately uses minimum tied rank; a fixed node order resolves displayed ties.

Group means are pooled by example count within a fold/evaluation unit. Repeated holdouts of a unit are averaged, then distinct units receive equal weight within tissue and tissues receive equal weight within a setting:
\begin{equation}
m_{fu}=\frac{\sum_{g\in(f,u)}n_gm_g}{\sum_{g\in(f,u)}n_g},\quad
m_u=\frac1{|F_u|}\sum_fm_{fu},\quad
m_t=\frac1{|U_t|}\sum_um_u,\quad
m_{\rm setting}=\frac1{|T|}\sum_tm_t.
\end{equation}
For unseen-line tests, physical replicates are pooled within a line/fold before repeated line holdouts are averaged. For held-replicate tests, the unit is a line/physical-replicate pair. This weights tissues deliberately rather than estimating the average recorded well. Repeated rotations do not support independence-based confidence intervals. Group medians cannot recover a pooled median, and MRR cannot recover mean rank.

\section{The request and its two geometries}\label{app:request}
\subsection{Prognostic evidence and coordinates}
HPA separates high- and low-expression patient groups at an expression cutoff and assesses survival association. For the primary A549 request, we use the validated tier in the pinned 25.1 table, which identifies prognostic associations replicated across cohorts \citep{hpa_validation,hpa_methods}. Its entries are P values, not expression amplitudes or causal effect sizes. We use only validated favorable/unfavorable entries for \emph{Lung Adenocarcinoma (validation)}, deduplicate by Ensembl identity, and exclude conflicting directions or ambiguous mappings. Graph-only markers cannot be edited expression inputs. The resulting primary request has \RequestedTotal{} mapped coordinates, not the whole raw HPA marker list. The other cell contexts and their evidence tiers are described in Appendix~\ref{app:query-audit}.

Let $c^{\rm nat}$ be the exact matched native control mean and $(\mu^R,s^R)$ the H1 reference moments. The request is
\begin{align}
c^R&=(c^{\rm nat}-\mu^R)\oslash s^R, &
t^R&=c^R+a\one_{S_+}-a\one_{S_-},\qquad a=\RequestAmplitude,\\
q^{\rm nat}&=(q^R\odot s^R)+\mu^R, &
q_f&=(q^{\rm nat}-\mu^f)\oslash s^f,\qquad f\in\{\mathrm{H1,H3}\}.
\end{align}
The reference pair is formed in float32, then transferred through native units in float64. This retains the declared physical request and its rounding. All other coordinates remain unchanged; there is no clipping or profile-wide renormalization. H1 defines the edit units, each member's normalizer defines scoring coordinates, and the projected protein mean defines the separate attribution reference. The original PYGL and CCDC86 nominees both have own-coordinate decreases; TTK does not.

The primary fits are sibling rotations of the specified four-hop run, not a search over the newest or best available checkpoint. Model membership is selected from saved train/validation/test exposure. The explorer labels contexts without whole-line-held-out models as seen-cell exploration rather than presenting them as held out. A missing HPA cohort or evidence tier is not silently replaced by another policy.

\subsection{Observed response geometry}
One observed-displacement point is a retained knockout well. It uses the complete vector $(v_n^{\rm endpoint}-\bar v^{\rm plate(n)})\oslash s^{\rm H1}$, retaining replicate and partition membership without unit normalization. The exploratory UMAP uses all retained wells and is not part of predictor fitting or checkpoint selection. A second view uses the same coordinates, colored by the within-tissue rank of full-feature cosine alignment with favorable-up/unfavorable-down directions. That descriptive overview uses explicitly matched TCGA cohorts and potential-plus-validated markers, separately from the primary validation-only A549 request. Unmatched cohorts or undefined cosines are unscored, not imputed.

\subsection{Learned gene-state geometry}
The representation view instead contains one row per gene at each stage of the same frozen H1 model and request. One CPU UMAP fit uses all \GeometryRows{} unscaled node-stage vectors, Euclidean distance, 30 neighbors, minimum distance 0.15, 500 epochs, random initialization, seed \TrainingSeed{}, and one job. Neither logits nor ranks enter the fit. No stage-specific centering or normalization is performed.

\section{Conductance calculations and numerical evidence}\label{app:numerics}
\textbf{Protein-reference path.} The request stays fixed, and covered genes' projected proteins move from their common mean $\bar p=|\mathcal V_p|^{-1}\sum_{i\in\mathcal V_p}p_i$ to their actual values, $p_i(\alpha)=(1-\alpha)\bar p+\alpha p_i$. The mean is taken after the learned projector; all covered proteins move jointly. Because weights are shared, $\Delta s_k$ on this path, written $\Delta s_k^{\rm protein}$, measures how much actual protein identities distinguish the nominee in the requested context; expression and graph position can still distinguish genes, so the reference is not a random guess. The two paths answer different questions, so their totals and signs are not interchanged.

\subsection{Initial features and complete computational layers}
The formulas below first specify the original fixed-expression protein path; the request path substitutes expression-input tangents as defined in Appendix~\ref{app:query-audit}. At each integration point, the same learned updates in Equation~\ref{eq:update} act on the interpolated initial state. The projected protein endpoints and mean are detached, fixed values; all trained parameters and graph coefficients are unchanged. For a covered node, its initial tangent is $p_i-\bar p$; for a missing-protein node it is zero. Initial feature attribution is
\begin{equation}
I_{i,a}=(p_{i,a}-\bar p_a)\int_0^1
\left.\frac{\partial s_k}{\partial h_{i,a}^{(0)}}\right|_\alpha d\alpha,
\qquad C_i^{(0)}=\sum_a I_{i,a},
\end{equation}
with $I_{i,a}=0$ where protein is missing. A node can have zero own initial protein attribution yet carry nonzero later conductance through incoming messages.

The node-local decoder makes the nominee's final state the only directly read final node. Its conductance therefore equals the full difference at that layer, up to numerical integration error. Earlier layers can distribute the same difference across many nodes. This architectural endpoint is explicitly separated in the profile heatmaps; it is not evidence that earlier computation lacked network dependence.

\subsection{Full forward path, reverse sensitivities, and forward tangents}
For each quadrature position, \texttt{stages\_from\_initial} evaluates the actual full graph, recording node states, summed messages, and pre-LayerNorm states. One backward call obtains the nominated score's sensitivities to those tensors. Tangents propagate forward through the same transforms. For edge $e=(j\to i,r)$,
\begin{equation}
\dot m_e^{(\ell)}=\widetilde A^r_{ij}W_{\ell,r}\dot h_j^{(\ell)},\qquad
\dot u_i^{(\ell)}=\dot h_i^{(\ell)}+\silu'(M_i^{(\ell)})\odot\dot M_i^{(\ell)},\qquad
\dot h_i^{(\ell+1)}=D\LN_\ell(u_i^{(\ell)})[\dot u_i^{(\ell)}].
\end{equation}
A dot denotes differentiation with respect to $\alpha$. The LayerNorm directional derivative includes centering, variance, epsilon, and learned scale. It is not an interpolation between independently chosen hidden-layer endpoints.

Message conductance contracts each transformed sender tangent with the receiver's message-sum sensitivity and fixed normalized coefficient. Contractions are chunked without materializing the full edge-by-hidden Jacobian. The gradient includes SiLU, LayerNorm, later messages, and the decoder; residual sensitivities bypass SiLU but include downstream normalization and computation. Receptive-field bookkeeping retains every layer-edge instance that can reach the nominee, including residual reachability. Only structurally unrelated contractions are omitted; the forward path remains the full graph.

The implementation compares analytic tangents and edge contractions against explicit autograd calculations on small test models. It separately verifies that the actual-protein endpoint reproduces ordinary model scoring. These checks concern the chosen mathematical instrument, not the biological validity of its reference.

\subsection{Quadrature and reconciliation}
Gauss--Legendre points and weights are transformed from $[-1,1]$ to $[0,1]$. All feature, node, message, and residual integrals share the same positions and weights. We compare 16 versus 32 points, continuing to 64 or 128 if required. For each attribution array or specified layer slice $V$,
\begin{equation}
\|\widehat V_{2Q}-\widehat V_Q\|_1
\le10^{-5}+10^{-3}\|\widehat V_{2Q}\|_1.
\end{equation}
Each complete signed cut must also recover the independently evaluated score difference within the absolute-plus-relative tolerance based on its own absolute mass. This avoids judging numerical accuracy solely by a cancellation-sensitive total. Node/residual slices and message slices are checked separately per layer. Endpoint parity uses absolute and relative tolerances of $10^{-10}$.

In addition to receiving-node reconciliation, we check the allocation leaving each sender:
\begin{equation}
C_j^{(\ell)}=\sum_{e=(j\to i,r)}E_e^{(\ell)}+R_j^{(\ell)}.
\end{equation}
Maximum incoming/outgoing discrepancies must be at most $10^{-5}$. Failure to satisfy the criteria by 128 points stops the calculation. The saved accepted orders and errors for every case/member are reported below.

\begin{table}[t]
\centering
\caption{Saved numerical evidence for the original PYGL/CCDC86 protein-path cases and models. Completeness checks compare full signed cuts to the endpoint difference; reconciliation checks both ends of each transition. These are recorded computations, not biological-validation measurements.}
\begin{tabular}{llrrr}
\toprule
Nominee & Fit & Points & \shortstack{Completeness\\error} & \shortstack{Reconciliation\\error} \\
\midrule
PYGL & H1 & 32 & 9.77e-06 & 1.78e-15 \\
PYGL & H3 & 64 & 3.20e-08 & 7.11e-15 \\
CCDC86 & H1 & 32 & 1.23e-06 & 8.88e-15 \\
CCDC86 & H3 & 32 & 3.13e-06 & 8.88e-15 \\
\bottomrule
\end{tabular}

\label{tab:numerics}
\end{table}

\section{Consensus, route selection, and display semantics}\label{app:display}

\textbf{Representation geometry.} The notebook first follows the H1 query through encoded inputs and four message layers using one joint UMAP \citep{mcinnes2018}. Points are genes, with final nomination colors held fixed (the supplementary output \texttt{gene\_geometry\_paper.pdf}). These are stages of one model, not expression trajectories. Geometry makes the representation visible; conductance supplies its quantitative account.

This section describes the original protein-path displays. The stricter shared-sign selection and revised widths in Figure~\ref{fig:ttk-case} are specified in Appendix~\ref{app:ttk-figure}. The two models receive byte-identical native expression pairs but retain their own normalizers, projectors, means, and decoders. Candidate identities, graph relations, and layer indices align. We average signed scalar allocations after each member passes its numerical checks:
\begin{equation}
\bar C_i^{(\ell)}=\tfrac12\sum_{f\in\{\mathrm{H1,H3}\}} C_{i,f}^{(\ell)},\qquad
\Delta\bar s_k^{\rm protein}=\tfrac12\sum_f[s_{k,f}(1)-s_{k,f}(0)].
\end{equation}
The same rule applies to messages and residuals. Initial feature channels are summed within each gene before cross-model averaging; hidden coordinate bases are not aligned or averaged. Mean actual scores agree with the ranking table. Opposing member contributions can cancel before any positive-part display rule.

\begin{table}[t]
\centering
\caption{The original PYGL/CCDC86 fixed-request nominations. Mean-logit ranks use the same \Candidates{} candidates. Member ranks expose fit variation. $N_{90}$ counts the smallest set carrying 90\% of positive node conductance after three message rounds, before the final decoder state. It does not measure how many genes are biologically necessary.}
\begin{tabular}{lrcrr}
\toprule
Nominee & Consensus rank & H1 / H3 rank & $\Delta s^{\rm protein}$ & $N_{90}$ after 3 \\
\midrule
PYGL & 6 & 5 / 10 & +8.8719 & 8 \\
CCDC86 & 3 & 4 / 2 & +11.7054 & 1 \\
\bottomrule
\end{tabular}

\label{tab:cases}
\end{table}

\subsection{Select a structure without changing the attribution}
For one member, a route connects an edited marker's initial state to the nominee's final state with exactly one transition per layer, using positive messages or residuals. A widest route maximizes its minimum transition conductance \citep{vassilevska2009}. Ties prefer residuals, then relation and original-edge order. The union retains every node and connection used by at least one selected route; shared instances are counted once. Unreachable anchors remain recorded rather than receiving invented connectors.

For consensus, candidate routes are complete member witnesses. A witness is eligible only when every corresponding mean message/residual conductance exceeds $10^{-5}$; among eligible witnesses, the greatest bottleneck is selected for each anchor. This does not assemble a route from unrelated positive fragments of different fits. The path is a display of positive marginal transition allocations: positivity does not establish the sign or necessity of an isolated complete route. Exact ordered-route integration is a separate optional inspection, not a reported experiment on the primary displayed paths.

\subsection{Read the overview and profile as complementary objects}
Node size uses $\max_\ell[C_i^{(\ell)}]_+$, with a logarithmic mapping and a visibility/text floor. It is not an incident-edge sum or a conserved cross-layer total. Edge width sums positive conductance over all full-field layer instances of the displayed biological relationship; route selection determines which relationships appear, not the width's denominator. Reverse computational channels return to source-prior orientation for the overview. STRING physical/functional pairs remain undirected; regulatory sign annotations preserve positive, negative, mixed, or unassigned source evidence.

Names are the union of the 18 largest displayed node summaries, 12 highest displayed degrees, up to four strongest favorable-marker summaries, up to four strongest unfavorable-marker summaries, and the nominee. Overlap reduces the label count. Unnamed genes remain in the selected graph and exports. Node fill denotes the requested edit; neutral includes unchanged inputs and graph context without mapped expression. Neither a prior repression bar nor a requested decrease fixes the sign of a learned score contribution.

Full signed node arrays and typed message records underlie the profiles. Panel A separates positive mass from opposing magnitude for each complete node layer and message/residual transition. Panel B selects positive and opposing peaks, retains raw values with one within-case symmetric-log scale, and attributes the same final prediction in every column. Panel C's upper part counts components needed for 90\% of positive conductance; its lower part compares selected positive conductance with full positive conductance, separately for messages and residuals. A row with zero positive mass has undefined coverage, not an invented zero or 100\%.

The overview's log scales and profile color scales are case-specific display conventions. Quantitative cross-case comparisons use the recorded values and denominators. Layout positions and edge bends carry no biological distance or temporal meaning. Paper print reflow retains the scientific values, identities, selected nodes/relationships, and their source records.

\section{Finite message suppression}\label{app:suppression}
This section describes suppression for the original protein-path examples and defines the two outcomes retained in the appendix. The endpoint-score drop (Actual drop) measures how much deleting a message lowers the score at the path's endpoint:
\begin{equation}
D^{\rm actual}_{e,f}=s_{k,f}(1)-s_{k,f}^{\setminus e}(1).\label{eq:suppression-actual}
\end{equation}
Gap drop (Equation~\ref{eq:suppression-gap}) additionally subtracts the deletion effect at the reference endpoint. On the request path, these endpoints are requested expression and no expression change, so Gap drop is the request-response drop used in Figure~\ref{fig:request-population}; both outcomes are retained in Table~\ref{tab:expression-suppression}. On the original protein path below, they are actual and reference protein representations at fixed expression. Neither outcome compares different nominees.

The conductance values, native expression request, protein references, and checkpoint members match Section~\ref{sec:explanation} and the original examples (Appendix~\ref{app:original-examples}). Both members held A549 out of training and validation. Sampling uses the full signed receptive field, rather than only the positive routes selected for the network overview.

At each of the four transitions, we select up to six largest positive and six most negative mean message conductances. A $10^{-5}$ sign cutoff excludes near-zero entries; it is not a significance or integration-error threshold. All transitions here contain six extremes of each sign, well above that cutoff. For each positive selection we sample a low-$|E|$ control from the lowest absolute-conductance quartile and a random control in the same transition and typed relation. Selections are disjoint, unavailable controls are omitted, and seed 20260924 fixes the selection before suppression outcomes are evaluated. Each transition contains 24 sampled messages, except PYGL's final transition, which contains 17 because only three low-$|E|$ and two random controls are available.

An intervention sets one original layer/relation/edge gate from one to zero. Reverse channels, other occurrences of the same relationship, residuals, and original degree factors remain unchanged. We recompute the full forward pass at both actual and reference protein endpoints with frozen weights. The binary gate intervention produces a continuous score change through the nonlinear model. Conductance follows those nonlinearities along the original protein path; neither finite drop is algebraically equal to the original message conductance. Equal-weight mean member drops are compared with signed mean conductance using Spearman correlation. Selecting extremes and controls changes the sampled distribution, so these correlations are not estimates for all messages or significance claims.

The supplementary \texttt{notebook\_suppression\_groups.tex} shows directional agreement in the group means, with more consistent individual Gap-drop signs for CCDC86 than PYGL. Smaller mean control effects do not imply zero effects for every control. Pooling transitions summarizes the sampled messages without measuring the combined effect of suppressing a group or establishing independent biological replicates.

Some tested messages also lie on selected overview routes. For example, suppressing the final STRING regulatory PHKB $\to$ PYGL message lowers the actual score and protein-reference gap. The final GYG2 $\to$ PYGL reverse-channel message lowers the actual score but increases the gap, because the reference score falls further. This explains why the two contrasts must be retained. A pooled overview relationship can occur at several transitions; a layer-specific suppression tests only one instance. Residual paths and other messages remain available, so agreement does not rule out redundancy or nonlinear interactions.

The frozen evidence contains the selection, consensus and member effects, correlations, and source study receipts. The notebook retains the complete signed scatterplots and matched-control diagnostics. The original-case summary in Appendix~\ref{app:original-examples} retains the signed outcomes; the full plot is a supplementary output.

\section{Compound evidence and scope}\label{app:compounds}
\label{sec:repurposing}
The explorer is an interactive interface in the accompanying notebook for configuring expression requests and inspecting ranked nominees and their attribution networks. Its compound panel matches nominees by exact symbol to the pinned Broad Drug Repurposing Hub \citep{corsello2017,broad2025}, returning source-reported mechanisms, all targets, stages, and indications without changing rankings or the selected network.

For TTK, the source lists ten exact-symbol matches, including four with reported clinical stages; CKS2 has none (Table~\ref{tab:ttk-compounds}). The panel retains other annotated targets and source-reported actions so users can examine their compatibility with the hypothesis. These annotations do not establish selectivity or efficacy; an inhibitor need not reproduce a knockout. The original PYGL matches are described below.

\begin{table}[t]
\centering
\small
\caption{All exact-symbol TTK matches in the pinned source, with source-reported stage and mechanism labels. CKS2 has no exact-symbol match. Stages, mechanisms, and targets are source annotations, not measurements performed here.}
\label{tab:ttk-compounds}
\begin{tabularx}{\linewidth}{lllX}
\toprule
Compound & Reported stage & Source mechanism label & Annotated targets \\
\midrule
BAY1217389 & Phase 1 & kinase inhibitor & TTK \\
BOS-172722 & Phase 1 & monopolar spindle 1 kinase inhibitor & TTK \\
CFI-402257 & Phase 1/Phase 2 & dual specificity protein kinase inhibitor & TTK \\
Mps1-IN-5 & Phase 1 & protein kinase inhibitor & TTK \\
MPI-0479605 & Preclinical & mitotic kinase inhibitor & TTK \\
Mps-BAY-2a & Preclinical & monopolar spindle 1 kinase inhibitor & TTK \\
Mps1-IN-1 & Preclinical & monopolar spindle 1 kinase inhibitor & ALK, TTK \\
NMS-P715 & Preclinical & protein kinase inhibitor & TTK \\
TC-Mps1-12 & Preclinical & monopolar spindle 1 kinase inhibitor & TTK \\
pyrazolanthrone & Preclinical & JNK inhibitor & LRRK2, MAPK10, MAPK8, MAPK8IP1, MAPK9, TTK \\
\bottomrule
\end{tabularx}
\end{table}
The explorer uses the pinned August 18, 2025 Broad Drug Repurposing Hub annotation snapshot, checked by SHA-256 before use. Exact gene symbols are matched against the full target-list field. No fuzzy alias expansion or clinical-status inference is performed, and no selected gene is sent to the source server during offline matching.

For PYGL, the full set includes records with broad target lists, metabolite-related annotations, and differing source-reported stages. The notebook's PYGL table uses the explicit source-label filter \emph{Preclinical} plus \emph{glycogen phosphorylase inhibitor}; it does not claim those rows are exhaustive pharmacology or that a one-gene annotation proves selectivity. The full matched rows and source identity are preserved in the frozen paper evidence. CCDC86 has no exact-symbol match in this resource, which is not evidence that no compound exists or that the target is undruggable. For TTK, Table~\ref{tab:ttk-compounds} lists every exact-symbol match without a source-label filter; CKS2 has none, with the same interpretation. Most TTK records annotate only TTK, but a one-gene annotation does not establish selectivity.

A source label such as \emph{Launched} or \emph{Phase 3} is not an independent check of present FDA approval, oncology indication, trial status, target engagement, or an effect compatible with knockout. The role of the lookup is to place existing evidence beside a model-generated hypothesis for a researcher to examine.

\section{Robustness across QC-passing baselines}\label{app:qc-sensitivity}
The query baseline (Appendix~\ref{app:data}) is one of 44 A549 plates whose matched controls all pass QC (20--33 controls each). To test whether the original PYGL/CCDC86 cases depend on that choice, we repeat the query with each of these plates as a single-plate baseline---the same construction used for every training pair---and with a pooled mean of all 1,790 QC-passing A549 controls across 78 plates. The pooled baseline is smoother than any training baseline and is reported separately. Weights, graph, request rule, validated markers, and H1/H3 members are unchanged.

\begin{table}[h]
\centering
\small
\caption{\textbf{Rank robustness across QC-passing baselines.} Mean-logit ranks over H1/H3 among 17,225 candidates. The final row summarizes all 44 single-plate QC-passing baselines. Overlap and Spearman $\rho$ compare the full ranking with the query baseline.}
\label{tab:qc-ranks}
\begin{tabular}{lcccc}
\toprule
Baseline & PYGL (H1/H3) & CCDC86 (H1/H3) & Top-20 overlap & $\rho$ \\
\midrule
Query baseline (single plate) & 6 (5/10) & 3 (4/2) & 20/20 & 1.00 \\
Pooled QC-passing controls & 5 (3/9) & 2 (4/2) & 18/20 & 0.98 \\
44 QC-passing plates, median [range] & 5 [2--7] & 2 [2--6] & -- & -- \\
\bottomrule
\end{tabular}
\end{table}

Both nominees rank within the top 7 on every QC-passing plate (Table~\ref{tab:qc-ranks}). Member disagreement persists across plates (PYGL: H1 ranks 1--7, H3 ranks 6--15), so it is not confined to the selected baseline.

We also repeated protein-path conductance under the pooled baseline. For PYGL, 12 of the 15 largest positive node contributors are shared with the query baseline, including GYG2, PHKB, PHKG1, PHKG2, PYGM, and PYGB; the protein-reference difference is +5.44, $N_{90}$ after three rounds is 7, and final-transition messages contribute +1.73 against a residual of +3.71, so the glycogen-associated structure is stable while the magnitude of final-layer message support varies with the baseline. For CCDC86, all 15 largest contributors are shared, and the difference (+12.21), $N_{90}=1$, and residual-dominated final transition (+11.96) are essentially unchanged. Outputs and scripts will be released with the code.

\section{Execution records and reproducibility boundary}\label{app:reproduction}
The supplementary materials, to be released with the code and model weights upon publication, retain frozen inputs, outputs, source hashes, and reproduction commands for the reported analyses. The recorded notebook load-mode run contains 39 code cells; a historical full-training execution is separate. Building this revision uses saved results and does not rerun training. Numerical parity, completeness, and source checks establish computational consistency, not independent biological validation. The supplementary notebook and saved-field renderers reproduce additional geometry, training, and circuit views omitted from this PDF.

\section{Own-edit dependence and measured support of the original nominees}\label{app:legacy}
The PYGL and CCDC86 networks were the original illustrations. Their supporting structures remain accounts of the specified protein-reference contrasts, but do not establish dependence on the rest of the prognosis request. Both targets were themselves unfavorable markers requested to decrease. Table~\ref{tab:selfcue} records the decisive intervention on those inputs. Removing an edit restores that endpoint coordinate to its baseline; it does not remove the gene, its protein, or its graph messages.

\begin{table}[t]
\centering
\caption{\textbf{The original nominees rely on their own requested decrease.} Each entry gives H1/H3 common-query ranks over all \Candidates{} candidates; each intervention defines its own fixed input. ``Rest reversed'' retains the own decrease but flips every other edit.}
\label{tab:selfcue}
\begin{tabular}{lrrrr}
\toprule
Nominee & Full request & Own edit removed & Own edit only & Rest reversed \\
\midrule
PYGL & 5/10 & 3084/2214 & 1/1 & 1/1 \\
CCDC86 & 4/2 & 2670/2455 & 1/1 & 1/1 \\
\bottomrule
\end{tabular}
\end{table}

\subsection{Original examples: fixed-request protein-reference accounts}\label{app:original-examples}
The original protein-reference accounts are summarized below; their full figures and signed profiles can be regenerated from the supplementary notebook. Their different supporting structures motivate different biological hypotheses. Measured other-marker responses are reported separately in Appendix~\ref{app:measured}.

\subsubsection{PYGL: glycogen-associated regulatory and metabolic propagation}
PYGL has mean-logit rank \PyglRank{}, with member ranks \PyglFirstMemberRank{} and \PyglSecondMemberRank{}. Its selected structure includes phosphorylase-kinase components, glycogen branching/storage machinery, and a CollecTRI HIF1A$\to$GYS1 activation route into glycogen synthesis (the supplementary output \texttt{pygl\_network\_paper.pdf}). PYGL knockdown reduces proliferation and migration in lung cancer cells \citep{he2023}; subsequent NSCLC work connects PYGL-dependent glycogenolysis to glycolytic flux and tumor growth \citep{huo2026}. Broader cancer studies connect hypoxia-induced glycogen metabolism to redox balance and avoidance of senescence \citep{favaro2012}. Together these motivate a glycogenolysis-fueled glycolysis hypothesis in hypoxia-related signaling. The request does not specify oxygen tension, and the attribution measures neither metabolic flux nor the relative importance of metabolic and mitotic mechanisms.

The profile in the supplementary output \texttt{pygl\_profile\_paper.pdf} shows supporting and opposing allocations recovering the \PyglDifference{}-logit protein-reference difference. At $3\to4$, signed messages contribute \PyglFinalMessages{} and the residual \PyglFinalResidual{} logit units. This motivates a propagation-oriented interpretation: glycogen-associated regulatory relationships connect the nominee to a broader metabolic program. Whether PYGL loss produces the requested expression program through these processes remains the biological prediction.

\subsubsection{CCDC86: a physical-interaction context for ribosome biogenesis}
CCDC86 ranks \CcdcRank{} in the mean, with member ranks \CcdcFirstMemberRank{} and \CcdcSecondMemberRank{}. Its selected network includes nucleolar/ribosome-associated genes, with STRING functional links to WDR18, NLE1, and large-subunit ribosomal proteins and physical links to EIF6, LAS1L, ZNF593, and GNL2 (the supplementary output \texttt{ccdc86\_network\_paper.pdf}), motivating a hypothesis about partners involved in ribosome biogenesis and complex organization. CCDC86 has established chromosome-periphery and cell-division literature \mbox{\citep{stamatiou2023}} and functional evidence in glioma \citep{liu2025}. The context-specific hypothesis is that CCDC86 loss can move A549 toward the requested program, with nucleolar/ribosome processes participating. An independently measured response aligned with the request would support the endpoint prediction; a confirmed perturbation without that response would count against it. A response without the proposed functional involvement would weaken the mechanistic lead.

After three message rounds, the smallest set carrying 90\% of positive node conductance contains \PyglNodesNinetyAfterThree{} nodes for PYGL but \CcdcNodesNinetyAfterThree{} for CCDC86. CCDC86's final residual carries \CcdcFinalResidual{} logit units, while final messages contribute \CcdcFinalMessages{} (the supplementary output \texttt{ccdc86\_profile\_paper.pdf}). Earlier information is already present in that carried state. Together, the cases illustrate propagation through regulatory/metabolic relationships and attribution involving physical partners; both use the same message-passing architecture. These are interpretations of distinct allocations, not measurements of metabolic flux, complex assembly, or biological timing. Many higher-ranked nominees are well-characterized cancer genes; we selected PYGL as a nominee with supporting lung-cancer literature and CCDC86 as one with limited lung-cancer characterization. Complete rankings and signed fields remain available.

\subsubsection{Finite message suppression for the original examples}
The supplementary output \texttt{message\_suppression\_paper.pdf} summarizes \PyglSuppressionCount{} PYGL and \CcdcSuppressionCount{} CCDC86 sampled messages. Gap correlations are positive at every transition for both nominees: CCDC86 ranges from $\rho=\CcdcSuppressionGapThree{}$ to \CcdcSuppressionGapOne{} and PYGL from \PyglSuppressionGapFour{} to \PyglSuppressionGapTwo{}. Positive/negative selections have correspondingly signed average drops and control means lie closer to zero (the supplementary \texttt{notebook\_suppression\_groups.tex}). Individual PYGL gap signs agree less often (18/24 positive, 15/24 negative) than CCDC86's (23/24, 22/24); single-message deletion is a finite, nonlinear change that path-averaged conductance need not match edge by edge. These correlations describe the stratified sample, not all edges, total layer reliance, or necessity of the displayed network.

\section{Request audit, selection, and expression attribution}\label{app:query-audit}
\subsection{Three distinct rank definitions}
Let $S$ be the mapped requested coordinates and let $d^{(-k)}$ equal $d$ except that the candidate's own coordinate is restored to no change, if present. This intervention edits the endpoint input, not the graph or a learned representation; the baseline channel is unchanged. For the fixed full request, the common-query rank is
\begin{equation}
 r^{\rm common}_{kf}=1+\sum_{j\in\mathcal V_p}\one[s_{jf}(c,c+d)>s_{kf}(c,c+d)].
\end{equation}
A fixed own-masked query instead scores \emph{all} candidates on $(c,c+d^{(-k)})$ and ranks $k$ against those scores. This is the rank reported for ``own edit removed'' in Table~\ref{tab:selfcue}. The population screen uses a third quantity: the candidate-specific masked score $u_{jf}=s_{jf}(c,c+d^{(-j)})$ and its rank
\begin{equation}
 r^{\rm masked}_{kf}=1+\sum_{j\in\mathcal V_p}\one[u_{jf}>u_{kf}].
\end{equation}
Here each competitor has its \emph{own} edit removed; the ranked scores are not outputs under one common input. Finally, ranking the response $a_{kf}=u_{kf}-s_{kf}(q_0)$ would be yet another ordering, which can favor low-baseline candidates. We do not relabel response ranks as common-query ranks. At two SD, TTK has common-query ranks 18/17 and candidate-specific masked ranks 2/3; CKS2 has common-query rank 19 and candidate-specific masked rank 10. At empirical amplitudes the respective common-query ranks are 19/12 and 55, whereas masked ranks are 14/7 and 54. Neither nominee has an own edit, but masking competitors changes its screen rank. The masked ranks were independently recomputed from archived full-pool masked scores; they were not obtained from a fresh all-candidate masked inference run. Fresh canonical scoring verified the request/reversal responses and common-query ranks.

The computational screen requires $r^{\rm masked}_{kf}\leq200$, $a_{kf}>0$, and $b_{kf}<0$ in \emph{every} member, then mean $(a_{kf}-b_{kf})/2\geq0.5$. These logit thresholds are descriptive selection choices, not significance levels or calibrated probabilities. Full per-member endpoint tables, not rounded averages or historical pass flags, determine eligibility. Table~\ref{tab:screen-counts} retains all 14 available tissue-matched contexts, including zero-yield cells and seen-cell exploration, and yields 19 cases in 6 cells. Cases can share targets and requests, so the count is not 19 independent biological discoveries. Choosing TTK and CKS2 as detailed illustrations was post hoc, informed by the available audits and measured support; neither their selection nor their percentiles estimate prospective success.

\begin{table}[t]
\centering
\caption{The full corrected computational screen. Directional candidates pass every-member rank and sign requirements; Above floor additionally requires mean $(a_{kf}-b_{kf})/2\geq0.5$ logit. This floor does not denote measured biological support. All rows have 17,225 eligible candidates.}
\label{tab:screen-counts}
\begin{tabular}{llrrr}
\toprule
Cell & Exposure & Members & Directional & Above floor \\
\midrule
A375 & held out & 1 & 58 & 1 \\
A549 & held out & 2 & 9 & 4 \\
AGS & seen & 3 & 0 & 0 \\
DANG & held out & 1 & 54 & 4 \\
ES2 & seen & 3 & 2 & 0 \\
HCC1806 & held out & 1 & 57 & 0 \\
HS944T & held out & 1 & 69 & 4 \\
HT29 & seen & 3 & 0 & 0 \\
IPC298 & held out & 1 & 69 & 5 \\
MCF7 & held out & 2 & 8 & 0 \\
PC3 & seen & 3 & 1 & 0 \\
SNU761 & seen & 3 & 1 & 0 \\
U251MG & held out & 2 & 7 & 0 \\
YAPC & held out & 2 & 2 & 1 \\
\bottomrule
\end{tabular}
\end{table}

\subsection{Endpoint evidence and reversal}
The screen uses 28 members across 14 lines, with whole-line-held-out models where available and explicitly flagged seen-cell members otherwise. Nine cell-line contexts, including A549 and DANG, use HPA's validated tier; five (A375, HS944T, IPC298, AGS, and PC3) use potential-tier markers. No single marker-evidence policy is silently imposed on all cells. HS944T uses 170 potential melanoma markers with two-reference-SD edits and the saved internally assessed-control baseline; A549 uses the original 157-coordinate validation-tier request and QC-passing baseline. The different marker and control policies limit a direct comparison between their accounts.

An external direction check against the adult PRECOG patient-cohort compendium \citep{gentles2015precog} finds 85.8\% agreement among 2,051 matched marker-direction entries, and 96.6\% among 953 entries with $|z|\geq1.96$ in PRECOG (positive $z$ denotes adverse prognosis). PRECOG concordance was assessed separately, without requiring agreement for marker inclusion. These are pooled entries with repeated cell/cohort uses, not independent genes or patients. Agreement supports the choice of prognostic directions, not the two-SD amplitude, intervention causality, or a survival benefit in cultured cells.

Across the 28 members, the median fraction of top-50 response-ranked nominees leaving the top 50 under reversal is 82\%; median all-candidate response/reversal correlation is $-0.40$. Nevertheless, a median 44\% of those top-50 nominees do not decrease relative to no change under reversal. Leaving a top list and having a negative reversed response are different tests. We therefore use the explicit sign requirements of Section~\ref{sec:cases}, rather than a rank shift alone, to distinguish directional candidates from magnitude responders.

\subsection{Expression IG and conductance share the request response}
For a fixed member with actual proteins and graph held constant, define
\begin{equation}
 I^x_{jkf}=d^{(-k)}_j\int_0^1
 \frac{\partial s_{kf}(c,c+\alpha d^{(-k)})}{\partial x^{(1)}_j}\,d\alpha,
 \qquad \sum_j I^x_{jkf}=a_{kf}.
 \label{eq:expression-ig}
\end{equation}
The integral allocates the joint no-change-to-request response; it is not the effect of making edit $j$ alone. Unchanged coordinates have zero attribution, including the masked own coordinate. Each member receives the same native request in its own normalization coordinates. The network conductance in Section~\ref{sec:request-path} follows this \emph{same} expression path and accounts for $\Delta s_k^x=a_{kf}$. Protein-reference conductance instead keeps expression fixed and interpolates projected proteins (Section~\ref{sec:explanation}); its endpoint difference and signs must not be interchanged with the expression response.

The six case/member/amplitude computations for TTK/A549 (H1/H3) and CKS2/HS944T (H3), each at two request sizes, use shared 32-point Gauss--Legendre quadrature for raw IG, nodes, messages, and residuals. Maximum raw-IG/initial-node disagreement is $1.94\times10^{-16}$; maximum complete-cut discrepancy against fresh raw score differences is $1.58\times10^{-8}$. The two-SD CKS2 input account retains all 170 coordinates, with positive total 1.135624 and negative total $-0.367135$, giving 0.768490; at the empirical amplitude the corresponding totals are 0.110203 and $-0.060073$, giving 0.050130. Its own-coordinate attribution remains zero.

\section{The TTK expression-path figure: full view and display semantics}\label{app:ttk-figure}

\paragraph{Biological meaning of the signed account.}
The biological question is which relationships make TTK loss a plausible response to this A549 expression program. The prior makes 92 immediate TTK neighbors available; the learned computation gives their contextualized contributions a sign and magnitude. Table~\ref{tab:ttk-biology} links representative numerical observations to experimentally established functions. The quantities are contributions to the joint request-minus-no-change logit, with protein representations fixed. Input IG and layer-message conductance are different accounts of that same difference, so they are compared within their respective accounts rather than added across them.

\begin{table}[!ht]
\centering
\small
\setlength{\tabcolsep}{4pt}
\begin{tabularx}{\linewidth}{@{}>{\raggedright\arraybackslash}p{0.21\linewidth}>{\raggedright\arraybackslash}p{0.25\linewidth}>{\raggedright\arraybackslash}X@{}}
\toprule
Entity / relationship & Signed observation & Literature-grounded interpretation \\
\midrule
MAD2L1 and BUB1B & Down-input IG: $+0.251/{+}0.236$ and $+0.169/{+}0.048$ (H1/H3). & MAD2 and BUBR1 participate in the checkpoint complex that inhibits APC/C \citep{sudakin2001mcc}. Decreases in both supply positive evidence for loss of the kinase organizing this machinery \citep{ji2017mps1}. \\
\addlinespace
PLK1 & Down-input IG: $+0.059/{+}0.070$ (H1/H3). & PLK1 strengthens MPS1 activity and checkpoint-component recruitment \citep{vonschubert2015plk1}. Its decrease supplies concordant, smaller support than MAD2L1 in each model. \\
\addlinespace
MAD1L1--TTK & Final message: $+0.420/{+}0.150$; MAD1L1 has no expression edit. & MAD1 is an MPS1 substrate and a scaffold in MAD2-associated signaling \citep{ji2017mps1}. Support through its contextualized state identifies an unedited checkpoint intermediary. \\
\addlinespace
E2F4--TTK & Messages at $1\to2$ and $3\to4$: $-0.096/{-}0.058$ (H1), $-0.115/{-}0.071$ (H3); H1 is $+0.141$ at $2\to3$. & E2F4-containing DREAM binds the TTK promoter and regulates its transcription \citep{muller2017timing}. This distinct regulatory relationship carries counterevidence in both models, unlike final MAD1L1 support. \\
\bottomrule
\end{tabularx}
\caption{\textbf{Numerical discrimination among biological roles around TTK.} Values use the original two-SD request. The papers establish functions of checkpoint proteins and transcriptional regulators; the interpretation column relates those functions to the model's A549 nomination. Positive/negative attribution means support/opposition to TTK knockout, independently of a prior's activating/repressing annotation.}
\label{tab:ttk-biology}
\end{table}

\paragraph{Interpretation and display.}
Checkpoint membership does not determine support: BUB3, TK1, and BRCA1 input signs differ between H1/H3. E2F4/DREAM promoter regulation was characterized in T98G and fibroblast systems \citep{muller2017timing}; its attribution here concerns the A549 request. TTK depletion in A549 also affects proliferation and migration through other reported accounts \citep{tsai2020ttk}; our comparison does not test those alternatives head to head. It supplies a checkpoint-related lead with named supporting and opposing contributions. These findings suggest examining whether the requested marker response tracks checkpoint assembly, transcriptional regulation, or both. The channel signs do not decide that question in A549.

\begin{figure}[p]
\centering
\includegraphics[width=0.88\linewidth]{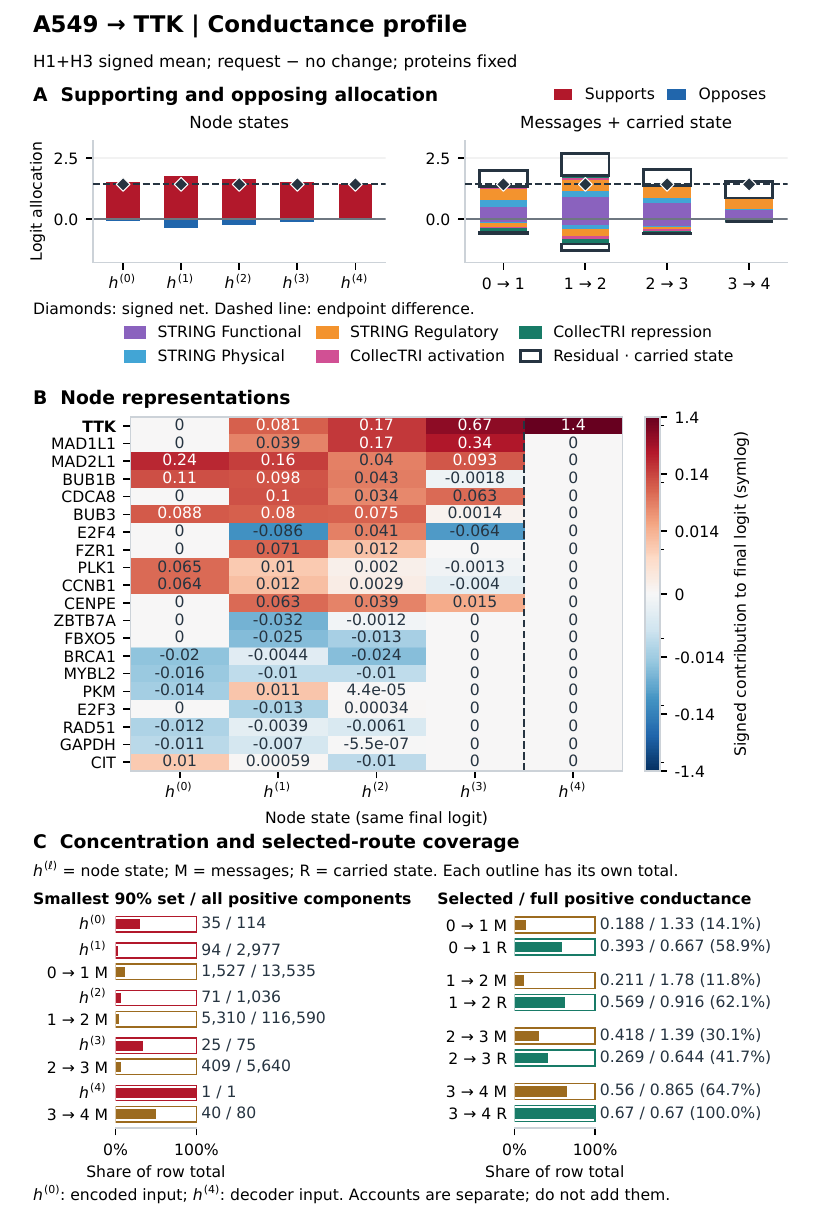}
\caption{\textbf{TTK: signed conductance behind the network.} Supporting and opposing allocations recover the same mean request-minus-no-change difference ($+1.423$) at every node state and transition, and the final transition splits it between messages ($+0.753$) and the carried residual ($+0.670$). The panels use the request path with actual proteins fixed and H1/H3 signed scalar means. A gives separate node and transition accounts of the same difference. B attributes the same final logit at each node state, from encoded input $h^{(0)}$ to decoder input $h^{(4)}$; the nominee-only final column is architectural, and genes without an expression edit start at zero. C separates component counts from positive mass selected by the complete overview (the supplementary output \texttt{ttk\_overview.pdf}), with a different denominator for each row. Red supports and blue opposes the nominee. Residuals include earlier messages; complete layer totals are not added.}
\label{fig:ttk-profile}
\end{figure}

Panel A uses node conductance at $h^{(3)}$, not input IG or final-message credit. Its candidate set is the partners displayed in B, excluding the nominee. We require each selected sign in both fits, retain the six largest positive and four most negative signed means, and order that subset from positive to negative. The top negative full-field node omitted from B is CDC27 ($-0.0123/{-}0.0018$); the strip is therefore not a global top-ten ranking. Full-field totals and additional node states are in Figure~\ref{fig:ttk-profile}. MAD1L1's node credit ($+0.432/{+}0.245$) differs from its final regulatory-reverse message ($+0.420/{+}0.150$), because node credit includes all outgoing message and residual allocations.

\begin{figure}[p]
\centering
\includegraphics[width=\linewidth]{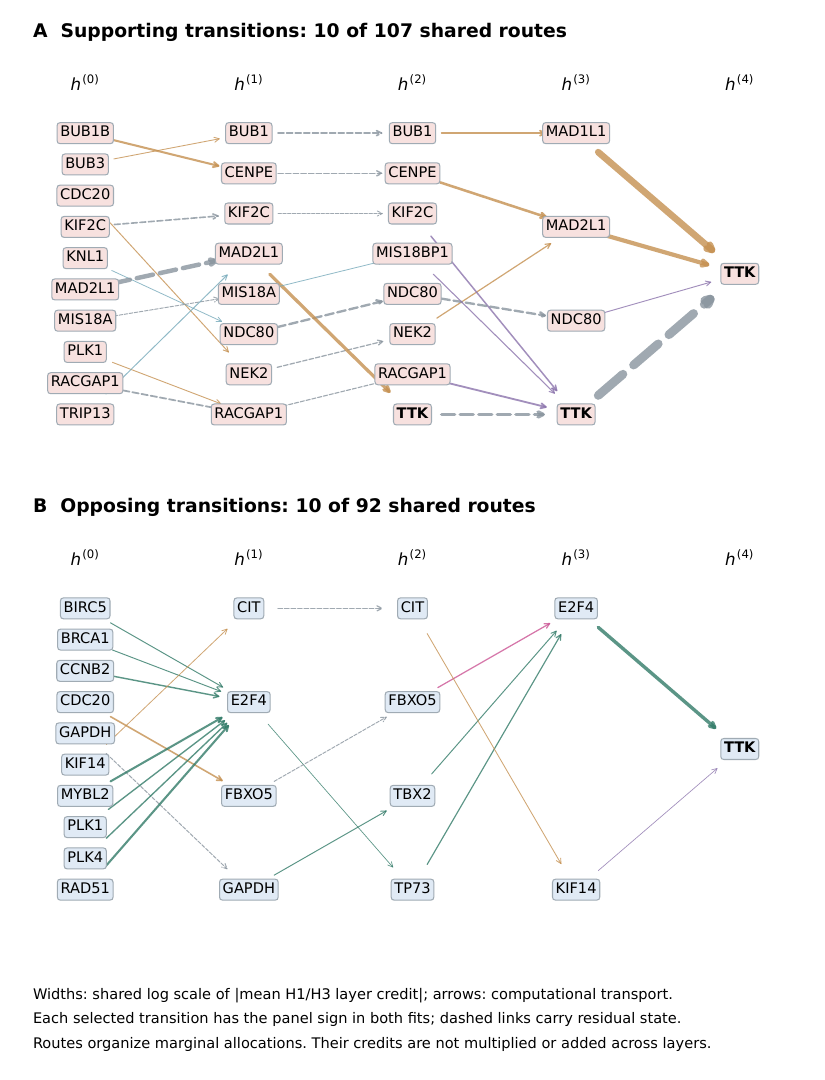}
\caption{\textbf{Separate supporting and opposing route displays from the same TTK field.} Each panel retains the ten widest routes under the smaller H1/H3 absolute transition credit, requiring every message and residual transition to exceed $10^{-5}$ with the panel sign in both fits (107 supporting and 92 opposing eligible anchors). Columns preserve computational layers; arrowheads denote transport, including reverse channels, rather than biological causation. Colors distinguish STRING functional (purple), physical (blue), regulatory (orange), CollecTRI activation (pink), repression (green), and residual state (gray dashed). Widths use the same logarithmic scale of absolute mean layer credit in both panels. Signs are marginal allocations along the request path, not products of edge signs or complete-path causal effects.}
\label{fig:ttk-signed-routes}
\end{figure}

The complete mean overview retains 124 of 157 edited anchors; the shared positive core retains 107, with the top 20 drawn in Figure~\ref{fig:ttk-case}. All complete cuts recover the mean response $+1.423$ within $7.9\times10^{-9}$. Final messages contribute $+0.753$ and the residual $+0.670$. Figure~\ref{fig:ttk-case} selects the twenty widest shared positive routes using the smaller member's transition credit, each transition exceeding $10^{-5}$ in both fits. They cover 59\%/72\% of positive final-transition and 12\%/9\% of all positive message conductance. Positive relationship widths use the maximum positive mean layer-instance credit; opposing overlays use mean absolute final-message credit on the same logarithmic scale. The archived overview instead pools positive credit across layers. Neither is conserved cross-layer flow. Layer identity matters: H3 MAD1L1-to-TTK is $-0.271$ at $2\to3$ but $+0.150$ at $3\to4$. Appendix Figure~\ref{fig:ttk-signed-routes} retains layers when displaying both signs.

At the empirical amplitude, MAD2L1 remains the strongest input ($+0.130/{+}0.183$), BUB1B supports ($+0.068/{+}0.057$), and final MAD1L1 support persists ($+0.269/{+}0.059$). Nevertheless, the H3 $2\to3$ message-field cosine across amplitudes is $-0.172$; persistent nomination does not imply an invariant internal account. Absolute input IG outside the complete one-hop neighborhood is 28.0\%/17.3\% at the empirical size. These are global-field allocations, not source-specific path integrals or effects of isolated edits.

\paragraph{Recurrence in DANG under H2.}
DANG has one held-out model, H2, which trains on A549. Removing the requested TTK decrease leaves rank 44 (response $+0.895$, reversed $-1.438$). Only 15 edits are shared with A549, excluding MAD2L1, BUB1B, and PLK1. Final MAD1L1 support recurs ($+0.092$), while E2F4 supports ($+0.101$), unlike A549; the pre-declared reading is partial (Table~\ref{tab:crossline}). At 0.271 SD, rank is 152 (response $+0.169$, reversed $-0.181$). Deletion correlations are $0.78,-0.35,0.46,0.56$ across transitions on the 88-message panel. Measured agreement reaches the 82.1st percentile (six wells). This changes request, cell, and fitted model together.

\section{Endpoint-amplitude sensitivity on the common expression path}\label{app:amplitude}
The original endpoint moves each marker by two reference SD in its HPA-associated direction. The empirical endpoint uses 0.527130407904365 reference SD in A549 and 0.2751588147442933 in HS944T. These are the cell-specific 90th percentiles of observed knockout-induced per-marker RMS displacement, computed from target-mean responses for knockouts with at least three wells. The marker directions, baseline, model, actual proteins, normalizer, and own-mask policy are unchanged. This retrospective magnitude anchor does not confirm the directions, establish simultaneous attainability, or specify a clinical target or dose. Both scales were retained; the scaled ranks were known before the attribution comparison, rather than choosing an amplitude afterward to rescue a case. The 19-case screen in Appendix~\ref{app:query-audit} concerns the original two-SD requests, not a repeated population screen at empirical amplitudes.

\begin{table}[t]
\centering
\small
\caption{Both amplitudes, with distinct rank definitions. Masked rank removes each competitor's own edit; Common rank uses one fixed query. Scores are raw request/reversal differences from no change.}
\label{tab:amplitudes}\label{SCALE_TABLE}
\begin{tabular}{llrrrrr}
\toprule
Case/member & Scale & SD & Masked & Common & Request & Reverse \\
\midrule
TTK H1 & original & 2.000000 & 2 & 18 & $+1.517$ & $-1.834$ \\
TTK H1 & empirical & 0.527130 & 14 & 19 & $+0.623$ & $-0.416$ \\
TTK H3 & original & 2.000000 & 3 & 17 & $+1.329$ & $-1.216$ \\
TTK H3 & empirical & 0.527130 & 7 & 12 & $+1.030$ & $-1.068$ \\
CKS2 H3 & original & 2.000000 & 10 & 19 & $+0.768$ & $-0.378$ \\
CKS2 H3 & empirical & 0.275159 & 54 & 55 & $+0.050$ & $-0.018$ \\
\bottomrule
\end{tabular}
\end{table}

TTK remains request-positive and reversal-negative in both overlapping training partitions. CKS2 retains those signs but falls below the original mean 0.5-logit directional threshold. Common-query no-change ranks are 381/196 for TTK H1/H3 and 65 for CKS2 H3; reversed common-query ranks are 3309/1622/132 originally and 1176/1389/70 empirically. These are not masked nomination ranks.

\subsection{Field similarity and multi-hop input dependence}
For two aligned signed fields $v,w$, define shared signed mass by separately dividing each by its absolute L1 norm and summing $\min(|v_i|/\|v\|_1,|w_i|/\|w\|_1)$ only where signs agree. This measures shape overlap, not a significance test or invariance of the raw magnitudes. Table~\ref{tab:field-shape} retains all message transitions. TTK H3's negative $2\to3$ cosine rules out describing its internal account as unchanged. Conversely, MAD2L1 remains its strongest absolute input-IG marker in both models at both amplitudes. The nominee-only final node is architecturally concentrated and is not a stability test.

\begin{table}[t]
\centering
\small
\caption{Original versus empirical expression fields. Message metrics use the topology-reachable field; each four-tuple orders transitions $0\to1$ through $3\to4$.}
\label{tab:field-shape}
\begin{tabular}{llcc}
\toprule
Case/member & Input overlap & Message signed overlap & Message cosine \\
\midrule
TTK H1 & 0.860 & 0.742, 0.731, 0.565, 0.821 & 0.861, 0.936, 0.486, 0.980 \\
TTK H3 & 0.781 & 0.561, 0.526, 0.472, 0.559 & 0.687, 0.437, -0.172, 0.520 \\
CKS2 H3 & 0.806 & 0.793, 0.711, 0.682, 0.765 & 0.968, 0.642, 0.780, 0.933 \\
\bottomrule
\end{tabular}
\end{table}

Absolute input-IG shares outside the complete one-hop incoming prior neighborhood are 30.2\%/28.0\% for TTK H1, 26.3\%/17.3\% for TTK H3, and 58.6\%/61.8\% for CKS2 H3 (original/empirical). The exported neighbor sets exactly match the complete prior: 92 distinct neighbors for TTK and 34 for CKS2. With node-local encoding, normalization within each node, fixed input normalizers/proteins, and nominee-local decoding, non-neighbor expression influence requires at least two graph-message hops. The denominator is total \emph{absolute input IG}; these percentages are neither fractions of net score, sums of layer credits, nor selected-drawing coverage. They establish multi-hop input dependence, not a particular attributed route.

\subsection{All finite-message outcomes}
Each expression field has its own stratified signed-extreme/control panel, chosen before observing suppression effects. The first three transitions each contain 24 messages; final-transition counts are 18/20 for TTK H1, 16/16 for TTK H3, and 16/21 for CKS2 (original/empirical). Thus total panel sizes are 90/92, 88/88, and 88/93. Each intervention removes one typed directed layer-message while keeping the original normalization and residual branches. Full scores are recomputed at request and no change, with actual proteins fixed. Table~\ref{tab:expression-suppression} distinguishes request-logit drop from request-response drop. Panels differ across amplitudes and members; the correlations cannot rank the amplitudes' comparative faithfulness. Negative and small values remain visible.

\begin{table}[t]
\centering
\small
\caption{Expression-path finite suppression: correlations between signed message conductance and finite drops. Request is the endpoint-score drop (Actual drop); Response is the request-response drop (Gap drop), which subtracts the deletion effect at no change. Both outcomes and all transitions are retained.}
\label{tab:expression-suppression}
\begin{tabular}{lllrrrr}
\toprule
Case/member & Scale & Outcome & $0\to1$ & $1\to2$ & $2\to3$ & $3\to4$ \\
\midrule
TTK H1 & original & Request & 0.328 & 0.125 & -0.397 & -0.639 \\
TTK H1 & original & Response & 0.551 & 0.540 & 0.305 & -0.418 \\
TTK H1 & empirical & Request & 0.348 & -0.377 & -0.081 & 0.774 \\
TTK H1 & empirical & Response & 0.810 & 0.570 & 0.697 & 0.358 \\
TTK H3 & original & Request & 0.164 & -0.089 & 0.025 & -0.168 \\
TTK H3 & original & Response & 0.528 & 0.329 & 0.080 & -0.476 \\
TTK H3 & empirical & Request & 0.475 & -0.330 & 0.269 & 0.471 \\
TTK H3 & empirical & Response & 0.507 & 0.090 & 0.319 & -0.191 \\
CKS2 H3 & original & Request & 0.923 & -0.041 & 0.083 & 0.606 \\
CKS2 H3 & original & Response & 0.923 & 0.478 & 0.418 & 0.744 \\
CKS2 H3 & empirical & Request & 0.044 & -0.203 & 0.643 & 0.292 \\
CKS2 H3 & empirical & Response & 0.803 & 0.445 & 0.404 & 0.832 \\
\bottomrule
\end{tabular}
\end{table}

All six saved tasks completed at 32 quadrature points; saved-array cuts, raw-score parity, and raw-IG/initial-node agreement were verified separately from these finite interventions. Numerical completeness is not a guarantee of deletion agreement, and positive deletion agreement does not establish biological efficacy.

\section{CKS2 companion account}\label{app:cks2}

\subsection{Expression-path account at both amplitudes}
The supplementary output \texttt{cks2\_overview.pdf} applies the submitted single-model selection to every requested edit (157 of 170 reach CKS2; 356 genes, 364 relationships); panel B of the supplementary output \texttt{cks2\_network.pdf} draws its 15 routes with the largest bottleneck, and the supplementary output \texttt{cks2\_profile.pdf} gives the signed profile. After three rounds, 12 of 26 positive nodes carry 90\% of positive node conductance; final messages contribute $+0.642$ and the residual $+0.127$. The strongest final message into CKS2 comes from SKP2 ($+0.223$), then TP73 and TP63 ($+0.109$, $+0.067$). HS944T/H3 uses the potential-marker melanoma request, its saved baseline, and no own CKS2 edit.

At two SD, SKP2 down contributes $-0.329090$, while CDT1 down ($+0.293241$), NR4A3 up ($+0.265462$), and NFKB2 up ($+0.177753$) support CKS2. At 0.275159 SD these become $-0.054926$, $+0.010105$, $+0.034690$, and $+0.021917$, respectively. Thus CDT1's contribution shrinks relative to the other leading inputs, and the net response falls from $+0.768$ to $+0.050$ despite positive response-drop correlations.

The final SKP2-to-CKS2 STRING regulatory reverse-channel message is positive at both amplitudes ($+0.222538$ and $+0.025240$), despite negative SKP2 input IG. This is not a contradiction: after earlier message passing, a sender state contains information from other genes, and its global-field message conductance is not a source-specific allocation of that sender's original edit. The TP73-to-CKS2 CollecTRI repression channel is also positive ($+0.108613$ and $+0.009049$); its biological-prior sign and computational attribution sign answer different questions. At the empirical amplitude 61.8\% of absolute input IG lies outside CKS2's complete one-hop prior neighborhood. The positive finite response-drop correlations in Table~\ref{tab:expression-suppression} concern the fitted response, not its magnitude, therapeutic value, or an established route.

\subsection{Cross-line comparison: HS944T/H3 and IPC298/H1}\label{app:crossline}
\begin{figure}[t]
\centering
\includegraphics[width=\linewidth,trim=0 296bp 0 0,clip]{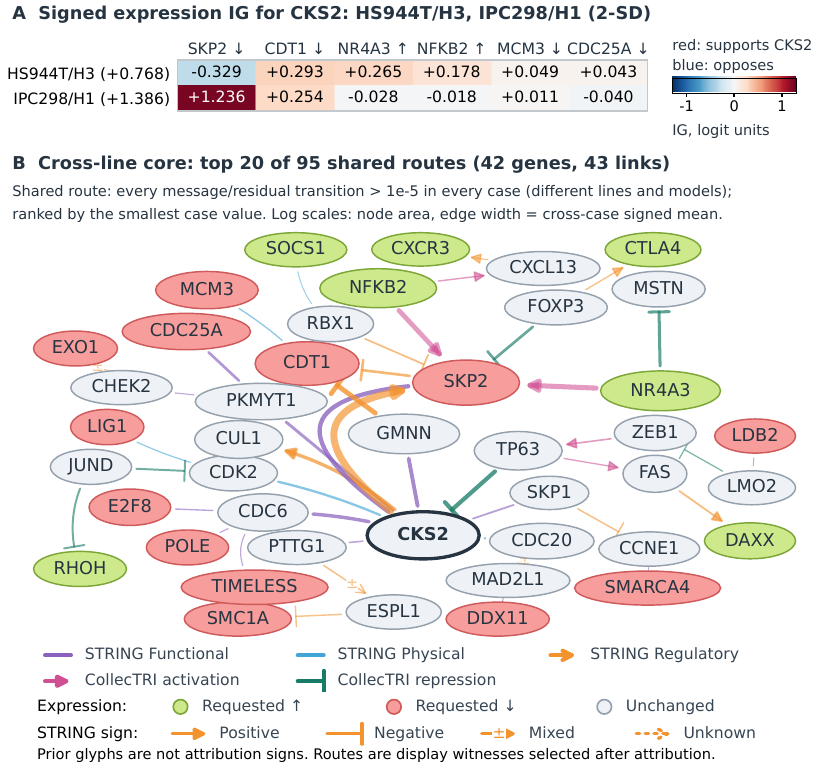}
\caption{\textbf{One CKS2 request, two melanoma lines: the same edit can contribute differently.} Signed expression IG for the six shared edits with the largest $|$IG$|$ in either line; both requests contain the same 170 edits and directions. SKP2 down opposes CKS2 in HS944T/H3 but supports it in IPC298/H1, whereas CDT1 down supports both. Parentheses give each request response. Red supports and blue opposes CKS2. Cell and fitted model change together, so this comparison does not isolate a cell-context effect. The corresponding shared-route network is retained in the supplementary reproduction output \texttt{cks2\_crossline\_network.pdf}.}
\label{fig:cks2-case}
\end{figure}

CKS2 has no own edit in the melanoma request and passes the screen in two lines, each with one held-out model: HS944T/H3 (rank 10, response $+0.768$, reversed $-0.378$) and IPC298/H1 (rank 148, $+1.386$, $-2.221$). Each melanoma line is held out by only one fold, so the two accounts differ in line and in model. Both requests contain the same 170 edits in the same directions, so inputs are compared edit by edit (Figure~\ref{fig:cks2-case}). Under the pre-declared criteria below, the input account is divergent.

SKP2 down opposes CKS2 in HS944T ($-0.329$) but is its dominant support in IPC298 ($+1.236$); NR4A3 and NFKB2 up support in HS944T ($+0.265$, $+0.178$) but weakly oppose in IPC298 ($-0.028$, $-0.018$). Over all 170 edits, Spearman $\rho=-0.003$. Two features recur: CDT1 down supports CKS2 ($+0.293/{+}0.254$), and SKP2 sends the strongest supporting final message into CKS2 in both lines ($+0.223/{+}0.359$). Of the 170 edits, 95 have a route positive in both lines at every transition (the supplementary output \texttt{cks2\_crossline\_network.pdf} displays the top 20). The displayed routes carry 49\%/83\% of HS944T/IPC298 positive final-transition and 15\%/19\% of all positive message conductance. The supplementary network pools positive relationship credit over layers, unlike the peak-layer widths in Figure~\ref{fig:ttk-case}. The opposing SKP2 edit thus describes the HS944T/H3 computation, not CKS2 in melanoma generally.

Measured CKS2 knockouts reach the 95.0th and 90.4th percentiles (6 wells each). With one model per line, potential-tier markers, and strong sensitivity to request size (Appendix~\ref{app:amplitude}), this comparison shows how recurring relationships can coexist with different input support.

The all-endpoint screen also passes CKS2 in IPC298 under H1 and TTK in DANG under H2. Before computing either field, we fixed the comparison statistics, named checks, and outcome labels. Both fields use the same frozen checkpoints, request path, quadrature, and reconciliation as the featured cases. Every complete cut recovers the request-minus-no-change difference within $3\times10^{-8}$, and initial node conductance matches input IG within $10^{-16}$. The screen's own code recomputes each member's 17,225-candidate table: logits agree within $9\times10^{-14}$ and the nominee's own-masked rank exactly; 2--7 near-tied paralog candidates per rank column swap order. Node and typed-edge index spaces are identical across lines for each nominee, so cross-line min-consensus cores are defined. Inputs are compared over edits requested in both cases with the same direction. \emph{Material} edits have $|$IG$|\ge0.01$ in both. Labels were fixed in advance. \emph{Convergent} requires $\rho\ge0.5$, material sign agreement $\ge75\%$, all applicable named checks passing, and at least 10 shared routes. \emph{Divergent} applies when $\rho\le0.2$, material agreement is below $60\%$, most named checks fail, or fewer than 5 routes are shared. Otherwise the result is \emph{partial}.

\begin{table}[t]
\centering
\small
\setlength{\tabcolsep}{3.5pt}
\caption{Cross-line comparisons on the two-SD request path (pre-declared readings). Final: incoming layer-4 relationships positive in both cases (positive in first/second). Core: routes positive in every case at every transition. The A549 H1/H3 row (same line and request) is a scale reference only.}
\label{tab:crossline}
\begin{tabular}{lrrrrrl}
\toprule
Comparison & Common edits & Material agree & $\rho$ & Final & Core & Reading \\
\midrule
CKS2 HS944T/H3 vs IPC298/H1 & 170 & 3/10 & $-0.003$ & 29 (39/39) & 95 & divergent \\
TTK A549/H1 vs DANG/H2 & 15 & 5/6 & $0.264$ & 60 (79/70) & 10\textsuperscript{a} & partial \\
TTK A549/H3 vs DANG/H2 & 15 & 5/5 & $0.350$ & 42 (59/70) & 10\textsuperscript{a} & partial \\
\emph{TTK A549/H1 vs A549/H3} & 157 & 29/36 & $0.411$ & 43 (79/59) & 107 & -- \\
\bottomrule
\end{tabular}

\smallskip
{\footnotesize \textsuperscript{a}Three-way core over A549/H1, A549/H3, and DANG/H2.}
\end{table}

The 29 shared-positive final relationships carry 70\%/85\% of each line's positive final-message conductance, but their rank agreement is weak ($\rho=0.171$). At 0.249 SD in IPC298, CKS2 falls to own-masked rank 406 (response $+0.258$, reversed $-0.277$); SKP2 down remains its dominant support ($+0.232$). IPC298's 94-message deletion panel gives correlations $0.44,0.40,0.28,0.46$ across transitions. These results retain the divergent account and amplitude sensitivity rather than treating the two melanoma cases as independent replication.

\section{Measured marker agreement and the cross-cell measured-signature reference}\label{app:measured}
For knockout target $k$, request cell $c$, and well $n$ in that cell, let $S_c\setminus\{k\}$ be the requested markers after excluding the target itself. With requested sign $\eta_j\in\{-1,+1\}$, measure
\begin{equation}
 A_{nkc}=\frac{1}{|S_c\setminus\{k\}|}
 \sum_{j\in S_c\setminus\{k\}}\eta_j
 \frac{v^{\rm endpoint}_{nj}-\bar v^{\rm plate(n)}_j}{s^{\rm H1}_j}.
 \label{eq:measured}
\end{equation}
We average over the target's wells and rank target means among all measured knockouts in that cell, using average tied percentile rank. This is a signed mean displacement, not cosine similarity, a fraction of markers achieved, or survival benefit. Excluding $k$ prevents its direct depletion from supplying measured program agreement. Replicate wells are not independent patients or an independent validation dataset. These are existing LINCS responses, assessed separately from the constructed model query.

\textbf{Original nominees.} Measured A549 knockouts of PYGL reach the 70.8th percentile of requested-marker agreement (17 wells; mean 0.140252 reference SD), and CCDC86 the 88.5th percentile (9 wells; mean 0.243854 reference SD), with the target's own marker excluded. These measured responses are separate from the own-edit dependence of their nominations (Appendix~\ref{app:legacy}).

For TTK in A549 the mean is 0.417864 reference-SD units over requested markers, percentile 98.2 among 5,115 measured target identities, with 10 wells. For CKS2 in HS944T it is 0.074210, percentile 95.0 among 562 targets, with 6 wells. These means are far below a uniform two-SD shift, although the percentiles are high. All TTK wells and five of six CKS2 wells exceed their respective cell's median well agreement. The ranking comparison in Table~\ref{tab:forward} uses the narrower intersection of eligible inverse candidates and available other-cell profiles (4,970 and 544 targets, respectively); that intersection is not the percentile denominator.

The same metric applied to all 19 computational-screen cases (Table~\ref{tab:screen-measured}) gives measured support for 10 of the 17 measured cases, including both TTK lines and both CKS2 lines, and no measured support for the other 7; GCLC in DANG and CKS1B in IPC298 have no measured knockouts. Nine of the ten supported cases have at least three wells (CHEK1 in HS944T has one). Repeated genes across lines make these cases dependent, so we report counts rather than a test.

\begin{table}[t]
\centering
\caption{Measured responses of candidates passing the computational screen. Agreement is mean other-marker alignment (Equation~\ref{eq:measured}), averaged over the target's wells in reference-SD units, excluding its own marker. The within-cell percentile ranks this mean among all measured knockouts; it is not clinical success. Measured-support thresholds were fixed before the original run and are separate from the computational screen. All 19 context--gene cases are retained and are not independent: TTK, CKS2, and CAMSAP2 each appear in two lines.}
\label{tab:screen-measured}
\small
\begin{tabular}{llrrr}
\toprule
Cell & Gene & Wells & Agreement & Percentile \\
\midrule
\multicolumn{5}{l}{\textbf{Measured support ($\geq$75th percentile; 10 cases)}} \\
A549 & TTK & 10 & 0.418 & 98.2 \\
A549 & TOP2A & 21 & 0.370 & 97.0 \\
A549 & CAMSAP2 & 12 & 0.369 & 97.0 \\
HS944T & CHEK1 & 1 & 0.079 & 96.4 \\
HS944T & CKS2 & 6 & 0.074 & 95.0 \\
IPC298 & CKS2 & 6 & 0.036 & 90.4 \\
HS944T & CDK2 & 9 & 0.056 & 89.5 \\
DANG & TTK & 6 & 0.039 & 82.1 \\
A549 & SORL1 & 6 & 0.177 & 78.6 \\
DANG & CAMSAP2 & 6 & 0.031 & 76.3 \\
\midrule
\multicolumn{5}{l}{\textbf{No measured support ($\leq$50th percentile; 7 cases)}} \\
IPC298 & CBLB & 6 & -0.003 & 39.5 \\
A375 & AVPR2 & 5 & -0.005 & 32.3 \\
HS944T & CDC7 & 2 & 0.002 & 31.7 \\
YAPC & CD2 & 5 & -0.024 & 23.1 \\
DANG & TPR & 5 & -0.023 & 19.2 \\
IPC298 & WWTR1 & 6 & -0.018 & 13.4 \\
IPC298 & CALCA & 6 & -0.035 & 1.3 \\
\midrule
\multicolumn{5}{l}{\textbf{Not measured (2 cases)}} \\
DANG & GCLC & -- & -- & not measured \\
IPC298 & CKS1B & -- & -- & not measured \\
\bottomrule
\end{tabular}
\end{table}

\subsection{Measured knockout correspondence for TTK}\label{app:ttk-knockout}
For the TTK nomination we also ask whether the requested markers that the request-path input attribution credits most are those that the measured TTK knockout moves most in the requested direction. For each requested marker other than TTK, the model value is the member-averaged two-SD input IG and the measured value is the well-mean signed displacement of Equation~\ref{eq:measured} before averaging over markers. Their Spearman correlation is 0.492 in A549 (157 markers, 10 wells), at the 98.8th percentile of the same correlation computed against each of the 4,654 other A549 knockouts with at least three wells; in DANG it is 0.278 (99 markers, 6 wells), at the 88.5th percentile of 520. This concerns one nomination and two lines; it is retrospective agreement between attribution and measured responses, not intervention efficacy.

\textbf{Cross-cell measured-signature reference.} We compute the same signed-marker score for each target in every \emph{other} cell, average wells within a cell, then give each other cell equal weight. The request cell's knockout endpoints do not contribute to that prediction. It is not a trained intervention-conditioned forward model. The common H1 scale is fixed; this comparison does not reproduce every inverse fold's training-data boundary or tune a forward model. We compare reference scores, mean inverse full logits, candidate-specific masked logits, and mean directional scores with the request cell's measured agreement over their common supported targets. Missing measurements are not imputed. The reference uses measurements the inverse model never receives and is scored on the evaluation criterion itself, so the columns are not a matched model comparison; the illustrated cases do not establish ranking-level efficacy.

\begin{table}[t]
\centering
\caption{Spearman correlation with measured requested-marker agreement over the common supported target set. The reference column is built from the same knockout's measured responses in other lines and scored on this criterion; it is a measured-response reference, not a trained forward model, so its column is not a matched comparison with the inverse scores. Exposure refers to the inverse model; reference signatures always exclude the request cell. Masked uses each candidate's own-edit-masked inverse score.}
\label{tab:forward}
\begin{tabular}{llrrrrr}
\toprule
Cell & Inverse exposure & Targets & Reference & Full logit & Masked & Directional \\
\midrule
A549 & held out & 4970 & 0.681 & 0.042 & 0.038 & 0.046 \\
YAPC & held out & 3356 & 0.132 & -0.042 & -0.039 & 0.011 \\
DANG & held out & 513 & 0.488 & 0.056 & 0.058 & 0.069 \\
HT29 & seen & 4808 & 0.173 & -0.163 & -0.163 & 0.010 \\
MCF7 & held out & 3356 & 0.242 & 0.008 & 0.008 & 0.037 \\
HCC1806 & held out & 513 & 0.532 & -0.013 & -0.013 & -0.002 \\
SNU761 & seen & 442 & 0.287 & -0.037 & -0.038 & 0.112 \\
U251MG & held out & 4247 & 0.113 & -0.099 & -0.099 & 0.051 \\
PC3 & seen & 4968 & 0.169 & 0.075 & 0.075 & 0.051 \\
A375 & held out & 4970 & 0.310 & 0.051 & 0.052 & 0.005 \\
HS944T & held out & 544 & 0.407 & 0.048 & 0.041 & 0.041 \\
IPC298 & held out & 513 & 0.306 & -0.061 & -0.039 & -0.024 \\
AGS & seen & 4145 & 0.101 & -0.137 & -0.139 & 0.018 \\
ES2 & seen & 4968 & 0.185 & -0.086 & -0.086 & -0.008 \\
\bottomrule
\end{tabular}
\end{table}

\section{Request-path population suppression}\label{app:request-population}
Under a fixed request and held-out model, we suppress one typed, directed message at one transition. Other message gates, weights, and degree factors remain fixed; downstream states are recomputed.

\textbf{Pre-declared protocol.} The question, pool, per-run computation, statistics, decision rule, and the wording for each possible outcome were frozen and committed before any production run; post-hoc diagnostics are labeled as such. The pool comprises every candidate that passes the screen of Section~\ref{sec:cases} without its illustrative mean-directional floor: in every eligible member, own-edit-masked rank $\le200$, $a_{kf}>0$, and $b_{kf}<0$. This gives 337 nominations and 371 member runs in 12 lines, of which 333 nominations and 359 runs lie in the 9 lines held out by at least one model (primary); ES2, PC3 and SNU761 are seen in training by every model (4 nominations, secondary). Nominees in a line share the same underlying request, with candidate-specific own edits removed. We therefore use the line as the unit of aggregation: correlations are averaged over members within a nomination, nominations within a line, and lines equally.

\textbf{Message selection and deletion.} Each run uses the frozen checkpoints, the request path with the nominee's own edit removed, the default conductance refinement with the reconciliation checks of Appendix~\ref{app:numerics}, and the canonical panel rule (six largest positive and six most negative messages per transition, low-$|E|$ and relation/layer-matched random controls; seed fixed; selection saved before deletion). Each deletion removes one typed directed layer-message; request and no-change scores are recomputed, and the outcome is the drop in $a_{kf}$. Qualification pilots on TTK and CKS2 reproduced their saved fields, panels, and deletion results within $5.4\times10^{-15}$. All 742 runs (both request sizes) succeeded.

\textbf{Primary result.} The pre-declared primary reading required the equal-line mean to be positive and all 9 held-out lines to be positive at a transition; it is met at every transition for both sizes.

\begin{table}[t]
\centering
\scriptsize
\setlength{\tabcolsep}{3pt}
\caption{Request-path population suppression by held-out line: mean Spearman correlation between signed message conductance and response drop. $n$: nominations. Measured size: the cell-specific 90th percentile of observed knockout displacement (0.25--0.53 SD); the pool is fixed by the two-SD screen. Permutation null: deletion effects shuffled within each panel, 1,000 times, aggregated identically.}
\label{tab:request-population}
\begin{tabular}{lrrrrrrrrr}
\toprule
 & & \multicolumn{4}{c}{Two SD} & \multicolumn{4}{c}{Measured size} \\
Line & $n$ & $0\to1$ & $1\to2$ & $2\to3$ & $3\to4$ & $0\to1$ & $1\to2$ & $2\to3$ & $3\to4$ \\
\midrule
A375 & 58 & 0.852 & 0.663 & 0.459 & 0.336 & 0.872 & 0.665 & 0.476 & 0.361 \\
A549 & 9 & 0.734 & 0.667 & 0.412 & 0.177 & 0.770 & 0.665 & 0.491 & 0.208 \\
DANG & 54 & 0.882 & 0.680 & 0.654 & 0.491 & 0.900 & 0.745 & 0.673 & 0.501 \\
HCC1806 & 57 & 0.928 & 0.863 & 0.596 & 0.372 & 0.928 & 0.872 & 0.617 & 0.393 \\
HS944T & 69 & 0.905 & 0.740 & 0.650 & 0.531 & 0.895 & 0.745 & 0.641 & 0.545 \\
IPC298 & 69 & 0.880 & 0.722 & 0.660 & 0.485 & 0.899 & 0.751 & 0.689 & 0.528 \\
MCF7 & 8 & 0.931 & 0.890 & 0.577 & 0.475 & 0.899 & 0.898 & 0.317 & 0.545 \\
U251MG & 7 & 0.923 & 0.839 & 0.634 & 0.521 & 0.924 & 0.862 & 0.670 & 0.580 \\
YAPC & 2 & 0.690 & 0.560 & 0.508 & 0.506 & 0.744 & 0.450 & 0.599 & 0.286 \\
\midrule
Equal-line mean & 333 & 0.858 & 0.736 & 0.572 & 0.433 & 0.870 & 0.739 & 0.575 & 0.438 \\
95\% bootstrap & & 0.78--0.91 & 0.64--0.81 & 0.50--0.64 & 0.35--0.51 & 0.81--0.91 & 0.62--0.83 & 0.48--0.66 & 0.32--0.54 \\
Null 95th pct & & 0.028 & 0.027 & 0.029 & 0.034 & 0.036 & 0.031 & 0.036 & 0.039 \\
\bottomrule
\end{tabular}
\end{table}

\textbf{Further pre-declared outputs.} Raw message $L_2$ size correlates with the response drop at 0.004--0.107 (two SD). One HCC1806 nomination (SLC38A6; response $8.7\times10^{-6}$) has no message above the $10^{-5}$ sign cutoff and therefore an empty panel; it and the other undefined panels (at most 9 per transition at two SD, 13 at the measured size) are recorded as missing; setting them to zero leaves every equal-line mean positive (worst case 0.41 at $3\to4$). The seen-cell lines give 0.85 / 0.75 / 0.73 / 0.43. The 19 above-floor cases are a reported subset with lower correspondence than the full pool. Correlations with request-logit and margin outcomes, which conductance does not allocate, are retained in the frozen outputs.

\textbf{Post-hoc additivity diagnostic.} Among signed extremes in the held-out lines, we compare each message's deletion effect with its own conductance (Table~\ref{tab:additivity}). Median drop/conductance ratios remain near one while identity agreement declines with depth. These descriptive ratios do not establish unbiased centering or identify a cause; this diagnostic was examined after the primary results and is not a pre-declared reading.

\begin{table}[t]
\centering
\small
\caption{Post-hoc: deletion effect versus conductance for the signed-extreme messages (two SD, 9 held-out lines). Slope and identity $R^2$ are per-panel medians of drop $=b\cdot$conductance and of agreement with drop $=$ conductance.}
\label{tab:additivity}
\begin{tabular}{lrrrrr}
\toprule
Transition & Median drop/$E$ & Slope & Identity $R^2$ & Within $2\times$ & Same sign \\
\midrule
$0\to1$ & 0.995 & 0.998 & 0.963 & 90\% & 97\% \\
$1\to2$ & 1.006 & 1.009 & 0.757 & 72\% & 90\% \\
$2\to3$ & 0.990 & 0.960 & 0.394 & 59\% & 84\% \\
$3\to4$ & 1.024 & 1.011 & 0.165 & 39\% & 73\% \\
\bottomrule
\end{tabular}
\end{table}

\textbf{Pre-declared non-additivity tests.} After the additivity diagnostic, we froze a second protocol and a seeded subsample (3 nominations per held-out line, all if fewer: 26 nominations, 37 member runs) before running two arms; all 37 runs succeeded, and with both arms switched off the code reproduced the main-run deletions within $10^{-14}$. \emph{Pairwise deletion} jointly removes every pair of signed-extreme messages within a panel; relative interaction is $|D_{ab}-(D_a+D_b)|/(|D_a|+|D_b|)$. Its equal-line median rises monotonically, 0.007, 0.026, 0.051, 0.097 from $0\to1$ to $3\to4$ (pre-declared reading: non-additivity grows with depth); at $3\to4$, 9\% of pairs exceed 0.5, and the signed interaction is centered on zero, with super- and sub-additive pairs equally common at every transition. \emph{Frozen normalization} repeats single deletions with each LayerNorm's mean and variance fixed at their intact values. Identity $R^2$ falls at the first three transitions (0.97$\to$0.71, 0.73$\to$0.57, 0.39$\to$0.34) and rises at $3\to4$ (0.14$\to$0.25); the pre-declared rule reads this as a partial contribution. This arm does not, however, isolate normalization: conductance already integrates each LayerNorm's Jacobian, so drops under frozen statistics need not equal the same conductance, which we recognized only after running it. We therefore draw no conclusion about normalization as the source of the spread.

\subsection{Separating path allocation from finite-deletion remainders}\label{app:normalization-repair}
Pairwise deletions show increasing non-additivity with depth, but a bounded follow-up does not establish normalization as a general explanation of the disagreement.

\textbf{Follow-up designed after the initial results.} After recognizing the frozen-statistics arm's mismatch, we froze a repaired implementation and analysis on the same saved two-SD panels: 37 member tasks, 26 nominations, 9 held-out lines. This is a bounded follow-up, not an independent replication or a replacement for the 333-nomination population study. The earlier pairwise measurements and frozen-statistics outputs above remain unchanged.

Let $g_e$ multiply one typed directed message, with all gates initially one, and write $S_e(x)=\left.\partial s_k(x;g_e)/\partial g_e\right|_{g_e=1}$ for endpoint-local sensitivity. For no-change and request endpoints $x_0,x_1$, define $A_e=S_e(x_1)-S_e(x_0)$, $D_e$ as the original response drop under full deletion, and $C_e$ as the saved request-path message conductance. In plain terms, $A_e$ is the local linear approximation to the request-response drop, while $D_e$ is the drop after completely removing the message. Then, exactly,
\begin{equation}
D-C=(A-C)+(D-A),\qquad (D-C)^2=(A-C)^2+(D-A)^2+2(A-C)(D-A).
\label{eq:deletion-decomposition}
\end{equation}
The first term separates path allocation from endpoint sensitivity; the second is the finite-deletion remainder. Their signed cross-term can cancel, so squared components are not additive fractions of discrepancy explained. Saved conductance retains its original quadrature error.

\textbf{Secondary LayerNorm intervention.} At each endpoint separately, replace every LayerNorm by its full-Jacobian affine tangent at the intact state, $\LN(u_*)+J_{\LN}(u_*)(u-u_*)$. Unlike freezing only mean and variance, this preserves the intact score and endpoint sensitivity $S$, while removing LayerNorm curvature for that endpoint's deletion. These endpoint-local surrogates are not one newly attributed request-path model. With tangent deletion $D^{\rm LN}$, we compare, within each signed-extreme panel,
\begin{equation}
Q=\sum_e(D_e^2+A_e^2),\quad K=\frac{\sum_e(D_e-A_e)^2}{Q},\quad
K_{\rm LN}=\frac{\sum_e(D_e^{\rm LN}-A_e)^2}{Q},\quad \Delta K=K-K_{\rm LN}.
\end{equation}
The original-model denominator is shared between arms; panels require at least three extremes and positive finite $Q$. Panel values are averaged over members, nominations, then lines equally (Table~\ref{tab:normalization-repair}). At $3\to4$, the remainder decreases in 8 of 9 lines, but increases in HS944T: equal-line $K$ changes from 0.352336 to 0.244374, $\Delta K=0.107962$, with descriptive 95\% line-bootstrap interval $[-0.005949,0.231621]$ (10,000 resamples). The frozen reading is \emph{Mixed}, not a general explanation of the depth trend. All 37 tasks succeeded; two MCF7/ABCB9 final panels have fewer than three extremes (H1: one; H3: two), so their contrasts are undefined, not failed runs. Every line retains a defined contrast.

\begin{table}[t]
\centering
\caption{Bounded follow-up: final-transition normalized finite-deletion remainder, averaged within each held-out line. Positive $\Delta K$ means a smaller remainder under tangent LayerNorm, not improved agreement with conductance. MCF7 has 2 defined nominations of 3; YAPC has 2 of 2; all other lines have 3 of 3.}
\label{tab:normalization-repair}
\begin{tabular}{lrrr}
\toprule
Line & $K$ & $K_{\rm LN}$ & $\Delta K$ \\
\midrule
A375 & 0.639799 & 0.334892 & 0.304907 \\
A549 & 0.717889 & 0.253772 & 0.464117 \\
DANG & 0.241978 & 0.135392 & 0.106587 \\
HCC1806 & 0.126292 & 0.120693 & 0.005599 \\
HS944T & 0.156073 & 0.344205 & -0.188132 \\
IPC298 & 0.197448 & 0.104083 & 0.093365 \\
MCF7 & 0.088589 & 0.082603 & 0.005986 \\
U251MG & 0.427983 & 0.420345 & 0.007638 \\
YAPC & 0.574974 & 0.403382 & 0.171593 \\
\midrule
Equal-line mean & 0.352336 & 0.244374 & 0.107962 \\
\bottomrule
\end{tabular}
\end{table}

\textbf{Original-conductance agreement is not restored.} Using the shared original scale $\sum_e(D_e^2+C_e^2)$, the final-transition squared discrepancy against $C$ changes only from 0.704446 to 0.690835 on the equal-line summary. For the existing TTK/A549 cases, the full saved panels retain negative correlations: H1 $-0.417957\to-0.298246$ (18 messages), H3 $-0.476471\to-0.497059$ (16 messages). These are descriptive continuity checks, not the signed-extreme primary contrast. TTK's normalized squared discrepancy against original conductance worsens in both models (H1 $1.190312\to1.230727$; H3 $1.343328\to1.448154$), despite smaller finite-deletion remainders; its path-allocation term has larger squared magnitude than its remainder. Thus reducing one nonlinear contribution can remove cancellation rather than improve total correspondence. The result supports a bounded contribution of LayerNorm curvature under this intervention, not normalization as a universal cause, biological causality, or better intervention ranking.

All task inputs, code identities and output receipts were verified; local model-free aggregation reproduced the tables. Maximum historical-deletion discrepancy was $8.88\times10^{-15}$, intact-score discrepancy $1.78\times10^{-15}$, and sensitivity-preservation discrepancy $2.08\times10^{-15}$. These numerical checks qualify the intervention, not a new biological mechanism.

\section{Protein-path population suppression: scope and dispositions}\label{app:population}
We now turn from constructed expression requests to observed knockout queries. Across 45,089 observed queries in 17 lines, protein-path deletion correspondence is also positive on average. Randomization, rewiring and refits assess this different contrast in Appendix~\ref{app:controls}; early-layer cross-refit agreement is weak. These are not request-path independent-seed replications.

This population uses the fixed-expression protein-reference path of Section~\ref{sec:explanation}, not the expression path of Section~\ref{sec:request-path}. The predeclared cohort comprises 45,089 cell-line by resolved-knockout-target combinations across all 17 lines. Each selected observed query explains its top predicted target, whether correct or incorrect, using only checkpoint members for which the exact observation lies in the outer test set. This is not the constructed-request screen or a validation-split study. Every terminal disposition remains in the denominator: 45,069 numerical successes and 20 numerical-convergence failures. Within-query/layer Spearman correlations use the selected message panel; defined correlations are averaged within cell and cells equally within exposure setting. Undefined values are recorded as missing, not zero. Table~\ref{tab:population-values} records the primary rounded summaries; the frozen source retains per-case and per-cell distributions, missingness, and correctness/QC strata.

\begin{table}[t]
\centering
\caption{Equal-cell mean observed-query suppression correlations. Actual is the endpoint-score drop with actual proteins; Gap subtracts the deletion effect with reference proteins, keeping expression fixed (Appendix~\ref{app:suppression}). These are within-query selected-message correlations, not a pooled correlation over all edges.}
\label{tab:population-values}
\begin{tabular}{lrrrr}
\toprule
 & \multicolumn{2}{c}{Held replicate} & \multicolumn{2}{c}{Unseen cell} \\
Transition & Actual & Gap & Actual & Gap \\
\midrule
$0\to1$ & 0.321 & 0.237 & 0.341 & 0.232 \\
$1\to2$ & 0.141 & 0.312 & 0.156 & 0.331 \\
$2\to3$ & 0.159 & 0.335 & 0.186 & 0.362 \\
$3\to4$ & 0.210 & 0.382 & 0.220 & 0.396 \\
\bottomrule
\end{tabular}
\end{table}

\begin{figure}[tp]
\centering
\includegraphics[width=\linewidth]{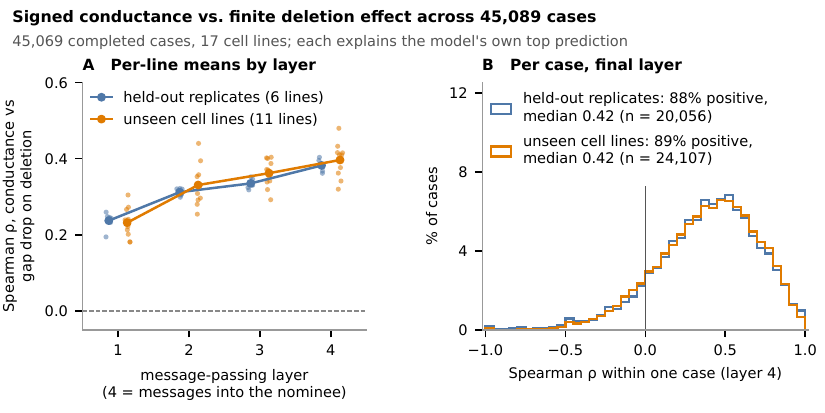}
\caption{\textbf{Population suppression on the protein path.} In 45,089 predeclared observed test cases (17 lines), each explaining the model's top prediction, a message sample fixed before deletion is deleted one message at a time. A: per-line means (dots) and equal-line means (lines) of the Spearman correlation between signed conductance and gap drop. B: per-case correlations at layer 4; undefined cases are excluded. Rank correspondence within the fitted model, not biological causality (Appendix~\ref{app:population}).}
\label{fig:population}
\end{figure}

Figure~\ref{fig:population} shows the per-line gap-drop means by layer and the per-case distribution at layer 4. Per-line means are positive in every line at every layer. At layer 4, 88\% of held-replicate and 89\% of unseen-line cases have a positive correlation, with median 0.42 in both settings; the 906 cases whose layer-4 correlation is undefined are excluded rather than set to zero. With the raw score drop instead of the gap drop, correspondence is weaker at layers 2--4 (0.21--0.22 at layer 4) and stronger at layer 1 (Table~\ref{tab:population-values}). This is rank correspondence within the fitted model; it is neither equality of magnitudes nor biological causality.

Terminal case counts are reported from the completed case computations and four per-shard artifact checks; the redundant supervisor-wide final scan did not run.

\section{Matched models, null controls, and independent refits}\label{app:controls}
\begin{table}[t]
\centering
\caption{\textbf{Predictive competence depends on exposure setting.} Tissue-balanced MRR, mean $\pm$ sample SD over three training seeds after within-seed aggregation over H1/H2/H3. Rotations are not independent seeds.}
\label{tab:baselines}
\begin{tabular}{lcc}
\toprule
Model & Held replicate & Unseen cell \\
\midrule
Expression-only MLP & $0.0309\pm0.0011$ & $0.0357\pm0.0012$ \\
Protein-informed node-local & $0.1150\pm0.0014$ & $0.1499\pm0.0016$ \\
GNN & $0.1344\pm0.0010$ & $0.1433\pm0.0023$ \\
\bottomrule
\end{tabular}
\end{table}
The matched predictive study contains 27 model/fold/seed records (18 new fits and 9 originals), with the same ordered test identities, features, candidates, and fold normalizers. The expression-only MLP sees the expression vector but no proteins or graph. The protein-informed node-local model retains protein and own-node expression features but has no cross-node messages. Three-seed means and sample SDs in Table~\ref{tab:baselines} follow the canonical fold/unit/tissue aggregation before averaging seeds. Every node-local fit selected epoch 29 or 30 and its learning-rate trials reached the 30-epoch cap. This is a fixed-budget comparison with different information access and parameterization, not proof of best-achievable graph-free performance. The historical GNN seed and audited canonical-port refits share the scientific recipe; bitwise historical replay was not established.

The predictive comparisons above do not use attribution. All null-field and refit-field results below use the separate fixed-expression protein path, not a new population of expression-path accounts.

A constructed-query graph-free check confirms that the node-local model's own-edit-masked program response is zero (as its architecture implies). It reproduces the original self-cue nominations, including H1 PYGL rank 4 and CCDC86 rank 2. This property is specific to node-local decoding: a global MLP can respond to other genes. In seven tested cell/member contexts, that MLP shares 2--7 full-request top-50 candidates with the GNN and has response-to-full-logit standard-deviation ratios 0.017--0.058. Its response ranks can place TTK or TOP2A higher despite small changes. These observations do not establish that all graph-free models fail or that inverse nomination requires our graph.

\textbf{Sensitivity to randomized models and priors.} The observed null panel fixes 34 queries, nominees, and 42 eligible members, with one trained reference and eight controls at three null seeds: 1,050 successful jobs. The primary statistic compares signed node fields on topology-defined reachable-union support, not conductance with deletion effects. Null seeds are averaged within case, cases within cell, then cells equally. One PC3 case lacks sufficient support, leaving 33 defined cases spanning all 17 cells; the final nominee-local node correlation is architecturally undefined. Table~\ref{tab:nulls} reports all eight arms. Rewired edges are not paired row by row; common node identities support the graph-null comparison. The decoder-first cascade cannot separately identify message-weight dependence. A separate 100-job constructed-null panel covers the original PYGL/CCDC86 request, not TTK/CKS2 or the observed population.

\begin{table}[t]
\centering
\caption{Trained-versus-null signed node-field correlation, equal-cell aggregation. Low values indicate sensitivity to a null intervention, not poor conductance--suppression correspondence.}
\label{tab:nulls}
\begin{tabular}{lrrrr}
\toprule
Control & $h^{(0)}$ & $h^{(1)}$ & $h^{(2)}$ & $h^{(3)}$ \\
\midrule
Reset decoder & 0.0067 & 0.0028 & -0.0022 & -0.0040 \\
Decoder + last layer & 0.0035 & 0.0033 & 0.0020 & 0.0200 \\
Decoder + last two & 0.0064 & 0.0001 & -0.0042 & 0.0066 \\
Decoder + last three & -0.0081 & -0.0020 & -0.0070 & 0.0115 \\
Decoder + all four & 0.0019 & -0.0002 & 0.0012 & 0.0450 \\
Full reinitialization & -0.0016 & 0.0034 & -0.0011 & -0.0063 \\
Protein permutation & 0.0197 & 0.0761 & 0.0974 & 0.2013 \\
Typed degree-preserving rewiring & 0.0254 & 0.0348 & 0.0271 & 0.0653 \\
\bottomrule
\end{tabular}
\end{table}

\textbf{Agreement across training seeds.} The refit panel retains the same 34 observed queries, 42 eligible members, and original nominees across three training seeds (126 successful attribution jobs). Suppression messages were selected on the original fit and reused across refits, so cross-seed differences are not a clean comparative faithfulness ranking. Table~\ref{tab:refits} reports full signed message-rank agreement; weak early-layer agreement persists between the two new fits. All within-fit equal-cell mean actual/gap suppression associations are positive, but individual negative cases remain. Full-field rank statistics do not establish stability of top-message subnetworks or pathways, and this bounded panel does not establish refit stability of the constructed cases.

\begin{table}[t]
\centering
\caption{Cross-refit agreement of full signed message rankings, equal-cell mean Spearman correlation (33 defined cases, 17 cells). This is not top-subnetwork stability.}
\label{tab:refits}
\begin{tabular}{lrrrr}
\toprule
Seed pair & $0\to1$ & $1\to2$ & $2\to3$ & $3\to4$ \\
\midrule
20260903/20260904 & 0.018 & 0.019 & 0.034 & 0.168 \\
20260903/20260905 & 0.022 & 0.024 & 0.052 & 0.206 \\
20260904/20260905 & 0.026 & 0.024 & 0.061 & 0.177 \\
\bottomrule
\end{tabular}
\end{table}

\clearpage
\section{Supporting and opposing relationships across the screen}\label{app:competing-accounts}
This descriptive inventory retains all 19 computational-screen cases, including ten meeting the measured-support criterion, seven at or below the 50th percentile, and two without measurements. It uses saved request-path conductance, with actual proteins fixed and the nominee's own expression edit removed where present. No new model fits or attribution runs were performed for this inventory.

Every listed message is computational transport from the named partner into the nominee at transition $3\to4$. Conductance is in nominee-score units: positive supports the request-minus-no-change response and negative opposes it. Prior activation/repression labels do not determine attribution signs; reverse transport does not reverse biological causality. Values across different nominees are not calibrated measures of biological importance.

\textbf{Coverage and selection.} The original 786-message inventory contains four complete fields (TTK/A549, TTK/DANG, CKS2/HS944T, CKS2/IPC298) and fifteen saved stratified panels. Table~\ref{tab:competing-accounts} preserves that export and its coverage labels: panel maxima are not full-field maxima, and absence is not zero. We subsequently recovered complete saved fields for SORL1/A549, CAMSAP2/A549, TOP2A/A549 and CHEK1/HS944T, without new inference. Together with TTK/A549 and CKS2/HS944T, these supply six expanded accounts covering ten eligible partition fits. TTK is shown in Figure~\ref{fig:ttk-case}; the other five figures follow below. The CAMSAP2/DANG contrast remains panel-based. Each figure's executed request leaves the nominee unchanged, removing its own edit where present; TTK/A549 was already unedited in the original request. The examples were selected retrospectively for illustration, not as an unbiased estimate of nomination success.

\textbf{Reading the figures.} Panel A shows node conductance at $h^{(3)}$, before the final message step; it is not expression-input IG. Selected partners are ordered by signed mean, with per-fit values and arrows only for requested edits. Panel B shows selected positive route witnesses and selected opposing final messages. Twenty routes is a display budget, not a candidate-rank cutoff. Node fill denotes the executed request; relationship colors and glyphs retain the biological prior. Blue ticks mark negative final incoming message conductance, with badges for fit-specific opposition. An unticked relationship need not be positive in every layer or fit. Positive widths use peak positive mean layer credit, and opposing overlays use mean absolute final-message credit on the same within-case logarithmic scale. Layers are not added into a conserved flow; full selection rules and exact values accompany the figures.

\textbf{What the distinctions establish.} The fitted accounts distinguish relationship type, input edit from later state, and recurring from fit-dependent contributions. Literature then supplies biological questions to examine, rather than a one-to-one mapping from a database channel to a mechanism. CD58/CD2 illustrates why these questions remain separate from measured support: in YAPC, regulatory credit supports ($+0.6522/{+}0.4967$) and physical credit opposes ($-0.0288/{-}0.0541$) in H1/H3, despite only 23.1st-percentile measured agreement. Cross-cell comparisons change the fitted model as well as the cell/request; H1/H3 share a seed. The case captions distinguish the computational finding, experimental context and an unresolved biological question.

\begin{table}[!htbp]
\centering\small
\caption{\textbf{Opposition in the complete final incoming-message fields.} Counts include every typed message into the nominee at $3\to4$, separately by fit. Negative magnitude share is the sum of absolute negative conductance divided by total absolute message conductance. It excludes residual state and is not a percentage of the prediction. The selected blue ticks are not a census of this field.}
\label{tab:six-case-signs}
\begin{tabular}{llrr}
\toprule
Case & Fit & Negative / all & Negative magnitude share \\
\midrule
TTK/A549 & H1 & 20/99 & 13.8\% \\
TTK/A549 & H3 & 40/99 & 24.7\% \\
SORL1/A549 & H1 & 12/26 & 35.5\% \\
SORL1/A549 & H3 & 18/26 & 19.8\% \\
CKS2/HS944T & H3 & 16/56 & 16.3\% \\
CAMSAP2/A549 & H1 & 5/16 & 2.1\% \\
CAMSAP2/A549 & H3 & 0/16 & 0.0\% \\
TOP2A/A549 & H1 & 34/197 & 1.7\% \\
TOP2A/A549 & H3 & 36/197 & 6.6\% \\
CHEK1/HS944T & H3 & 97/233 & 13.5\% \\
\bottomrule
\end{tabular}
\end{table}

\begin{table}[!htbp]
\centering\small\setlength{\tabcolsep}{4pt}
\caption{Original 786-message inventory: largest available final incoming contribution of each sign, by fit. C/P mark complete/panel coverage in that export, before recovery of the expanded case fields. Panel maxima are not full-field maxima. SF/SP/SR: STRING functional/physical/regulatory; CA/CR: CollecTRI activation/repression; rev: reverse transport. Percentiles and wells concern measured other-marker agreement. Exact full records will be released with the code.}
\label{tab:competing-accounts}
\begin{tabular}{@{}>{\raggedright\arraybackslash}p{.28\linewidth}>{\raggedright\arraybackslash}p{.34\linewidth}>{\raggedright\arraybackslash}p{.32\linewidth}@{}}
\toprule Case / evidence & Supporting & Opposing \\ \midrule
\textbf{TTK} / A549\par C; 98.2 pct.; n=10 & H1: MAD1L1 [SR-rev] +0.42\par H3: MAD2L1 [SR-rev] +0.161 & H1: E2F4 [CR] -0.0575\par H3: E2F4 [CR] -0.071 \\[5pt]
\textbf{TOP2A} / A549\par P; 97.0 pct.; n=21 & H1: NFYB [CR] +0.766\par H3: NFYB [CR] +0.771 & H1: TP53 [SP] -0.00444\par H3: CDK1 [SR] -0.0533 \\[5pt]
\textbf{CAMSAP2} / A549\par P; 97.0 pct.; n=12 & H1: CKAP2L [SP] +0.831\par H3: CKAP2L [SP] +1.13 & H1: CLASP1 [SF] -0.0093\par H3: none in saved scope \\[5pt]
\textbf{CHEK1} / HS944T\par P; 96.4 pct.; n=1 & H3: TIMELESS [SR] +1.33 & H3: FOXM1 [CA] -0.0538 \\[5pt]
\textbf{CKS2} / HS944T\par C; 95.0 pct.; n=6 & H3: SKP2 [SR-rev] +0.223 & H3: GMNN [SP] -0.0427 \\[5pt]
\textbf{CKS2} / IPC298\par C; 90.4 pct.; n=6 & H1: SKP2 [SR-rev] +0.359 & H1: SKP1 [SR-rev] -0.056 \\[5pt]
\textbf{CDK2} / HS944T\par P; 89.5 pct.; n=9 & H3: JUND [CR] +0.203 & H3: MCM3 [SR-rev] -0.0906 \\[5pt]
\textbf{TTK} / DANG\par C; 82.1 pct.; n=6 & H2: CDCA8 [SR-rev] +0.266 & H2: MAD2L1 [SR-rev] -0.031 \\[5pt]
\textbf{SORL1} / A549\par P; 78.6 pct.; n=6 & H1: APP [SP] +0.0436\par H3: APP [SR-rev] +0.674 & H1: GGA2 [SF] -0.0155\par H3: GGA1 [SP] -0.0416 \\[5pt]
\textbf{CAMSAP2} / DANG\par P; 76.3 pct.; n=6 & H2: CKAP2L [SP] +0.488 & H2: CKAP2 [SF] -0.0301 \\[5pt]
\bottomrule\end{tabular}
\end{table}
\begin{table}[!htbp]
\centering\small\setlength{\tabcolsep}{4pt}
\caption{Final incoming contrasts, continued; definitions and scope as in Table~\ref{tab:competing-accounts}.}
\begin{tabular}{@{}>{\raggedright\arraybackslash}p{.28\linewidth}>{\raggedright\arraybackslash}p{.34\linewidth}>{\raggedright\arraybackslash}p{.32\linewidth}@{}}
\toprule Case / evidence & Supporting & Opposing \\ \midrule
\textbf{CBLB} / IPC298\par P; 39.5 pct.; n=6 & H1: LCK [SR-rev] +0.0803 & H1: ZAP70 [SP] -0.0348 \\[5pt]
\textbf{AVPR2} / A375\par P; 32.3 pct.; n=5 & H2: AVP [SR] +0.444 & H2: REN [SF] -0.0402 \\[5pt]
\textbf{CDC7} / HS944T\par P; 31.7 pct.; n=2 & H3: CLSPN [SR-rev] +0.184 & H3: CDT1 [SF] -0.0266 \\[5pt]
\textbf{CD2} / YAPC\par P; 23.1 pct.; n=5 & H1: CD58 [SR] +0.652\par H3: CD58 [SR] +0.497 & H1: CD58 [SP] -0.0288\par H3: CD58 [SP] -0.0541 \\[5pt]
\textbf{TPR} / DANG\par P; 19.2 pct.; n=5 & H2: MAD2L1 [SP] +0.0473 & H2: CDK1 [SF] -0.0245 \\[5pt]
\textbf{WWTR1} / IPC298\par P; 13.4 pct.; n=6 & H1: ASNS [CR-rev] +0.327 & H1: PPP2R1B [SR] -0.045 \\[5pt]
\textbf{CALCA} / IPC298\par P; 1.3 pct.; n=6 & H1: CALCRL [SP] +0.0762 & H1: CALCR [SR-rev] -0.113 \\[5pt]
\textbf{GCLC} / DANG\par P; unmeasured & H2: GCLM [SP] +1.15 & H2: FOSL1 [CR] -0.643 \\[5pt]
\textbf{CKS1B} / IPC298\par P; unmeasured & H1: SKP2 [SR-rev] +0.294 & H1: CKS2 [SP] -0.0369 \\[5pt]
\bottomrule\end{tabular}
\end{table}

\clearpage
\begin{figure}[p]
\centering
\includegraphics[width=\linewidth]{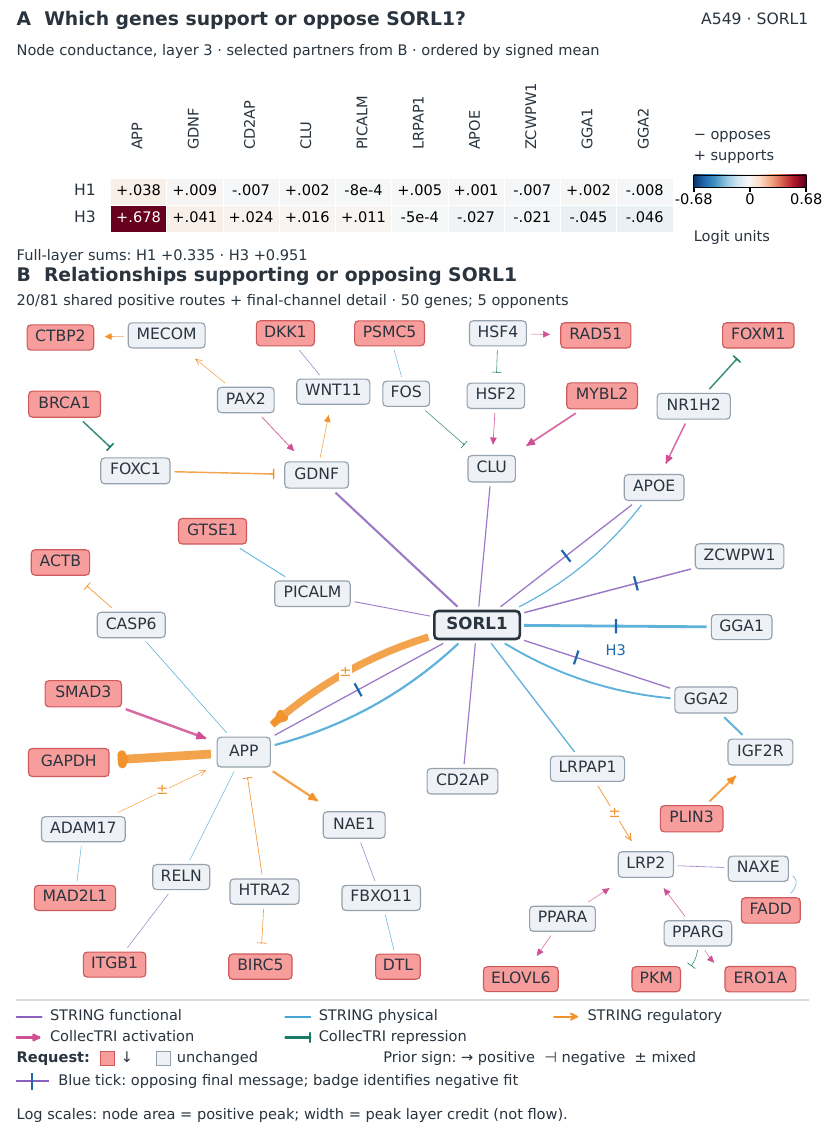}
\caption[SORL1 signed request-path account]{\textbf{SORL1/A549: one partner has more than one contribution.} APP physical support ($+0.0436/{+}0.0108$) coexists with functional opposition ($-0.0149/{-}0.0062$) in H1/H3; reverse-regulatory support is much larger in H3. Experimental work links SORLA-dependent APP retention and processing to GGA/PACS-1 adaptor interactions \citep{sorla2007}. This makes APP retention and altered adaptor-dependent routing useful biological questions, without equating either process with one database channel. Do trafficking changes accompany the requested marker response, or can the two separate? The cited experiments do not establish an A549 cancer mechanism. Measured knockout agreement reaches the 78.6th percentile (six wells). The display includes shared opposition and the strongest per-fit opponents; badges identify opposition confined to one fit.}
\label{fig:six-sorl1}
\end{figure}

\clearpage
\begin{figure}[p]
\centering
\includegraphics[width=\linewidth]{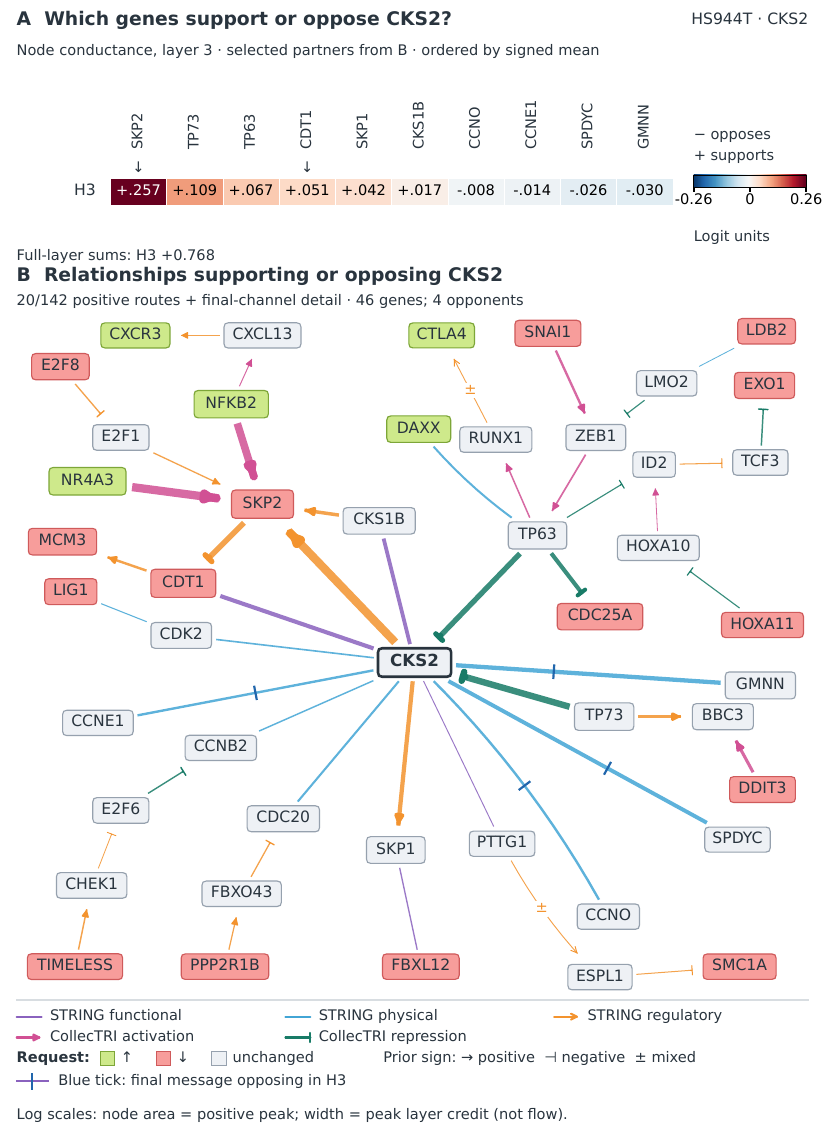}
\caption[CKS2 signed request-path account]{\textbf{CKS2/HS944T: distinguish an input edit from the transported state.} In H3, SKP2 down has negative input IG ($-0.3291$), but its final reverse-regulatory message supports CKS2 ($+0.2225$): its later state carries information from other genes. In a mouse-derived system, CKS2 protects p27 and restrains cyclin A/CDK2 activity, with a role distinct from CKS1; its loss increases replication speed and damage \citep{cks2012}. Does the requested marker response accompany altered p27/CDK regulation, rather than a uniform reduction in proliferation? The regulatory channel does not establish direct SKP2--CKS2 binding. Measured agreement is at the 95.0th percentile (six wells). Appendix~\ref{app:crossline} retains the IPC298 comparison and Appendix~\ref{app:amplitude} the weaker response at smaller amplitude; this figure shows only H3.}
\label{fig:six-cks2}
\end{figure}

\clearpage
\begin{figure}[p]
\centering
\includegraphics[width=\linewidth]{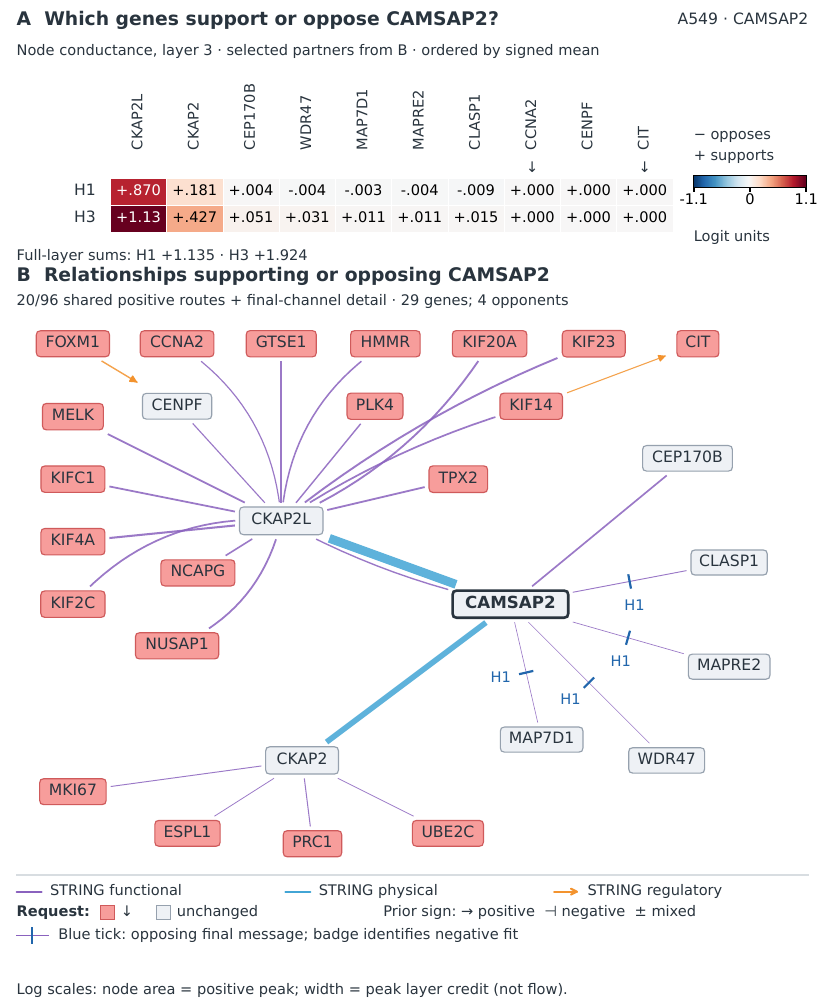}
\caption[CAMSAP2 signed request-path account]{\textbf{CAMSAP2/A549: recurring cytoskeletal support, fit-dependent opposition.} CKAP2L and CKAP2 physical support recurs in H1/H3. CLASP1, MAPRE2, WDR47 and MAP7D1 oppose in H1 but support in H3; H3 has no negative final incoming message. CAMSAP2 silencing increases cisplatin sensitivity in A549/H292, and CAMSAP2 expression partially rescues miR-145-5p-associated sensitization \citep{camsap2026}. Those experiments motivate a drug-response question, separate from whether cytoskeletal changes accompany the requested expression response. They do not establish CKAP2/CKAP2L mediation. A549 knockout agreement reaches the 97.0th percentile (12 wells). The physical-support/functional-opposition CKAP2 example in the main text concerns DANG/H2, not this figure; its agreement is 76.3rd percentile (six wells), with cell and model changing together.}
\label{fig:six-camsap2}
\end{figure}

\clearpage
\begin{figure}[p]
\centering
\includegraphics[width=\linewidth]{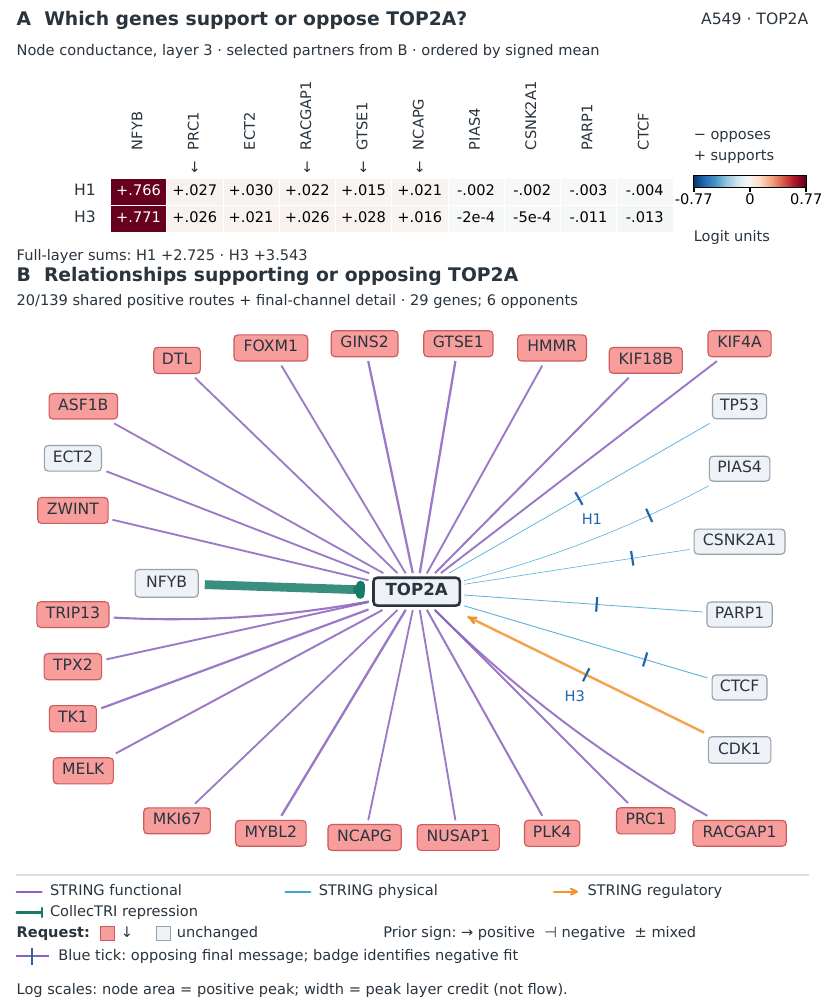}
\caption[TOP2A signed request-path account]{\textbf{TOP2A/A549: recurring NFYB support and fit-dependent CDK1 credit.} NFYB repression-prior credit supports both fits ($+0.766/{+}0.771$); CDK1 regulatory credit supports H1 but opposes H3 ($+0.103/{-}0.053$). Thus prior repression does not determine attribution sign. In leukemia and rhabdomyosarcoma models, miR-485-3p-dependent NFYB reduction increased TOP2A expression and sensitivity to Top2 inhibitors \citep{nfytop2011}. This motivates separating transcriptional regulation of TOP2A abundance from CDK1-associated cell-cycle effects. Does either account track the requested marker response after accounting for cell-cycle composition? The study does not establish these relationships in A549, and TOP2A depletion differs from topoisomerase-poison treatment. Measured knockout agreement reaches the 97.0th percentile (21 wells). Fit badges distinguish shared from fit-specific opposition.}
\label{fig:six-top2a}
\end{figure}

\clearpage
\begin{figure}[p]
\centering
\includegraphics[width=\linewidth]{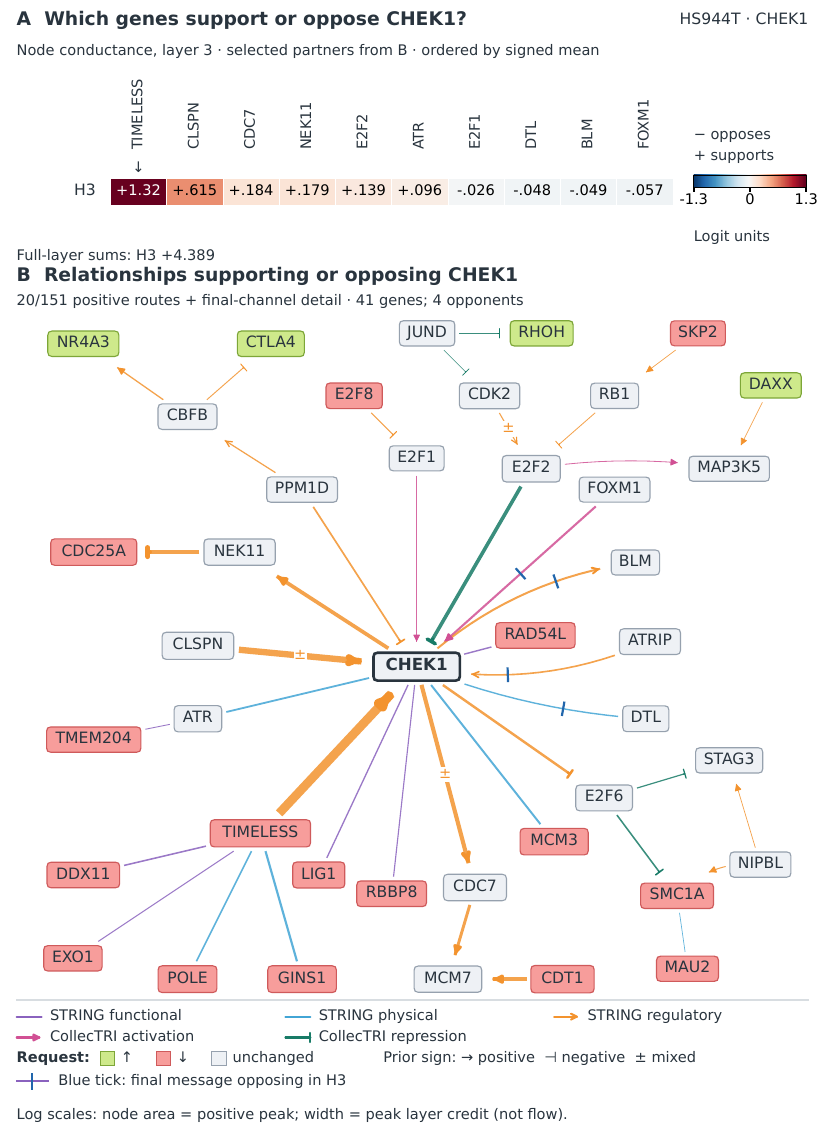}
\caption[CHEK1 signed request-path account]{\textbf{CHEK1/HS944T: checkpoint partners receive different credit.} TIMELESS and CLSPN support H3; FOXM1, BLM, ATRIP and DTL supply selected opposition. Direct final-channel detail keeps CLSPN visible even though it is absent from the selected complete positive routes. Claspin supports stress-induced CHK1 activation \citep{claspin2003}; TIMELESS interacts with CHK1 and ATR--ATRIP in checkpoint studies \citep{timeless2005}. Supporting TIMELESS and opposing ATRIP therefore distinguish contributions within connected checkpoint biology. Do similar checkpoint effects produce different responses among the requested markers? These studies do not establish the mechanism in HS944T or validate an opposing sign. Measured agreement reaches the 96.4th percentile, but comes from one well and requires replication. This is a single-fit account, not a recurrence test.}
\label{fig:six-chek1}
\end{figure}

\clearpage
\end{document}